\documentclass{article} % For LaTeX2e

\usepackage{iclr2027_conference,times}

\newif\ifpreprint
\preprinttrue
\ifpreprint
  \iclrfinalcopy
\fi
\usepackage{amsmath,amsfonts,bm}

\def\eqref#1{equation~\ref{#1}}
\def\1{\bm{1}}

\DeclareMathAlphabet{\mathsfit}{\encodingdefault}{\sfdefault}{m}{sl}
\SetMathAlphabet{\mathsfit}{bold}{\encodingdefault}{\sfdefault}{bx}{n}

\newcommand{\E}{\mathbb{E}}

\newcommand{\R}{\mathbb{R}}

\newcommand{\KL}{D_{\mathrm{KL}}}

\newcommand{\Cov}{\mathrm{Cov}}
\DeclareMathOperator*{\argmin}{arg\,min}

\newcommand{\cut}[1]{}

\usepackage[framemethod=TikZ]{mdframed}
\mdfdefinestyle{MyFrame}{%
    linecolor=black,
    outerlinewidth=0.5pt,
    roundcorner=1pt,
    innertopmargin=2pt, %\baselineskip,
    innerbottommargin=2pt, %\baselineskip,
    innerrightmargin=7pt,
    innerleftmargin=7pt,
    backgroundcolor=black!0!white}

\mdfdefinestyle{MyFrame2}{%
    linecolor=white,
    outerlinewidth=1pt,
    roundcorner=2pt,
    innertopmargin=10pt,
    innerbottommargin=0.5pt,
    innerrightmargin=10pt,
    innerleftmargin=10pt,
    backgroundcolor=black!3!white}

\mdfdefinestyle{MyFrameEq}{%
    linecolor=white,
    outerlinewidth=0pt,
    roundcorner=0pt,
    innertopmargin=0pt,
    innerbottommargin=0pt,
    innerrightmargin=7pt,
    innerleftmargin=7pt,
    backgroundcolor=black!3!white}

\mdfdefinestyle{FrameAdded}{%
    linecolor=black,
    outerlinewidth=0pt,
    roundcorner=0pt,
    innertopmargin=0pt,
    innerbottommargin=0pt,
    innerrightmargin=0pt,
    innerleftmargin=0pt,
    backgroundcolor=PastelGreen}

\definecolor{customwhite}{HTML}{FCFBF7}
\definecolor{customturq}{HTML}{1D9D79}
\definecolor{customorange}{HTML}{D96002} 
\definecolor{custombeige}{HTML}{d6c9b1} 

\definecolor{custompurple}{HTML}{AEADF0}
\definecolor{customwhite2}{HTML}{fbf9f4}
\definecolor{customblue}{HTML}{4D9DE0} 
\usepackage{silence}
\usepackage[utf8]{inputenc}
\usepackage[T1]{fontenc}
\usepackage{tabularx}
\usepackage{hyperref}
\usepackage{lipsum}
\usepackage{url}
\usepackage{placeins}
\usepackage{booktabs}
\usepackage{soul}
\usepackage{amsfonts}
\usepackage{nicefrac}
\usepackage{microtype}
\usepackage{natbib}
\usepackage{proba}
\usepackage{wrapfig}
\usepackage{caption}
\usepackage{amssymb}
\usepackage{pifont}

\usepackage{subcaption}
\usepackage{float}
\usepackage{rotating}
\usepackage{etoolbox}
\usepackage{xparse}
\usepackage{grffile}
\usepackage{amsmath}
\usepackage{xspace}
\usepackage{euscript}
\usepackage{enumitem}
\usepackage{mathtools}
\usepackage{tikz}
\usepackage{tablefootnote}
\usepackage[flushleft]{threeparttable}
\usepackage{multirow}
\usepackage[title,toc,page,header]{appendix}
\usepackage{array, booktabs}
\usepackage{comment}
\usepackage{graphicx}
\usepackage{adjustbox}
\usepackage{dsfont}
\usepackage{amsmath,amsthm}
\usepackage{balance}
\usepackage{multicol}
\usepackage{nameref}
\usepackage{pdfpages}
\usepackage{braket}
\usepackage[normalem]{ulem}
\usepackage{bbding}
\usepackage[capitalise]{cleveref}   
\usepackage{xfrac}
\usepackage[usestackEOL]{stackengine}
\usepackage{indentfirst}
\usepackage{csquotes}
\usepackage{pgf}
\usepackage{varwidth}
\usepackage{verbatimbox}
\usepackage{verbatim}
\usepackage{bm}
\usepackage{algorithm}
\usepackage{algorithmic}

\AtBeginEnvironment{algorithmic}{}
\usepackage[colorinlistoftodos,prependcaption,textsize=small]{todonotes}

\usepackage{empheq}
\definecolor{myblue}{rgb}{.8, .8, 1}

\definecolor{pastelblue}{RGB}{76,113,175}
\definecolor{pastelgreen}{RGB}{144,238,144}
\definecolor{pastelred}{RGB}{196,78,82}
\definecolor{pastelgrey}{RGB}{230,230,230}
\definecolor{pastelbeige}{RGB}{243,236,221}
\definecolor{pastelpurple}{RGB}{154,139,192}
\definecolor{salmon}{RGB}{250, 128, 114}
\definecolor{darkgreen}{rgb}{0,0.6,0}
\definecolor{darkred}{rgb}{0.5,0,0}
\definecolor{verylightgreen}{HTML}{F6FFF9}
\definecolor{verylightred}{HTML}{FFF4F3}
\definecolor{verylightgray}{HTML}{F4F6F6}
\definecolor{babyblueeyes}{rgb}{0.63, 0.79, 0.95}
\definecolor{lightpink}{rgb}{1.00, 0.714, 0.757}

\usepackage{tikzpeople}
\usetikzlibrary{shapes,decorations,arrows,calc,arrows.meta,fit,positioning}

\tikzset{
    -Latex,auto,node distance =1 cm and 1 cm,semithick,
    state/.style ={ellipse, draw, minimum width = 0.7 cm},
    point/.style = {circle, draw, inner sep=0.04cm,fill,node contents={}},
    bidirected/.style={Latex-Latex,dashed},
    el/.style = {inner sep=2pt, align=left, sloped}
}

\newtheorem{lemma}{Lemma}

\newtheorem{proposition}{Proposition}

\makeatletter
\def\thmt@refnamewithcomma #1#2#3,#4,#5\@nil{%
	\@xa\def\csname\thmt@envname #1utorefname\endcsname{#3}%
	\ifcsname #2refname\endcsname
	\csname #2refname\expandafter\endcsname\expandafter{\thmt@envname}{#3}{#4}%
	\fi}
\makeatother
\Crefname{conjecture}{Conjecture}{Conjectures}
\Crefname{definition}{Definition}{Definitions}
\Crefname{observation}{Observation}{Observations}
\Crefname{assumption}{Assumption}{Assumptions}
\Crefname{axiom}{Axiom}{Axioms}
\Crefname{case}{Case}{Cases}
\Crefname{claim}{Claim}{Claims}
\Crefname{conclusion}{Conclusion}{Conclusions}
\Crefname{condition}{Condition}{Conditions}
\Crefname{criterion}{Criterion}{Criteria}
\Crefname{exercise}{Exercise}{Exercises}
\Crefname{example}{Example}{Examples}
\Crefname{notation}{Notation}{Notations}
\Crefname{problem}{Problem}{Problems}
\Crefname{property}{Property}{Properties}
\Crefname{remark}{Remark}{Remarks}
\Crefname{solution}{Solution}{Solutions}
\Crefname{summary}{Summary}{Summaries}
\Crefname{motivation}{Motivation}{Motivations}
\Crefname{query}{Query}{Queries}
\crefname{algocf}{Alg.}{Algs.}
\Crefname{algocf}{Algorithm}{Algorithms}

\newcommand*\dbar[1]{\overline{\overline{\lower0.2ex\hbox{$#1$}}}}

\def\cD{{\mathcal{D}}}

\def\cL{{\mathcal{L}}}

\def\cN{{\mathcal{N}}}

\DeclareFontFamily{U}{BOONDOX-calo}{\skewchar\font=45 }
\DeclareFontShape{U}{BOONDOX-calo}{m}{n}{
  <-> s*[1.05] BOONDOX-r-calo}{}
\DeclareFontShape{U}{BOONDOX-calo}{b}{n}{
  <-> s*[1.05] BOONDOX-b-calo}{}
\DeclareMathAlphabet{\mathcalb}{U}{BOONDOX-calo}{m}{n}
\SetMathAlphabet{\mathcalb}{bold}{U}{BOONDOX-calo}{b}{n}
\DeclareMathAlphabet{\mathbcalb}{U}{BOONDOX-calo}{b}{n}

\usepackage{cancel}
\renewcommand{\paragraph}[1]{{\noindent \textbf{#1.}}}

\newcommand{\methodname}{KC-maps}
\newcommand{\methodlongname}{Kolmogorov-Chapman Maps}
\newcommand{\method}{\methodname\xspace}
\newcommand{\methodlong}{\methodlongname\xspace}

\definecolor{pastel_purple}{HTML}{756FB3}
\definecolor{pastel_green}{HTML}{1D9D79}

\colorlet{PastelPurpleLight}{pastel_purple!15!white}
\colorlet{PastelGreenLight}{pastel_green!15!white}
\sethlcolor{PastelPurpleLight}
\sethlcolor{PastelGreenLight}

\ifdefined\nohyperref\else\ifdefined\hypersetup
  \definecolor{mydarkblue}{rgb}{0,0.08,0.45}
  \hypersetup{ %
    colorlinks=true,
    linkcolor=mydarkblue,
    citecolor=mydarkblue,
    filecolor=mydarkblue,
    urlcolor=mydarkblue,
    }

  \fi
\fi

\makeatletter
\g@addto@macro\normalsize{%
  \setlength\abovedisplayskip{4pt plus 2pt minus 2pt}%
  \setlength\belowdisplayskip{4pt plus 2pt minus 2pt}%
  \setlength\abovedisplayshortskip{0pt plus 2pt}%
  \setlength\belowdisplayshortskip{2pt plus 2pt minus 1pt}%
}
\makeatother

\let\originalleft\left
\let\originalright\right
\renewcommand{\left}{\mathopen{}\mathclose\bgroup\originalleft}
\renewcommand{\right}{\aftergroup\egroup\originalright}

\global\long\def\norm#1{\left\lVert #1\right\rVert }

\definecolor{antiquefuchsia}{rgb}{0.57, 0.36, 0.51}
\definecolor{amethyst}{rgb}{0.6, 0.4, 0.8}

\newcommand{\deriv}[2]{\frac{\partial #1}{\partial #2}}

\newcommand{\mean}{\mathbb{E}}
\newcommand{\var}{{\rm I\kern-.3em D}}

\newcommand{\cond}{\,|\,}
\newcommand{\Normal}{\mathcal{N}}

\newcommand{\eps}{\varepsilon}
\newtheorem*{theorem*}{Theorem}
\newtheorem*{proposition*}{Proposition}
\newtheorem*{example*}{Example}
\DeclareMathSymbol{\shortminus}{\mathbin}{AMSa}{"39}
\usepackage[many]{tcolorbox}

\usepackage[framemethod=TikZ]{mdframed}
\mdfdefinestyle{mybox}{%
    linecolor=white,
    outerlinewidth=1pt,
    roundcorner=2pt,
    innertopmargin=10pt,
    innerbottommargin=0.5pt,
    innerrightmargin=10pt,
    innerleftmargin=10pt,
    backgroundcolor=black!3!white}

\tcbuselibrary{theorems}

\newtcbtheorem[number within=section]{mytheo}{Statement}%
{colback=green!5,colframe=green!35!black,fonttitle=\bfseries}{th}

\newtcbox{\hlroundedpurple}[1][PastelPurpleLight]{ % Default color is CustomColorALight
  on line,
  arc=4pt, % Adjusts the roundness of the corners
  colback=#1,
  colframe=#1,
  boxrule=0pt,
  boxsep=0pt,
  left=1pt, % Adjusts left padding
  right=1pt, % Adjusts right padding
  top=2pt, % Adjusts top padding
  bottom=2pt, % Adjusts bottom padding
  leftrule=0pt,
  rightrule=0pt,
  toprule=0pt,
  bottomrule=0pt,
}

\newtcbox{\hlroundedgreen}[1][PastelGreenLight]{ % Default color is CustomColorALight
  on line,
  arc=4pt, % Adjusts the roundness of the corners
  colback=#1,
  colframe=#1,
  boxrule=0pt,
  boxsep=0pt,
  left=1pt, % Adjusts left padding
  right=1pt, % Adjusts right padding
  top=2pt, % Adjusts top padding
  bottom=2pt, % Adjusts bottom padding
  leftrule=0pt,
  rightrule=0pt,
  toprule=0pt,
  bottomrule=0pt,
}

\newtcbox{\hlrounded}[1][PastelPurpleLight]{ % Default color is CustomColorALight
  arc=4pt, % Adjusts the roundness of the corners
  colback=#1,
  colframe=#1,
  boxrule=0pt,
  boxsep=0pt,
  left=1pt, % Adjusts left padding
  right=1pt, % Adjusts right padding
  top=2pt, % Adjusts top padding
  bottom=2pt, % Adjusts bottom padding
  leftrule=5pt,
  rightrule=5pt,
  toprule=5pt,
  bottomrule=5pt,
}

\makeatletter
\let\save@mathaccent\mathaccent
\newcommand*\if@single[3]{%
  \setbox0\hbox{${\mathaccent"0362{#1}}^H$}%
  \setbox2\hbox{${\mathaccent"0362{\kern0pt#1}}^H$}%
  \ifdim\ht0=\ht2 #3\else #2\fi
  }
\newcommand*\rel@kern[1]{\kern#1\dimexpr\macc@kerna}
\newcommand*\widebar[1]{\@ifnextchar^{{\wide@bar{#1}{0}}}{\wide@bar{#1}{1}}}
\newcommand*\wide@bar[2]{\if@single{#1}{\wide@bar@{#1}{#2}{1}}{\wide@bar@{#1}{#2}{2}}}
\newcommand*\wide@bar@[3]{%
  \begingroup
  \def\mathaccent##1##2{%
    \let\mathaccent\save@mathaccent
    \if#32 \let\macc@nucleus\first@char \fi
    \setbox\z@\hbox{$\macc@style{\macc@nucleus}_{}$}%
    \setbox\tw@\hbox{$\macc@style{\macc@nucleus}{}_{}$}%
    \dimen@\wd\tw@
    \advance\dimen@-\wd\z@
    \divide\dimen@ 3
    \@tempdima\wd\tw@
    \advance\@tempdima-\scriptspace
    \divide\@tempdima 10
    \advance\dimen@-\@tempdima
    \ifdim\dimen@>\z@ \dimen@0pt\fi
    \rel@kern{0.6}\kern-\dimen@
    \if#31
      \overline{\rel@kern{-0.6}\kern\dimen@\macc@nucleus\rel@kern{0.4}\kern\dimen@}%
      \advance\dimen@0.4\dimexpr\macc@kerna
      \let\final@kern#2%
      \ifdim\dimen@<\z@ \let\final@kern1\fi
      \if\final@kern1 \kern-\dimen@\fi
    \else
      \overline{\rel@kern{-0.6}\kern\dimen@#1}%
    \fi
  }%
  \macc@depth\@ne
  \let\math@bgroup\@empty \let\math@egroup\macc@set@skewchar
  \mathsurround\z@ \frozen@everymath{\mathgroup\macc@group\relax}%
  \macc@set@skewchar\relax
  \let\mathaccentV\macc@nested@a
  \if#31
    \macc@nested@a\relax111{#1}%
  \else
    \def\gobble@till@marker##1\endmarker{}%
    \futurelet\first@char\gobble@till@marker#1\endmarker
    \ifcat\noexpand\first@char A\else
      \def\first@char{}%
    \fi
    \macc@nested@a\relax111{\first@char}%
  \fi
  \endgroup
}
\makeatother
\usepackage[utf8]{inputenc} % allow utf-8 input
\usepackage[T1]{fontenc}    % use 8-bit T1 fonts
\usepackage{hyperref}       % hyperlinks
\usepackage{url}            % simple URL typesetting
\usepackage{booktabs}       % professional-quality tables
\usepackage{amsfonts}       % blackboard math symbols
\usepackage{nicefrac}       % compact symbols for 1/2, etc.
\usepackage{microtype}      % microtypography
\usepackage{xcolor}         % colors
\usepackage{etoc}           % appendix-only table of contents

\title{\methodlong: Trajectory-Free \\ Learning for Simulating Molecular Dynamics}

\author{%
  \textbf{Majdi Hassan\textsuperscript{1,\,2}\thanks{Correspondence:
  \href{mailto:majdi.hassan@mila.quebec}{\texttt{majdi.hassan@mila.quebec}}} \quad
  Marta Skreta\textsuperscript{1,\,2} \quad
  Dominique Beaini\textsuperscript{1,\,2,\,4}\thanks{Work performed while at Valence Labs.} \quad
  Kirill Neklyudov\textsuperscript{1,\,2,\,3}} \\[0.5em]
  \textsuperscript{1}Mila -- Quebec AI Institute \quad
  \textsuperscript{2}Universit\'e de Montr\'eal \quad
  \textsuperscript{3}Institut Courtois \quad
  \textsuperscript{4} Valence Labs
}

\begin{document}

\maketitle
\ifpreprint
  % The template sets its conference header inside \maketitle.
  \fancyhead{}
  \renewcommand{\headrulewidth}{0pt}
  % Keep natural float spacing instead of stretching gaps to fill each page.
\fi
\etocdepthtag.toc{mainbody}

\begin{abstract}
  Stochastic differential equations (SDEs) describe stochastic processes across a wide range of scientific domains, including physics, biology, and finance. Although individual numerical integration steps are inexpensive, observing meaningful events or transitions can require millions of sequential steps, resulting in substantial wall-clock simulation time and poor hardware utilization. Deep learning offers an alternative by amortizing many integration steps into a single neural network forward pass, potentially enabling much faster simulation.  However, existing supervised approaches typically require long simulated trajectory data, which restricts their applicability. In this work, we introduce \methodlong (\method), a method for learning the transition kernel of an SDE that relies only on the \textit{trajectory-free} marginal samples together with the known drift and diffusion coefficients. Given an initial time and state of a process, \method models its distribution integrated to any given time, corresponding to the weak solution of the SDE.
  To recover the correct dynamics, \method satisfies two key properties: (i) it satisfies the boundary conditions, which we incorporate directly into the neural network architecture, and (ii) it respects the Kolmogorov--Chapman equation, which provides an efficient training objective. We empirically demonstrate that \method learns to integrate the dynamics of peptide molecular systems and outperforms classical molecular dynamics in terms of GPU-hours. The implementation is available at \href{https://github.com/majhas/kc-maps}{\texttt{https://github.com/majhas/kc-maps}}.
\end{abstract}

\section{Introduction}
\label{sec:introduction}

Stochastic differential equations (SDEs) provide a fundamental framework for describing stochastic processes, with applications across physics, finance, and biology. 
A prominent example is molecular dynamics \citep{leimkuhler_molecular_2015,hollingsworth_molecular_2018}, where Langevin equations describe the evolution of the atomic coordinates and momenta under interatomic forces and random fluctuations.
Classical approaches, such as Euler-Maruyama \citep{Kloeden1992NumericalSO} and Langevin splitting schemes \citep{leimkuhler2013robust}, approximate the dynamics through a sequence of short-time updates and provide systematic control over discretization error.
However, such methods are inherently sequential, with computational cost scaling with the number of integration steps. In molecular dynamics, femtosecond-scale time steps ($10^{\scalebox{0.85}{$\scriptstyle -15$}}$\,seconds) are typically required to maintain numerical stability and resolve fast motions such as bond vibrations, while conformational changes and transitions between metastable states can occur over orders-of-magnitude longer timescales, e.g. $10^{\scalebox{0.85}{$\scriptstyle -6$}}$\,seconds. Reaching these slow phenomena can therefore require an enormous number of sequential force evaluations, making long-horizon molecular simulation computationally expensive.

Recently, numerous approaches were developed to learn surrogates that can accelerate molecular simulation by predicting the evolution over larger time intervals. Generative modeling approaches \citep{klein_timewarp_2023, schreiner_implicit_2023} learn conditional distributions of future molecular configurations from simulated trajectories. 
Other approaches \citep{bigi2026flashmd} learn direct deterministic updates of the positions and momenta.
However, both formulations learn these transitions from pairs of states extracted from reference trajectories.
Generating these pairs still requires numerical integration with time steps typically of the order of $10^{-15}\,\mathrm{s}$, incurring substantial computational cost to produce the training data. 
Moreover, equilibrium samples can also be obtained using methods such as sequential Monte Carlo \citep{moral_sequential_2002} or parallel tempering \citep{Hukushima_1996}. Yet, their sampling transitions do not necessarily follow the physical dynamics of interest, so the resulting data may not provide the time-dependent pairs required by supervised methods. 

The underlying SDE, however, provides a different source of supervision. Although the finite-time transition kernel is generally unavailable in closed form, the drift and diffusion specify the dynamics locally. 
This raises the possibility of learning finite-time transitions from the governing SDE itself, without first generating reference trajectories. Recent approaches in flow-based generative modeling \citep{ho_denoising_2020, song_score-based_2021, lipman_flow_2023, albergo_building_2023} use consistency relations to connect the underlying vector fields with the finite-time maps \citep{song_consistency_2023,geng_mean_2025,boffi_how_2025}, enabling accelerated generative sampling. Hamiltonian Flow Maps \citep{ripken2026hfm} apply this consistency principle to molecular dynamics, learning Hamiltonian evolution from independent states and instantaneous force fields without trajectory supervision.
For stochastic dynamics, we seek an analogous formulation at the level of conditional distributions, connecting their finite-time evolution to the known drift and diffusion of the SDE.

In this work, we present \methodlong (\method), a trajectory-free method for learning finite-time transitions of SDEs. 
Our method uses \textit{uncorrelated} samples of the initial state together with the known vector field and diffusion, without requiring trajectory supervision or a prescribed probability path. We derive a distributional consistency objective from the Kolmogorov--Chapman equation, enforcing agreement between transitions composed over different time scales; for example, two successive $5\,\mathrm{fs}$ transitions should agree in distribution with a single $10\,\mathrm{fs}$ transition. To enable tractable density evaluation and satisfy SDE boundary conditions, we parameterize transitions densities using a conditional normalizing flow with a Gaussian reference kernel. The parameterization is constrained to recover the appropriate short-time behavior of the underlying SDE, while additional design choices enable learning and sampling over substantially longer time intervals.

The formulation of \method applies to general SDEs. Here, we focus on molecular Langevin dynamics because of both its computational difficulty and its importance to applications such as drug discovery. We first conduct an empirical study on a small molecular system to identify important design choices required for stable and accurate learning, including the training objective, design of the normalizing flow, and stabilization techniques for sampling over long time intervals. We then apply \method to larger scale molecular systems and highlight the following main results:

\begin{itemize}
\item Using only equilibrium samples and force-field evaluations, without trajectory supervision, we learn transition distributions over intervals of up to $100\,\mathrm{fs}$ for alanine dipeptide (AL2), tetrapeptide (AL4), and hexapeptide (AL6), replacing up to $100$ sequential MD integration steps with a single learned transition.
\item We demonstrate higher simulation throughput and faster exploration of conformational modes. On AL6, our method achieves approximately $2.3\times$ aggregate simulation throughput ($\mathrm{ns}$ per GPU-hour) of MD on a single L40s GPU and reaches three of the four TICA modes while MD baseline still misses two modes after an hour. Additionally, our method is able to recover all four modes of AL6 with longer simulations despite the training data only covering three of the four modes.
\item  Beyond equilibrium sampling, \method captures molecular kinetics, with mean first passage times (MFPT), mode autocorrelations, and transition probabilities closely matching ground-truth MD. 
\end{itemize}

\begin{figure}[t]
  \centering
  \includegraphics[width=\linewidth]{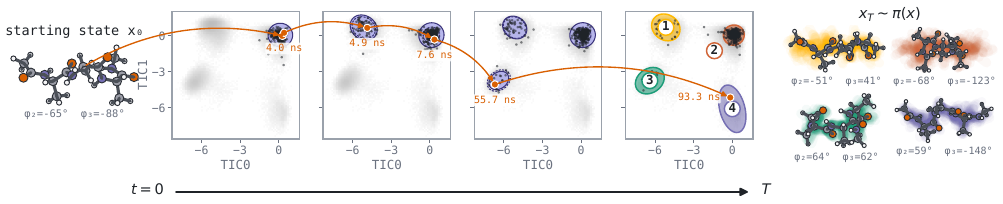}
  \caption{A single \method rollout explores the landscape of metastable alanine tetrapeptide conformers. Starting from a single conformer $X_0$ (left), \method is rolled out for 200 ns.  Orange arrows trace the first visit to each newly discovered basin. Representative conformers from each basin are shown on the right, colored by basin identity, along with their backbone torsion angles ($\Phi_2$, $\Phi_3$).\vspace{-1.5em}}
  \label{fig:kcm_overview}
\end{figure}
   
\section{Background}
\label{sec:background}

\paragraph{Stochastic Differential Equations}
Stochastic differential equations (SDEs) describe the stochastic process $X_t$ via the differential equation for their time-increments $dX_t$, i.e.
\begin{align}
    dX_t &= v_t(X_t)\,dt + g_t\,dW_t, \label{eq:sde}
\end{align}
where  $g_t\in\R^{d\times d}$ is a state-independent diffusion matrix, and $W_t$ is standard Brownian motion in $\R^d$. We will refer to the solution of the Cauchy problem of this SDE as a \textit{transition kernel}, i.e. given the initial condition $X_s = x$, we denote the density of the distribution of $X_t$ as $p_{s,t}(y\cond x) \coloneqq p(X_t = y\cond X_s = x)$.

\paragraph{Molecular Dynamics}
The formalism of SDEs is especially useful for describing the evolution of molecular systems. Their state is described by the position $X_t$ and momentum $P_t$ and its time-evolution is modeled by the second-order dynamics, which we will write in the matrix form
\begin{align}
    \begin{bmatrix}
    dX_t\\
    dP_t
    \end{bmatrix} =~& 
    \begin{bmatrix}
    M^{-1}P_t\\
    v_t(X_t) - \gamma P_t
    \end{bmatrix}
    dt + 
    \begin{bmatrix}
    0 & 0\\
    0 & g_t\\
    \end{bmatrix}
    dW_t\,,
\end{align}
where $v_t$ is the force field driving the system, $\gamma$ is the friction coefficient, $M$ is the diagonal mass matrix, and $g_t$ is the diffusion coefficient that depends on the temperature of the system. For $v_t(X) = -\nabla U(X)$ where $U(X)$ is a scalar potential and $g_t=\sqrt{2\gamma k_B\,T\,M}$, the stationary distribution of $X_t$ is the Boltzmann distribution, i.e. the equilibrium of the molecular system. Thus, in practice, simulating this SDE for long enough time one can sample from the Boltzmann distribution describing different states --- conformations --- of the system.

\paragraph{Normalizing Flows}
Generative modeling of the empirical data has two principal components: maximizing the model likelihood on the observed data and generating new samples under  the learned density model. Normalizing flows satisfy both criteria by parameterizing bijective maps with tractable Jacobians. The building block of normalizing flows is the \textit{affine coupling layer} $\Psi_k$, which transforms the state $x^{(k)} = [x_1^{(k)}, x_2^{(k)}]$ into $x^{(k+1)} = \Psi_k(x^{(k)})$ as
\begin{align}
  x_1^{(k+1)} &= x_1^{(k)},\\
  x_2^{(k+1)} &= x_2^{(k)}
      \odot\exp\,\bigl(a_k(x_1^{(k)}; \theta, c)\bigr)
      +b_k(x_1^{(k)}; \theta, c),\label{eq:affine_layer}
\end{align}
where $a_k(\cdot ; \theta, c): \R^{d_1} \to \R^{d_2}$ and $b_k(\cdot ; \theta, c): \R^{d_1} \to \R^{d_2}$ are functions parameterized by the neural networks with parameters $\theta$ and, optionally, taking a condition $c$ as an input. This layer is invertible, i.e. one can find $[x_1^{(k)}, x_2^{(k)}]$ from $[x_1^{(k+1)}, x_2^{(k+1)}]$, and the Jacobian of this transformation is available in a closed-form without additional computations
\begin{align}
    \log \det \left|\deriv{\Psi_k(x^{(k)})}{x^{(k)}}\right| =
    \log \begin{bmatrix}
        I & 0\\
        \deriv{x_2^{(k+1)}}{x_1^{(k)}} & I\exp\,\bigl(a_k(x_1^{(k)}; \theta, c)\bigr)
    \end{bmatrix}
    = \sum_i a_k(x_1^{(k)}; \theta, c)_i\,.
\end{align}
Finally, normalizing flows parameterize a density $p(x \cond \theta, c)$ as a push-forward map of some simple density $p^{\mathrm{ref}}(x)$ through the composition of affine layers $\Psi = \Psi_K \circ \ldots \circ \Psi_1$
\begin{align}
    \log p(x\cond \theta, c) = \log p^{\mathrm{ref}}(\Psi^{-1}(x; \theta, c)) - \sum_k \log \det \left|\deriv{\Psi_k(x^{(k)})}{x^{(k)}}\right|\,.
\end{align}

\section{\methodlong} \label{sec:methods}

Our method is designed around the classical definition of a diffusion process as a Markov process with continuous trajectories and infinitesimal first and second moments prescribed by the drift and diffusion coefficient \citep{kolmogoroff1931analytischen, feller1937theorie}. The process defined this way is a weak solution of the corresponding SDE \citep{Stroock1979MultidimensionalDP}. Therefore, our parametric model has to satisfy two conditions: (i) parameterize a valid Markov process, i.e. satisfy the Kolmogorov--Chapman equation, which we enforce through training (ii) the short-time behavior has to match the corresponding SDE, i.e. commit no jumps and have same first two moments, which we incorporate directly into our parameterization. We now proceed to formulating this statement formally.

For the SDE defined in \cref{eq:sde}, the infinitesimal change of the density $p_{s,t}(y\cond x)$ is defined by the following boundary conditions. $\forall \eps > 0$, we have
\begin{align}
    \lim_{\tau \rightarrow 0}\frac{1}{\tau} \int_{\|y-x\|>\eps} dy\, p_{s,s+\tau}(y\mid x) =~& 0\,, \tiny\texttt{ (delta function with no jumps)}\normalsize \label{eq:boundary-local}\\
    \lim_{\tau \rightarrow 0}\frac{1}{\tau} \int_{\|y-x\|\leq \eps} dy\, p_{s,s+\tau}(y\mid x)(y-x) =~& v_s(x)\,,\tiny\texttt{ (matching first moment)}\normalsize \label{eq:boundary-moment_1}\\
    \lim_{\tau \rightarrow 0}\frac{1}{\tau} \int_{\|y-x\|\leq \eps} dy\, p_{s,s+\tau}(y\mid x)(y-x)(y-x)^\top =~& g_sg_s^\top\,, \tiny\texttt{ (matching second moment)}\normalsize \label{eq:boundary-moment_2}
\end{align}
i.e. for $t \to s$, the density converges to the delta-function and makes no jumps, the change of the first moment equals the vector field and the change of the second moment equals the diffusion coefficient. Furthermore, any Markov transition kernel satisfies the Kolmogorov--Chapman equation
\begin{equation}
    p_{s,t}(y\mid x) = \int dx'\,p_{r,t}(y\mid x')\,p_{s,r}(x'\mid x)\,.
    \label{eq:chapman-kolmogorov}
\end{equation}
This equation is the stochastic analog of the semi-group consistency condition in the ODEs, which is used in the Flow Maps literature \citep{boffi_how_2025,woo2026riemannian}.

These two conditions define the SDE \cref{eq:sde} and vice versa, the weak solution of an SDE satisfies both these conditions. The first statement we prove in \cref{app:kernel-identification} using the classical result of probability theory \citep{Stroock1979MultidimensionalDP}, the second statement is a textbook statement. 
\begin{proposition}[\textnormal{Identification of the transition kernel}]
\label{prop:kernel-identification}
    Assume that $g_s$ is continuous in $s$ and that $v_s(x)$ is continuous, locally Lipschitz in $x$, and of at most linear growth, uniformly for $s$ in compacts. Let $p_{s,t}(y\cond x)$ be a measurable family of nonnegative, normalized conditional densities. Suppose it satisfies the boundary conditions in \cref{eq:boundary-local,eq:boundary-moment_1,eq:boundary-moment_2}, with limits locally uniform in $(s,x)$, and the Kolmogorov--Chapman equation in \cref{eq:chapman-kolmogorov}. Then $p_{s,t}(y\mid x)$ coincides (almost everywhere) with the transition density of the SDE in \cref{eq:sde}. Conversely, the SDE's transition kernel satisfies both the boundary conditions and \cref{eq:chapman-kolmogorov}.
\end{proposition}

\subsection{Kolmogorov--Chapman Constraints (Our Objective)}
\label{sec:objectives}

We parameterize our transition kernel $q_{s,t}(y\cond x; \theta)$ such that the boundary conditions are satisfied (see the following section; \cref{app:learning-boundary-conditions} details the case for learning the drift instead), which leaves the Kolmogorov--Chapman equation as the sole optimization objective. The most straightforward way to train the model would be to minimize the discrepancy between both sides of the Kolmogorov--Chapman equation for the parameterized model, i.e.
\begin{align}
    \cL_{\mathrm{KC}}(\theta) &=
        \KL\left(
            \int dx'\,q_{r,t}(y \mid x';\theta) q_{s,r}(x' \mid x;\theta)
            \,\middle\|\,
            q_{s,t}(y\mid x;\theta)
        \right)\,, \forall\; x,t,s,r\,.\label{eq:ck-kl}
\end{align}
However, as written, this function does not have an unbiased estimator of the gradient. Therefore, we propose to use the ``stop-gradient trick'' ubiquitously employed in the generative modeling literature \citep{geng_mean_2025,boffi_how_2025}. Setting $\eta = \mathrm{stopgrad}(\theta)$ as the parameters in the first argument of the KL-divergence, we get (see \cref{app:ck-kl-loss})
\begin{align}
    \cL_{\mathrm{KC}}(\theta) &=
        -\mean_{y}
            [\log q_{s,t}(y\mid x;\theta)]\,, \text{ where } y \sim q_{r,t}(y \mid x';\eta)\,,\text{ and } x' \sim q_{s,r}(x' \mid x;\eta) \,.\label{eq:ck-kl-mle}
\end{align}
The objective in \cref{eq:ck-kl}, along with the parameterization that satisfies the boundary conditions, is globally minimized by the transition kernel of the SDE \cref{eq:sde}, as shown by a classical result in stochastic processes \cref{app:ck-minimizer}.

\paragraph{Probability-flow objective}
An alternative objective follows from the forward Kolmogorov equation written as a probability-flow ODE. Using the model's transition score, we define the velocity 
\begin{align}
    v_{\mathrm{PF}}(y) &=v_t(y)-\tfrac12 g_tg_t^\top\nabla_y\log q_{s,t}(y\mid x;\theta).
\end{align}
For a generated sample $y_\theta$, define the average velocity $u_\theta=(y_\theta - x)/\tau$. Since $\tfrac{dy_\theta}{d\tau}=u_\theta + \tau \deriv{u_\theta}{\tau}$, using the same "stop-gradient trick" and matching the sample velocity to $v_\mathrm{PF}$ gives
\begin{align}
    \cL_{\mathrm{PF}}(\theta)
    &=\E_y\left[
        \left\|u_\theta-
        \operatorname{stopgrad}\left(
            v_{\mathrm{PF}}(y_\theta)-\tau\frac{du_\theta}{d\tau}
        \right)\right\|^2
    \right],
    \qquad y \sim q_{s,t}(y \mid x;\theta)  \;\;\text{and}\;\;  \forall x,\ s<t. \label{eq:pf-obj}
\end{align}
However, we use the KC-KL objective in \cref{eq:ck-kl-mle} as our primary training objective and compare it with the probability-flow objective in our experiments.

\subsection{Boundary Constraints (Our Parameterization)}
\label{sec:parameterization}

We parameterize our transition kernel $q_{s,t}(y\cond x;\theta)$ as a normalizing flow with time-scaled prior and time-scaled affine coupling layers. This guarantees the boundary constraints for $t \to s$ and puts no restrictions on larger time scales. For simplicity, we consider the autonomous SDE with the drift $v(x)$ and diffusion coefficient $gg^\top$, then our model is
\begin{align}
    y = x + \tau v(x) + \Psi(\sqrt{\tau} \cdot g z; x, \theta)\,, \;\; z \sim \Normal(z\cond 0, I)\,,\label{eq:our_model}
\end{align}
where $\tau = t-s$ is the new time variable for the autonomous system and $\Psi(\cdot;x,\theta)$ is the normalizing flow forward map. Importantly, we add \textit{time-gates} to the definition of the affine coupling layers, which then become defined as
\begin{align}
  x_1^{(k+1)} &= x_1^{(k)},\\
  x_2^{(k+1)} &= x_2^{(k)}
      \odot\exp\,\left(\frac{(\omega \tau)^{n_a}}{1 + (\omega \tau)^{n_a}}\cdot a_k(x_1^{(k)}; x, \theta)\right)
      + \frac{(\omega \tau)^{n_b}}{1 + (\omega \tau)^{n_b}}\cdot b_k(x_1^{(k)}; x, \theta)\,,\label{eq:time_affine_layer}
\end{align}
where $\omega$, $n_a$ and $n_b$ are hyperparameters of the gates. The key property of the gates is that they behave as $(\omega \tau)^n$ for small $\tau$, which controls the boundary conditions and converge to $1$ for large $\tau$, which implies full expressivity of the model at large time-scales.

\begin{proposition}
    \textnormal{(Time-gates satisfy the boundary constraints)}
    \label{lemma:gate-orders}
    Let $q_{s,t}(y\mid x;\theta)$ be the density of the random variable $y$ defined in \cref{eq:our_model}, and the map $\Psi = \Psi_K \circ \ldots \circ \Psi_1$ consists of the composition of affine coupling layers with time-gates of powers $n_a$ and $n_b$. Assuming that $a_k$ and $b_k$ are uniformly bounded over their inputs and sufficiently small $\tau > 0$, for
    \begin{align}
          n_a > \frac{1}{2}\,, \text{ and } n_b > 1\,,
    \end{align}
    $q_{s,t}(\cdot\mid x;\theta)$ satisfies the boundary conditions in \cref{eq:boundary-local,eq:boundary-moment_1,eq:boundary-moment_2}. See proof in \cref{app:gate-orders}.
\end{proposition}

\paragraph{Second-order dynamics}
For second-order Langevin dynamics, we use unit masses in the following expressions for simplicity and extend the parameterization to the joint source state $x=(X, P)$,
\begin{align}
    y &= x + m(x, \tau) + \Psi\left(L(\tau)z; x,\theta\right),  \;\; z \sim \Normal(z\cond 0, I)\,,\label{eq:our_model_second}
\end{align}
where $m(x, \tau)$ and $L(\tau)L(\tau)^\top$ are mean displacement and covariance of the Gaussian reference obtained holding the vector field at $v(X)$. The expressions for $m$ and $L$ are derived in \cref{app:second-order-gaussian-reference}. Moreover, we additionally scale the position shift with a free-flight term $\ell_i(\tau)$ where $b_{k,i} = \ell_{i}(\tau) \tilde{b}_{k,i}$ where the factor scales like $O(\tau)$ at small $\tau$. We provide more details in \cref{app:free-flight-term}.

The coupling layers in \cref{eq:time_affine_layer} alternate between position and momentum updates. We use $(n_a,n_b)=(3, 3)$ for position updates, and $(n_a, n_b)=(2,3)$ for momentum updates. This choice of gate orders preserves the boundary constraints for the second-order SDE (see \cref{app:midpoint-reference}). 

\subsection{Practical Design Choices}
\label{sec:design}
We use the design choices listed below to improve training stability and accuracy of learned transitions.

\paragraph{Stabilizing the Gaussian reference}
To stabilize sampling over longer intervals, we use a capped duration $h(\tau)$ to limit transport, force contributions, and position noise in the Gaussian reference. We set $h(\tau)=\tau$ for $\tau\leq\tau_0$ and $h(\tau)=\tau_0+\delta\tanh((\tau-\tau_0)/\delta)$ otherwise, so the duration saturates at $\tau_0+\delta$. The reference is unchanged near $\tau=0$, preserving its boundary conditions. We give the exact rescaling in \cref{app:duration-schedule}.

Furthermore, two-step targets in \cref{eq:ck-kl} are formed from shorter transitions, motivating a more accurate short-time reference.
For source position $X$ and momentum $P$, we evaluate the vector field at $v(X_{\mathrm{mid}})$, where $X_{\mathrm{mid}}=X+\tfrac12 h(\tau)P$ (see details in \cref{app:gaussian-reference}). We ablate duration capping and midpoint evaluation separately in \cref{tab:al2_ablation_reference}.

\begin{table}[!t]
\centering
\small
\setlength{\tabcolsep}{3pt}
\renewcommand{\arraystretch}{1.08}
\caption{\textbf{Ablation study on AL2.} NLL on samples from MD simulation and energy $\mathcal{E}$-$\mathcal{W}_2$ ($k_\mathrm{B}T$) over $64$ source states with $256$ samples per state compared to $256$ independent MD trajectories with a $0.1$ fs timestep.
Results are averaged over $\tau = 10, 20, \dots, 100$ fs, with lower values being better.
Each subtable compares the full model with variants that change only the indicated component, keeping all other settings fixed. All models were trained with a reduced compute budget of 250K steps. Best value in each column within each subtable is \textbf{bold}.}
\label{tab:al2_ablation}
\begin{subtable}[t]{0.485\linewidth}
\centering
\caption{\textbf{Gaussian reference.}}
\label{tab:al2_ablation_reference}
\begin{tabularx}{\linewidth}{@{}>{\raggedright\arraybackslash}Xrr@{}}
\toprule
Variant & NLL $\downarrow$ & $\mathcal{E}$-$\mathcal{W}_2$ $\downarrow$ \\
\midrule
Full model & $\mathbf{-0.122}$ & $\mathbf{1.33}$ \\
\midrule
w/o duration capping & $0.350$ & $9.7\times10^3$ \\
w/o midpoint evaluation & $-0.053$ & $4.71$ \\
\bottomrule
\end{tabularx}
\end{subtable}\hfill
\begin{subtable}[t]{0.485\linewidth}
\centering
\caption{\textbf{Time prior.}}
\label{tab:al2_ablation_time_prior}
\begin{tabularx}{\linewidth}{@{}>{\raggedright\arraybackslash}Xrr@{}}
\toprule
Variant & NLL $\downarrow$ & $\mathcal{E}$-$\mathcal{W}_2$ $\downarrow$ \\
\midrule
Full model & $\mathbf{-0.122}$ & $\mathbf{1.33}$ \\
\midrule
Log-uniform only & $-0.086$ & $1.37$ \\
Uniform only & $-0.087$ & $2.64$ \\
\bottomrule
\end{tabularx}
\end{subtable}

\vspace{0.6em}
\begin{subtable}[t]{0.485\linewidth}
\centering
\caption{\textbf{Input features.}}
\label{tab:al2_ablation_features}
\begin{tabularx}{\linewidth}{@{}>{\raggedright\arraybackslash}Xrr@{}}
\toprule
Variant & NLL $\downarrow$ & $\mathcal{E}$-$\mathcal{W}_2$ $\downarrow$ \\
\midrule
Full model & $\mathbf{-0.122}$ & $1.33$ \\
\midrule
w/o mean features & $-0.056$ & $1.34$ \\
w/o covariance features & $-0.094$ & $1.45$ \\
w/o residual normalization & $-0.106$ & $\mathbf{1.28}$ \\
\bottomrule
\end{tabularx}
\end{subtable}\hfill
\begin{subtable}[t]{0.485\linewidth}
\centering
\caption{\textbf{Normalizing flow design.}}
\label{tab:al2_ablation_flow}
\begin{tabularx}{\linewidth}{@{}>{\raggedright\arraybackslash}Xrr@{}}
\toprule
Variant & NLL $\downarrow$ & $\mathcal{E}$-$\mathcal{W}_2$ $\downarrow$ \\
\midrule
Full model & $\mathbf{-0.122}$ & $1.33$ \\
\midrule
w/o source-frame alignment & $0.028$ & $1.77$ \\
Lower-order gates & $-0.107$ & $\mathbf{1.25}$ \\
\bottomrule
\end{tabularx}
\end{subtable}
\end{table}

\paragraph{Time prior}
During training, we sample times from a mixture $\tau\sim w\,\mathtt{LogUniform}[T_{\min},T_{\max}]+(1-w)\,\mathtt{Uniform}[T_{\min},T_{\max}]$, where $T_{\min}$ and $T_{\max}$ are the minimum and maximum sampled times, with $0<T_{\min}<T_{\max}$ and $w\in[0,1]$.
The uniform component upweights longer intervals, while $T_{\max}$ is gradually increased early in training so the model first learns shorter transitions. We compare this mixture against either distribution alone in \cref{tab:al2_ablation_time_prior}.

\paragraph{Input features}
Alongside the source state and elapsed time $\tau$, we condition the flow on the Gaussian reference mean displacement and covariance. Flow inputs are normalized by the reference marginal standard deviations for predicting affine scale and shift. We provide more details in \cref{app:model-details} and separately ablate mean and covariance features and input normalization in \cref{tab:al2_ablation_features}.

\paragraph{Normalizing flow design}
The fixed rate $\omega$ controls the time $\tau$ at which the flow corrections take effect, without changing their prescribed short-time orders. We set this rate at $\omega=333\,\mathrm{ps}^{-1}$ from empirical analysis. For second-order dynamics, we also compare our gate choices with a lower-order alternative, labeled ``Lower-order gates,'' which uses $n_a=1$ for both log scales and $n_b=2$ for both shifts, as shown in \cref{tab:al2_ablation_flow}. Finally, we compare the displacement parameterization used here with standard and whitened flows on AL2. We briefly describe these alternatives and their corresponding gates in \cref{app:flow-parameterizations}, with results in \cref{fig:al2-parameterization}.

\paragraph{Equivariance}
To account for global translation and rotation symmetries, we center positions at the center of mass and apply the flow in an orthonormal frame determined by the source configuration \citep{kaba2023equivariance,jumper_highly_2021}.
We rotate the output back to the original orientation. The frame construction and treatment of the center-of-mass constraint are detailed in \cref{app:model-details}.

\paragraph{Stabilizing training}
To stabilize learning from model-generated targets, we use $\eta=\operatorname{EMA}(\theta)$ in $\mathcal{L}_{\mathrm{KC}}$.
We compare different EMA decay rates, including $\eta=\operatorname{stopgrad}(\theta)$, in \cref{fig:al2-ema-sweep}. 

\paragraph{Detailed-Balance Loss}
The Kolmogorov--Chapman objective constrains how the kernel composes across $\tau$ but does not ensure that stationary distribution is $\pi$. Since $\pi$ is known, we add a detailed-balance objective as done in \cite{klein_timewarp_2023}, which holds for underdamped Langevin dynamics up to momentum reversal $R(X,P)=(X,-P)$, and implies that $\pi$ is the stationary distribution of the kernel:
\begin{align}
    \label{eq:db-obj}
    \cL = \cL_{\mathrm{KC}} + \lambda \, \cL_{DB}, \quad \cL_{DB} = \KL \left(q_{s,t}(y \mid x)\pi(x) \| q_{s,t}(Rx \mid Ry)\pi(y)\right).
\end{align}
We optimize $\cL_{DB}$ with a low-variance gradient estimator following \citep{tucker2019doubly}. The detailed-balance loss is only applied to towards the end of training with more details in \cref{app:objective}.

\paragraph{Inference-time stabilization}
Following the use of inference-time corrections in HFM \citep{ripken2026hfm}, we apply additional corrections to limit accumulated errors during long simulations. We describe these corrections in \cref{app:inference-stabilization} and summarize a simulation step in \cref{alg:inference}.

\section{Experiments}
\label{sec:experiments}

\subsection{Baselines}
Throughout our experiments, we primarily consider Langevin dynamics as a baseline and other trajectory-free methods at the target distribution in our comparisons. Long reference simulations provide the transition samples and equilibrium statistics used for evaluation. For training, we assume we have access to the unpaired initial states $x\sim \pi(x)$. We match training FLOPs between KC-KL and PF-ODE on Lennard-Jones systems and between our method and HFM on AL4 and AL6 (see \cref{tab:hfm-training-compute}). For AL2, we use the released HFM checkpoint \citep{ripken2026hfm}.

\begin{table}[H]
  \centering
  \caption{Performance of \method on LJ-13 and LJ-55. Published baseline results are taken from \citet{potaptchik2025tilt}. Metrics are the 2-Wasserstein distance between energy distributions ($\mathcal{E}$-$\mathcal{W}_2$) and the geometric 2-Wasserstein distance (Geo-$\mathcal{W}_2$).
  In all settings, the target temperature is $T=1$; best values are \textbf{bolded} and the second best values are \underline{underlined}. Dashes indicate unavailable results.}
  \label{tab:lj-results}
  \small
  \setlength{\tabcolsep}{3pt}
  \renewcommand{\arraystretch}{1.0}
  \begin{tabular}{@{}lcccc@{}}
    \toprule
    & \multicolumn{2}{c}{LJ-13} & \multicolumn{2}{c}{LJ-55} \\
    \cmidrule(lr){2-3}\cmidrule(l){4-5}
    Method
      & $\mathcal{E}$-$\mathcal{W}_2$ $\downarrow$
      & Geo-$\mathcal{W}_2$ $\downarrow$
      & $\mathcal{E}$-$\mathcal{W}_2$ $\downarrow$
      & Geo-$\mathcal{W}_2$ $\downarrow$ \\
    \midrule
    MALA  & $0.325$ & $\mathbf{1.51}$ & $3.40$ & $\mathbf{3.80}$ \\
    \midrule
    \multicolumn{5}{@{}l}{\textit{Source temperature $T_S=4$}} \\
    PITA \citep{akhound2025pita}
      & $2.260$ & $1.65$ & $-$ & $-$ \\
    ITM \citep{potaptchik2025tilt}
      & $0.879$ & \underline{$1.54$} & $26.37$ & $3.98$ \\
    \method ($\cL_{\mathrm{KC}}$)  & $\mathbf{0.212}$ & $\mathbf{1.51}$ & \underline{$1.51$} & \underline{$3.82$} \\
    \midrule
    \multicolumn{5}{@{}l}{\textit{Source temperature $T_S=1$}} \\
    \method ($\mathcal{L}_{\mathrm{KC}}$) & $0.266$ & $\mathbf{1.51}$ & $\mathbf{1.19}$ & \underline{$3.82$} \\
    \method ($\mathcal{L}_{\mathrm{PF}}$) & \underline{$0.254$} & $\mathbf{1.51}$ & $140.37$ & $4.29$ \\
    \bottomrule
  \end{tabular}
  \vspace{-1.5em}
\end{table}

\subsection{Lennard-Jones}

To demonstrate our method to learn the map for the first-order SDE, we consider the Lennard-Jones 13-particle (LJ-13) and 55-particle (LJ-55) systems. Building on the setup in \citep{akhound2025pita}, we train our flow to model transitions of particle positions at the target temperature $T=1$ and diffusion coefficient $D=\tfrac{g^2}{2} = 0.02$. We compare two different training settings: (i) configurations sampled at the target temperature for the initial state during training, and (ii) configurations sampled from a higher temperature $T_S=4$ as the initial states, motivated by what is done in \citep{akhound2025pita,potaptchik2025tilt}. In both settings, the vector field and diffusion correspond to the SDE at the target temperature. This comparison demonstrates that our method can learn the target dynamics without requiring training configurations from the target distribution.

We evaluate 2-Wasserstein distances on 1D energy histograms ($\mathcal{E}$-$\mathcal{W}_2$) and 3D coordinates ($\text{Geo}$-$\mathcal{W}_2$). We compare Metropolis--Hastings sampling \citep{hastings1970} with our learned kernel against MALA \citep{roberts1996exponential} and results from Tilt Matching \citep{potaptchik2025tilt}. Results are reported in \cref{tab:lj-results}, with sampling and evaluation details in \cref{app:setup-eval}.

\begin{figure}[t]
  \centering
  \includegraphics[width=\linewidth]{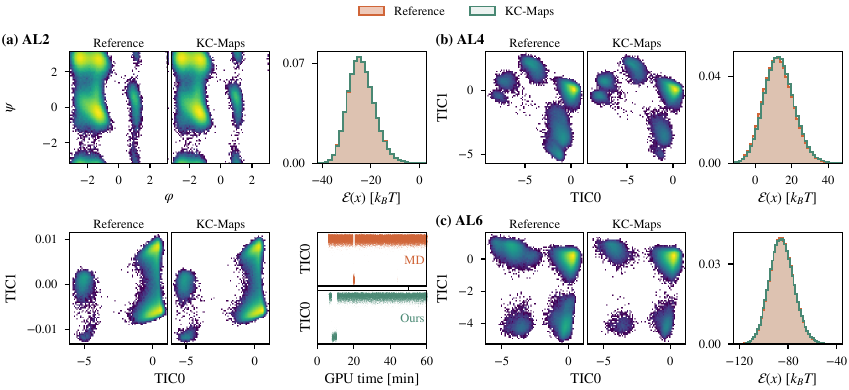}
  \caption{\textbf{Equilibrium sampling of alanine peptides.} Density panels show the reference MD (16 parallel chains $\times$ 2 $\mu$s per chain) beside \method (Ours). Potential energy histograms of \method against each system's MD reference.
  (a) AL2: Ramachandran and TICA densities, energy histogram, and TIC0 against simulated MD trajectory (top) and ours (bottom). (b) AL4: TICA densities and energy histogram. (c) AL6: TICA densities and energy histogram. For each molecule, samples generated from \method were run with varying compute budgets (1 GPU-hour for AL2; 22 GPU-hour for AL4; 40 GPU-hours for AL6).}
\label{fig:peptide_equilibrium}
\vspace{-0.5em}
\end{figure}

\subsection{Molecular Systems}
We evaluate our second-order parameterization on three molecular systems of increasing size: alanine dipeptide (AL2), tetrapeptide (AL4), and hexapeptide (AL6). We train the flow to model transitions of the joint state of positions and momenta under Langevin dynamics. Training uses \textit{uncorrelated} equilibrium configurations as initial states and the molecular force field, without requiring paired states from MD. We generate the trajectories by sequentially applying the learned transition kernel with step size $\tau$, and we compare the resulting equilibrium distribution against HFM \citep{ripken2026hfm} and molecular dynamics simulated using BAOAB scheme \citep{leimkuhler2013robust}.

For the peptide experiments, we train using $\cL_{\mathrm{KC}}$ with the detailed-balance loss $\cL_{\mathrm{DB}}$. We use $\cL_{\mathrm{KC}}$ since we found training with $\cL_{\mathrm{PF}}$ unstable, even with adaptive loss weighting \citep{geng_mean_2025,woo2026riemannian}. The LJ-55 results in \cref{tab:lj-results} also show substantially larger errors, especially with energy distribution, when training with $\cL_{\mathrm{PF}}$.

\begin{table}[!b]
  \centering
  \caption{Equilibrium sampling of alanine peptides per method with a fixed budget of one GPU-hour on single L40S, scored against each molecule's long MD. $\mathcal{W}_2$ metrics are computed between $10^4$ reference frames and $10^4$ frames drawn from the run after a 10\,\% burn-in, and report the mean $\pm$ std.\ over 3 draws. Best of HFM and \method\ in \textbf{bold}. 
  }
  \label{tab:mol-results}
  \small
  % Molecule in its own column left of the rules (rules span columns 2-6 only), as in Table 1 of arXiv:2606.29110.
  \setlength{\tabcolsep}{8pt}
  \renewcommand{\arraystretch}{1.15}
\begin{tabular}{@{}c@{\hspace{6pt}}clccc@{}}
    \cmidrule[\heavyrulewidth]{2-6}
    & & Method
      & $\mathcal{E}$-$\mathcal{W}_2$ $\downarrow$
      & $\mathcal{T}$-$\mathcal{W}_2$ $\downarrow$
      & TICA-$\mathcal{W}_2$ $\downarrow$ \\
    \cmidrule[\lightrulewidth]{2-6}

    % Bold ranks Ours and HFM only.
    % AL4 and AL6 share one visual scale; AL2 is enlarged slightly for legibility.
    \multirow{3}{*}{\includegraphics[height=1.0cm]{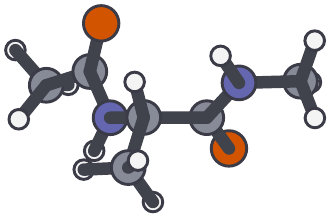}}
      & \multirow{3}{*}{
          \begin{tabular}{@{}c@{}}
            \textsf{\textbf{AL2}} \\ ($d=66$)
          \end{tabular}}
      & MD
      & $0.15 \pm 0.04$ & $0.20 \pm 0.00$ & $0.38 \pm 0.02$ \\
    \cmidrule{3-6}
    & & HFM
      & $0.15 \pm 0.01$ & $\mathbf{0.12 \pm 0.03}$ & $0.13 \pm 0.05$ \\
    & & \method
      & $\mathbf{0.13 \pm 0.02}$ & $0.21 \pm 0.03$ & $\mathbf{0.08 \pm 0.03}$ \\

    \cmidrule[\lightrulewidth]{2-6}

    \multirow{3}{*}{\includegraphics[height=1.35cm]{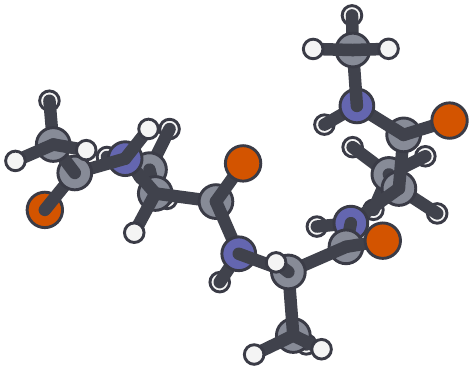}}
      & \multirow{3}{*}{
        \begin{tabular}{@{}c@{}}
          \textsf{\textbf{AL4}} \\ ($d=126$)
        \end{tabular}}
      & MD
      & $0.25 \pm 0.09$ & $0.58 \pm 0.02$ & $0.67 \pm 0.03$ \\
    \cmidrule{3-6}
    & & HFM
      & $\mathbf{0.21 \pm 0.04}$ & $\mathbf{0.74 \pm 0.02}$ & $0.52 \pm 0.02$ \\
    & & \method
      & $0.46 \pm 0.14$ & $1.08 \pm 0.03$ & $\mathbf{0.48 \pm 0.02}$ \\

    \cmidrule[\lightrulewidth]{2-6}

    \multirow{3}{*}{\includegraphics[height=1.35cm]{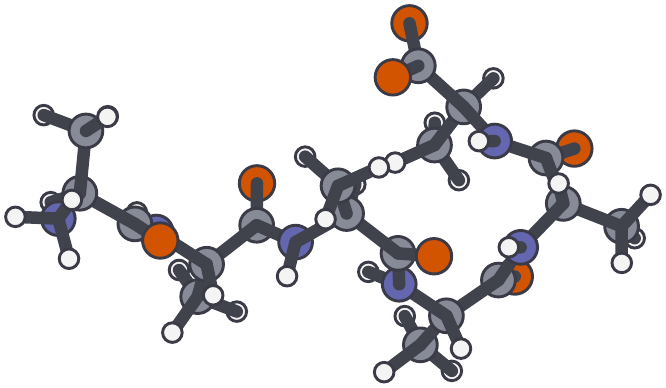}}
      & \multirow{3}{*}{
        \begin{tabular}{@{}c@{}}
          \textsf{\textbf{AL6}} \\ ($d=189$)
        \end{tabular}}
      & MD
      & $0.28 \pm 0.02$ & $1.17 \pm 0.01$ & $1.01 \pm 0.02$ \\
    \cmidrule{3-6}
    & & HFM
      & $0.90 \pm 0.05$ & $1.58 \pm 0.02$ & $0.95 \pm 0.04$ \\
    & & \method
      & $\mathbf{0.34 \pm 0.08}$ & $\mathbf{1.22 \pm 0.03}$ & $\mathbf{0.60 \pm 0.03}$ \\

    \cmidrule[\heavyrulewidth]{2-6}
  \end{tabular}
  \vspace{-1.4em}
\end{table}

\paragraph{Equilibrium} We compare generated samples with long MD references using 2-Wasserstein distances in energy ($\mathcal{E}$-$\mathcal{W}_2$), torsion angles ($\mathcal{T}$-$\mathcal{W}_2$), and first two component projections from time-lagged independent component analysis (TICA-$\mathcal{W}_2$), as done in \citep{tan_scalable_2025,tan2026amortized}.

% The metrics capture complementary aspects of the quality of the sampled distribution. For example, $\mathcal{E}$-$\mathcal{W}_2$ assesses the accuracy of local geometry while $\mathcal{T}$-$\mathcal{W}_2$ and TICA-$\mathcal{W}_2$ assess coverage of metastable states.

In \cref{tab:mol-results}, we evaluate all methods under an equal budget of one GPU hour. For each method, we vary the number of parallel chains and, for the learned models, the step size $\tau$, and report the configuration with the best average across all three metrics. Sampling and evaluation details are provided in \cref{app:setup-eval}.

At matched GPU-hour budgets, \method achieves the lowest TICA-$\mathcal{W}_2$ on all three systems while remaining competitive on energy and torsion metrics. We report performance across $\tau$ with and without the detailed-balance loss in \cref{fig:equilibrium-w2-vs-tau}. Longer rollouts of \method show broad mode coverage and closely match the reference energy distribution (\cref{fig:peptide_equilibrium}). For AL6, in particular, these rollouts reach modes that are absent in the training data, demonstrating exploration beyond the training conformations (see \cref{fig:al6-extrapolation-tica}).

\begin{figure}[t]
  \centering
  \includegraphics[width=0.96\linewidth]{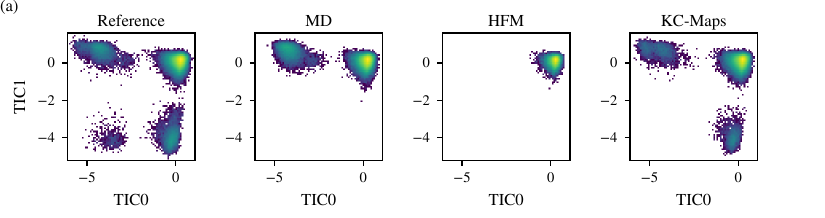}\\[2pt]
\includegraphics[width=0.96\linewidth]{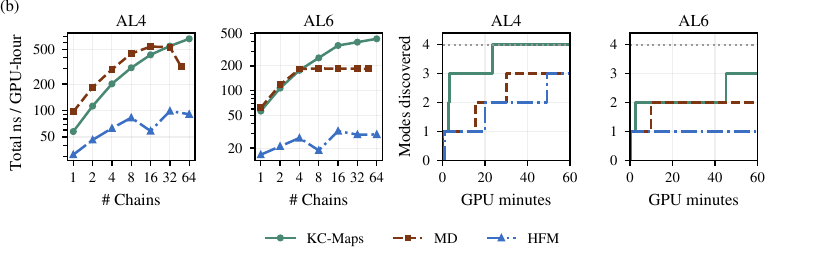} 
  \caption{\textbf{One GPU-hour on single L40S.} (a) TICA densities of the reference (16 chains; $dt=1$fs; 2\,$\mu$s each), MD (16 parallel chains; $dt=1$fs; one GPU-hour), HFM (16 chains; $\tau=5$\,fs; one GPU-hour) and \method\ (16 chains; $\tau=30$\,fs; one GPU-hour). \method\ consistently visits most if not all of the modes within the same GPU-hour. (b) Left: simulated aggregate time per GPU-hour against the number of parallel chains for \method\ ($\tau = 30$\,fs for AL4 and AL6), MD and HFM (7\,fs for AL4, 5\,fs for AL6). Right: metastable TICA modes discovered against GPU time. Dotted: modes of the reference MD.} 
  \label{fig:gpu-hour}
  \vspace{-1.5em}
\end{figure}

Furthermore, we evaluate exploration efficiency on AL4 and AL6 through aggregate simulated time and number of modes discovered within one GPU-hour. As shown in \cref{fig:gpu-hour}, \method's throughput increases with the number of chains on a single GPU, exceeding MD and HFM at larger chain counts. Within the same compute budget, \method also discovers the modes faster and samples more modes than either baseline. 

\begin{figure}[t]
  \centering
\includegraphics[width=\linewidth,height=1.95in]{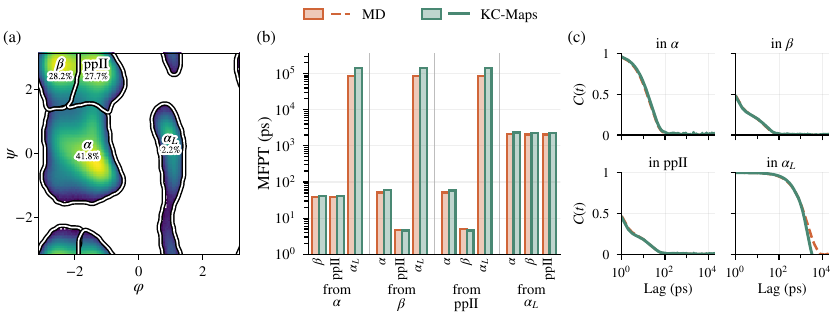}
  \caption{\textbf{AL2 kinetics.} (a) The four metastable modes ($\alpha$, $\beta$, ppII, $\alpha_L$) on the reference MD Ramachandran plot, with their populations. (b) Mean first-passage time between modes. (c) Autocorrelation $C(t)$ of mode occupancy. MD: 16 chains of 2\,$\mu$s each; \method: 16 chains at $\tau = 30$\,fs run for one GPU-hour on L40S. Both metrics are measured on trajectories with frames saved every 1\,ps.}
  \label{fig:al2-kinetics}
  \vspace{-1.5em}
\end{figure}

\paragraph{Kinetics} Here, we assess the kinetics of \method on AL2. We measure mean first-passage time (MFPT) between modes and the autocorrelation of mode occupancy. As shown in \cref{fig:al2-kinetics}, \method closely matches MD for MFPT between $\alpha$, $\beta$, and ppII, as well as their autocorrelation curves for these modes.
 
We observe the largest errors when involving $\alpha_L$: \method overestimates MFPT into this mode, and its autocorrelation decays faster at long lags. We believe these errors to stem from the limited support for that mode in the training data. Namely, $\alpha_L$ accounts for only $0.7\%$ of training configurations, compared with $2.2\%$ in the reference MD. We define the kinetic metrics in \cref{app:kinetic-metrics} and provide additional results on transition probabilities in \cref{fig:al2-kinetics-pij}.

\suppressfloats[t]
\vspace{0.1em}
\section{Related Work}
\label{sec:related_work}

Boltzmann generators \citep{noe_boltzmann_2019,klein_transferable_2024,tan_scalable_2025,tan2026amortized} directly sample uncorrelated equilibrium configurations without integrating MCMC. However, equilibrium distributions alone do not capture dynamical properties such as relaxation times and transition rates \citep{dixit2014inferring}, motivating surrogates that model time-coarsened MD transitions.

%\looseness=-1
Hamiltonian Flow Maps \citep{ripken2026hfm} learn deterministic updates from Hamiltonian equations using uncorrelated states and forces, and apply a separate thermostat update during simulation as done in the OBABO scheme \citep{bussi2007accurate}. Similarly, we leverage local physical information, but learn the stochastic transition kernel of Langevin dynamics.

%\looseness=-1
Consistency models \citep{song_consistency_2023,lu2025simplifying} and Flow Map methods \citep{boffi_flow_2024,geng_mean_2025,boffi_how_2025} learn finite-time maps of generative ODEs. Stochastic Flow Maps (SFMs) \citep{potaptchik_meta_2026,holderrieth_diamond_2026} extend this framework to conditional distributions, while subsequent works learn pathwise SDE solution maps conditioned on Brownian motion \citep{mccallum2026strongstochasticflowmaps,pan2026itomapsanystepsdes}. Our method falls within the SFM framework, directly parameterizing the transition kernel and enforcing its composition through the Kolmogorov--Chapman equation.

Lastly, similar works like Neural Stochastic Flows (NSF) \citep{kiyohara2025neuralstochasticflowssolverfree} also parameterize the SDE transition kernel with a conditional normalizing flows. However, their objective combines likelihood on \textit{observed} transition pairs with a variational upper bound on bidirectional KL regularization evaluated through a learned bridge model.
\method learns from unpaired states and the known SDE drift and diffusion. Unlike NSF, our parameterization preserves the prescribed infinitesimal dynamics, while the consistency objective uses transitions generated by composing the model itself. This allows us to learn finite-time transition distributions without observed future states.

\section{Conclusion}
\label{sec:conclusion}
Our work introduces \method, a trajectory-free approach to learning transition kernels of SDEs. Our method combines a parameterization that preserves the infinitesimal dynamics with a training objective based on the Kolmogorov--Chapman consistency, using \textit{uncorrelated} initial states and the known drift and diffusion. On molecular systems, we learn transitions of up to $100\,\mathrm{fs}$ and demonstrate improved equilibrium sampling efficiency compared with classical molecular dynamics, combining higher simulation throughput with broader conformational coverage at matched GPU-hour budgets. 

A limitation of our method is its dependence on training-data coverage. In particular, the method's kinetics suffers in performance in modes that are sparsely represented in the training set. Also, our method is trained separately for each molecular system and does not currently support transfer to unseen systems. Future work could address these limitations by amortizing across molecular systems, scaling to larger systems, and learning longer time horizons more accurately. We believe training an amortized model across systems may help improve performance in sparsely covered modes through shared structural and dynamical information across systems. Finally, our Lennard-Jones results motivate learning a map by leveraging higher-temperature data of molecular systems, without requiring equilibrium samples at the target temperature.

\subsection*{Acknowledgments}
M.H. acknowledges the support of the Natural Sciences and Engineering Research Council of Canada (NSERC), [funding reference number RGPIN-2026-05517]. M.S. was supported by the IVADO 2025 Postdoctoral Research Funding Program. This project was undertaken thanks to funding from IVADO and the Canada First Research Excellence Fund.

\subsection*{Reproducibility statement}
We provide code at \href{https://github.com/majhas/kc-maps}{\texttt{https://github.com/majhas/kc-maps}}. We also fully detail the training implementation and evaluation setup with details on hyperparameters in \cref{app:implementation-details}. Proofs of theoretical results are in \cref{app:proofs} and derivations in \cref{app:derivations}.

%%%%%%%%%%%%%%%%%%%%%%%%%%%%%%%%%%%%%%%%%%%%%%%%%%%%%%%%%%%%

\clearpage
\bibliographystyle{iclr2027_conference}
\bibliography{references, references_zotero}

@book{karlin1981second,
  title={A second course in stochastic processes},
  author={Karlin, Samuel and Taylor, Howard},
  year={1981},
  publisher={Elsevier}
}

@article{woo2026riemannian,
  title={Riemannian MeanFlow},
  author={Woo, Dongyeop and Skreta, Marta and Park, Seonghyun and Neklyudov, Kirill and Ahn, Sungsoo},
  journal={arXiv preprint arXiv:2602.07744},
  year={2026}
}

@software{jax2018github,
  author = {James Bradbury and Roy Frostig and Peter Hawkins and Matthew James Johnson and Yash Katariya and Chris Leary and Dougal Maclaurin and George Necula and Adam Paszke and Jake Vander{P}las and Skye Wanderman-{M}ilne and Qiao Zhang},
  title = {{JAX}: composable transformations of {P}ython+{N}um{P}y programs},
  url = {http://github.com/jax-ml/jax},
  version = {0.3.13},
  year = {2018},
}

@article{bussi2007accurate,
  title={Accurate sampling using Langevin dynamics},
  author={Bussi, Giovanni and Parrinello, Michele},
  journal={Physical Review E},
  volume={75},
  number={5},
  pages={056707},
  year={2007},
  publisher={APS},
  doi={10.1103/PhysRevE.75.056707}
}

@misc{lu2025simplifying,
      title={Simplifying, Stabilizing and Scaling Continuous-Time Consistency Models}, 
      author={Cheng Lu and Yang Song},
      year={2025},
      eprint={2410.11081},
      archivePrefix={arXiv},
      primaryClass={cs.LG},
      url={https://arxiv.org/abs/2410.11081}, 
}

@article{kolmogoroff1931analytischen,
  title={{\"U}ber die analytischen Methoden in der Wahrscheinlichkeitsrechnung},
  author={Kolmogoroff, Andrei},
  journal={Mathematische Annalen},
  volume={104},
  number={1},
  pages={415--458},
  year={1931},
  publisher={Springer}
}

@article{feller1937theorie,
  title={Zur Theorie der stochastischen Prozesse: Existenz-und Eindeutigkeitss{\"a}tze},
  author={Feller, Willy},
  journal={Mathematische Annalen},
  volume={113},
  number={1},
  pages={113--160},
  year={1937},
  publisher={Springer}
}

@article{dixit2014inferring,
  title={Inferring Microscopic Kinetic Rates from Stationary State Distributions},
  author={Dixit, Purushottam D. and Dill, Ken A.},
  journal={Journal of Chemical Theory and Computation},
  volume={10},
  number={8},
  pages={3002--3005},
  year={2014},
  doi={10.1021/ct5001389}
}

@article{roberts1996exponential,
  author    = {Roberts, Gareth O. and Tweedie, Richard L.},
  title     = {Exponential convergence of {L}angevin distributions and their discrete approximations},
  journal   = {Bernoulli},
  volume    = {2},
  number    = {4},
  pages     = {341--363},
  year      = {1996},
  publisher = {Bernoulli Society for Mathematical Statistics and Probability}
}

@article{bussi2007canonical,
  title={Canonical sampling through velocity rescaling},
  author={Bussi, Giovanni and Donadio, Davide and Parrinello, Michele},
  journal={The Journal of Chemical Physics},
  volume={126},
  number={1},
  pages={014101},
  year={2007}
}

@inproceedings{kohler2021smooth,
  title={Smooth Normalizing Flows},
  author={K{\"o}hler, Jonas and Kr{\"a}mer, Andreas and No{\'e}, Frank},
  booktitle={Advances in Neural Information Processing Systems},
  volume={34},
  year={2021}
}

@inproceedings{kohler2020equivariant,
  title={Equivariant Flows: Exact Likelihood Generative Learning for Symmetric Densities},
  author={K{\"o}hler, Jonas and Klein, Leon and No{\'e}, Frank},
  booktitle={International Conference on Machine Learning},
  pages={5361--5370},
  year={2020}
}

@article{hoffman2014nuts,
  title={The No-{U}-Turn Sampler: Adaptively Setting Path Lengths in {H}amiltonian {M}onte {C}arlo},
  author={Hoffman, Matthew D. and Gelman, Andrew},
  journal={Journal of Machine Learning Research},
  volume={15},
  number={1},
  pages={1593--1623},
  year={2014}
}

@article{eastman2017openmm,
  title={{OpenMM} 7: Rapid Development of High Performance Algorithms for Molecular Dynamics},
  author={Eastman, Peter and Swails, Jason and Chodera, John D. and McGibbon, Robert T. and Zhao, Yutong and Beauchamp, Kyle A. and Wang, Lee-Ping and Simmonett, Andrew C. and Harrigan, Matthew P. and Stern, Chaya D. and others},
  journal={PLoS Computational Biology},
  volume={13},
  number={7},
  pages={e1005659},
  year={2017}
}

@article{lindorff2010ff99sbildn,
  title={Improved Side-Chain Torsion Potentials for the {A}mber ff99{SB} Protein Force Field},
  author={Lindorff-Larsen, Kresten and Piana, Stefano and Palmo, Kim and Maragakis, Paul and Klepeis, John L. and Dror, Ron O. and Shaw, David E.},
  journal={Proteins: Structure, Function, and Bioinformatics},
  volume={78},
  number={8},
  pages={1950--1958},
  year={2010}
}

@article{maier2015ff14sb,
  title={ff14{SB}: Improving the Accuracy of Protein Side Chain and Backbone Parameters from ff99{SB}},
  author={Maier, James A. and Martinez, Carmenza and Kasavajhala, Koushik and Wickstrom, Lauren and Hauser, Kevin E. and Simmerling, Carlos},
  journal={Journal of Chemical Theory and Computation},
  volume={11},
  number={8},
  pages={3696--3713},
  year={2015}
}

@article{onufriev2004obc,
  title={Exploring Protein Native States and Large-Scale Conformational Changes with a Modified Generalized {B}orn Model},
  author={Onufriev, Alexey and Bashford, Donald and Case, David A.},
  journal={Proteins: Structure, Function, and Bioinformatics},
  volume={55},
  number={2},
  pages={383--394},
  year={2004}
}

@article{perez2013tica,
  title={Identification of Slow Molecular Order Parameters for {M}arkov Model Construction},
  author={P{\'e}rez-Hern{\'a}ndez, Guillermo and Paul, Fabian and Giorgino, Toni and De Fabritiis, Gianni and No{\'e}, Frank},
  journal={The Journal of Chemical Physics},
  volume={139},
  number={1},
  pages={015102},
  year={2013}
}

@inproceedings{loshchilov2019adamw,
  title={Decoupled Weight Decay Regularization},
  author={Loshchilov, Ilya and Hutter, Frank},
  booktitle={International Conference on Learning Representations},
  year={2019}
}

@article{hoffmann2021deeptime,
  title={Deeptime: a {P}ython Library for Machine Learning Dynamical Models from Time Series Data},
  author={Hoffmann, Moritz and Scherer, Martin Konrad and Hempel, Tim and Mardt, Andreas and de Silva, Brian
    and Husic, Brooke Elena and Klus, Stefan and Wu, Hao and Kutz, J Nathan and Brunton, Steven and No{\'e}, Frank},
  journal={Machine Learning: Science and Technology},
  volume={3},
  number={1},
  pages={015009},
  year={2022}
}

@inproceedings{tucker2019doubly,
  title     = {Doubly Reparameterized Gradient Estimators for {Monte Carlo} Objectives},
  author    = {Tucker, George and Lawson, Dieterich and Gu, Shixiang and Maddison, Chris J.},
  booktitle = {International Conference on Learning Representations},
  year      = {2019}
}

@inproceedings{peebles2023scalable,
  title={Scalable Diffusion Models with Transformers},
  author={Peebles, William and Xie, Saining},
  booktitle={Proceedings of the IEEE/CVF International Conference on Computer Vision (ICCV)},
  pages={4195--4205},
  year={2023}
}

@article{grossfield2018best,
  title={Best Practices for Quantification of Uncertainty and Sampling Quality in Molecular Simulations [Article v1.0]},
  author={Grossfield, Alan and Patrone, Paul N. and Roe, Daniel R. and Schultz, Andrew J. and Siderius, Daniel W. and Zuckerman, Daniel M.},
  journal={Living Journal of Computational Molecular Science},
  volume={1},
  number={1},
  pages={5067},
  year={2018},
  doi={10.33011/livecoms.1.1.5067}
}

@article{Hukushima_1996,
   title={Exchange Monte Carlo Method and Application to Spin Glass Simulations},
   volume={65},
   ISSN={1347-4073},
   url={http://dx.doi.org/10.1143/JPSJ.65.1604},
   DOI={10.1143/jpsj.65.1604},
   number={6},
   journal={Journal of the Physical Society of Japan},
   publisher={Physical Society of Japan},
   author={Hukushima, Koji and Nemoto, Koji},
   year={1996},
   month=June, pages={1604–1608} }

@misc{kaba2023equivariance,
      title={Equivariance with Learned Canonicalization Functions},
      author={Sékou-Oumar Kaba and Arnab Kumar Mondal and Yan Zhang and Yoshua Bengio and Siamak Ravanbakhsh},
      year={2023},
      eprint={2211.06489},
      archivePrefix={arXiv},
      primaryClass={cs.LG},
      url={https://arxiv.org/abs/2211.06489},
}

@article{hastings1970,
      title={{Monte Carlo} sampling methods using {Markov} chains and their applications},
      author={Hastings, W. K.},
      journal={Biometrika},
      volume={57},
      number={1},
      pages={97--109},
      year={1970},
      doi={10.1093/biomet/57.1.97},
      url={https://doi.org/10.1093/biomet/57.1.97},
}

@misc{tan2026amortized,
      title={Amortized Sampling with Transferable Normalizing Flows}, 
      author={Charlie B. Tan and Majdi Hassan and Leon Klein and Saifuddin Syed and Dominique Beaini and Michael M. Bronstein and Alexander Tong and Kirill Neklyudov},
      year={2026},
      eprint={2508.18175},
      archivePrefix={arXiv},
      primaryClass={cs.LG},
      url={https://arxiv.org/abs/2508.18175}, 
}

@misc{akhound2025pita,
      title={Progressive Inference-Time Annealing of Diffusion Models for Sampling from Boltzmann Densities}, 
      author={Tara Akhound-Sadegh and Jungyoon Lee and Avishek Joey Bose and Valentin De Bortoli and Arnaud Doucet and Michael M. Bronstein and Dominique Beaini and Siamak Ravanbakhsh and Kirill Neklyudov and Alexander Tong},
      year={2025},
      eprint={2506.16471},
      archivePrefix={arXiv},
      primaryClass={cs.LG},
      url={https://arxiv.org/abs/2506.16471}, 
}

@misc{potaptchik2025tilt,
      title={Tilt Matching for Scalable Sampling and Fine-Tuning}, 
      author={Peter Potaptchik and Cheuk-Kit Lee and Michael S. Albergo},
      year={2025},
      eprint={2512.21829},
      archivePrefix={arXiv},
      primaryClass={stat.ML},
      url={https://arxiv.org/abs/2512.21829}, 
}

@misc{ripken2026hfm,
      title={Learning Hamiltonian Flow Maps: Mean Flow Consistency for Large-Timestep Molecular Dynamics}, 
      author={Winfried Ripken and Michael Plainer and Gregor Lied and Thorben Frank and Oliver T. Unke and Stefan Chmiela and Frank Noé and Klaus-Robert Müller},
      year={2026},
      eprint={2601.22123},
      archivePrefix={arXiv},
      primaryClass={cs.LG},
      url={https://arxiv.org/abs/2601.22123}, 
}

@misc{bigi2026flashmd,
      title={FlashMD: long-stride, universal prediction of molecular dynamics}, 
      author={Filippo Bigi and Sanggyu Chong and Agustinus Kristiadi and Michele Ceriotti},
      year={2026},
      eprint={2505.19350},
      archivePrefix={arXiv},
      primaryClass={physics.chem-ph},
      url={https://arxiv.org/abs/2505.19350}, 
}

@misc{albergo_building_2023,
	title = {Building {Normalizing} {Flows} with {Stochastic} {Interpolants}},
	url = {http://arxiv.org/abs/2209.15571},
	doi = {10.48550/arXiv.2209.15571},
	urldate = {2026-04-23},
	publisher = {arXiv},
	author = {Albergo, Michael S. and Vanden-Eijnden, Eric},
	month = mar,
	year = {2023},
	note = {arXiv:2209.15571 [cs]},
}

@misc{geng_mean_2025,
	title = {Mean {Flows} for {One}-step {Generative} {Modeling}},
	url = {http://arxiv.org/abs/2505.13447},
	doi = {10.48550/arXiv.2505.13447},
	urldate = {2026-04-23},
	publisher = {arXiv},
	author = {Geng, Zhengyang and Deng, Mingyang and Bai, Xingjian and Kolter, J. Zico and He, Kaiming},
	month = may,
	year = {2025},
	note = {arXiv:2505.13447 [cs]},
}

@misc{potaptchik_meta_2026,
	title = {Meta {Flow} {Maps} enable scalable reward alignment},
	url = {http://arxiv.org/abs/2601.14430},
	doi = {10.48550/arXiv.2601.14430},
	urldate = {2026-04-23},
	publisher = {arXiv},
	author = {Potaptchik, Peter and Saravanan, Adhi and Mammadov, Abbas and Prat, Alvaro and Albergo, Michael S. and Teh, Yee Whye},
	month = jan,
	year = {2026},
	note = {arXiv:2601.14430 [stat]},
}

@misc{boffi_how_2025,
	title = {How to build a consistency model: {Learning} flow maps via self-distillation},
	shorttitle = {How to build a consistency model},
	url = {http://arxiv.org/abs/2505.18825},
	doi = {10.48550/arXiv.2505.18825},
	urldate = {2026-04-23},
	publisher = {arXiv},
	author = {Boffi, Nicholas M. and Albergo, Michael S. and Vanden-Eijnden, Eric},
	month = oct,
	year = {2025},
	note = {arXiv:2505.18825 [cs]},
}

@misc{holderrieth_diamond_2026,
	title = {Diamond {Maps}: {Efficient} {Reward} {Alignment} via {Stochastic} {Flow} {Maps}},
	shorttitle = {Diamond {Maps}},
	url = {http://arxiv.org/abs/2602.05993},
	doi = {10.48550/arXiv.2602.05993},
	urldate = {2026-04-23},
	publisher = {arXiv},
	author = {Holderrieth, Peter and Chen, Douglas and Eyring, Luca and Shah, Ishin and Anantharaman, Giri and He, Yutong and Akata, Zeynep and Jaakkola, Tommi and Boffi, Nicholas Matthew and Simchowitz, Max},
	month = apr,
	year = {2026},
	note = {arXiv:2602.05993 [cs]
version: 2},
}

@misc{song_consistency_2023,
	title = {Consistency {Models}},
	url = {http://arxiv.org/abs/2303.01469},
	doi = {10.48550/arXiv.2303.01469},
	urldate = {2026-04-23},
	publisher = {arXiv},
	author = {Song, Yang and Dhariwal, Prafulla and Chen, Mark and Sutskever, Ilya},
	month = may,
	year = {2023},
	note = {arXiv:2303.01469 [cs]},
}

@misc{tan_scalable_2025,
	title = {Scalable {Equilibrium} {Sampling} with {Sequential} {Boltzmann} {Generators}},
	url = {http://arxiv.org/abs/2502.18462},
	doi = {10.48550/arXiv.2502.18462},
	urldate = {2025-02-26},
	publisher = {arXiv},
	author = {Tan, Charlie B. and Bose, Avishek Joey and Lin, Chen and Klein, Leon and Bronstein, Michael M. and Tong, Alexander},
	month = feb,
	year = {2025},
	note = {arXiv:2502.18462 [cs]},
}

@misc{vaswani_attention_2023,
	title = {Attention {Is} {All} {You} {Need}},
	url = {http://arxiv.org/abs/1706.03762},
	doi = {10.48550/arXiv.1706.03762},
	urldate = {2025-01-28},
	publisher = {arXiv},
	author = {Vaswani, Ashish and Shazeer, Noam and Parmar, Niki and Uszkoreit, Jakob and Jones, Llion and Gomez, Aidan N. and Kaiser, Lukasz and Polosukhin, Illia},
	month = aug,
	year = {2023},
	note = {arXiv:1706.03762 [cs]},
}

@misc{noe_boltzmann_2019,
	title = {Boltzmann {Generators} -- {Sampling} {Equilibrium} {States} of {Many}-{Body} {Systems} with {Deep} {Learning}},
	url = {http://arxiv.org/abs/1812.01729},
	doi = {10.48550/arXiv.1812.01729},
	urldate = {2025-01-28},
	publisher = {arXiv},
	author = {Noé, Frank and Olsson, Simon and Köhler, Jonas and Wu, Hao},
	month = jul,
	year = {2019},
	note = {arXiv:1812.01729 [stat]},
}

@misc{song_score-based_2021,
	title = {Score-{Based} {Generative} {Modeling} through {Stochastic} {Differential} {Equations}},
	url = {http://arxiv.org/abs/2011.13456},
	doi = {10.48550/arXiv.2011.13456},
	urldate = {2025-01-10},
	publisher = {arXiv},
	author = {Song, Yang and Sohl-Dickstein, Jascha and Kingma, Diederik P. and Kumar, Abhishek and Ermon, Stefano and Poole, Ben},
	month = feb,
	year = {2021},
	note = {arXiv:2011.13456 [cs]},
}

@misc{moral_sequential_2002,
	title = {Sequential monte carlo samplers},
	url = {http://arxiv.org/abs/cond-mat/0212648},
	doi = {10.48550/arXiv.cond-mat/0212648},
	urldate = {2024-12-20},
	publisher = {arXiv},
	author = {Moral, Pierre Del and Doucet, Arnaud},
	month = dec,
	year = {2002},
	note = {arXiv:cond-mat/0212648},
}

@misc{klein_transferable_2024,
	title = {Transferable {Boltzmann} {Generators}},
	url = {http://arxiv.org/abs/2406.14426},
	doi = {10.48550/arXiv.2406.14426},
	urldate = {2024-06-21},
	author = {Klein, Leon and Noé, Frank},
	month = jun,
	year = {2024},
	note = {arXiv:2406.14426 [physics, stat]},
}

@misc{boffi_flow_2024,
	title = {Flow {Map} {Matching}},
	url = {http://arxiv.org/abs/2406.07507},
	doi = {10.48550/arXiv.2406.07507},
	urldate = {2024-06-14},
	author = {Boffi, Nicholas M. and Albergo, Michael S. and Vanden-Eijnden, Eric},
	month = jun,
	year = {2024},
	note = {arXiv:2406.07507 [cs, math]},
}

@misc{akhound-sadegh_iterated_2024,
	title = {Iterated {Denoising} {Energy} {Matching} for {Sampling} from {Boltzmann} {Densities}},
	url = {http://arxiv.org/abs/2402.06121},
	doi = {10.48550/arXiv.2402.06121},
	urldate = {2024-02-12},
	author = {Akhound-Sadegh, Tara and Rector-Brooks, Jarrid and Bose, Avishek Joey and Mittal, Sarthak and Lemos, Pablo and Liu, Cheng-Hao and Sendera, Marcin and Ravanbakhsh, Siamak and Gidel, Gauthier and Bengio, Yoshua and Malkin, Nikolay and Tong, Alexander},
	month = feb,
	year = {2024},
	note = {arXiv:2402.06121 [cs, stat]},
}

@misc{klein_equivariant_2023,
	title = {Equivariant flow matching},
	url = {http://arxiv.org/abs/2306.15030},
	doi = {10.48550/arXiv.2306.15030},
	urldate = {2023-06-29},
	author = {Klein, Leon and Krämer, Andreas and Noé, Frank},
	month = jun,
	year = {2023},
	note = {arXiv:2306.15030 [physics, stat]},
}

@misc{schreiner_implicit_2023,
	title = {Implicit {Transfer} {Operator} {Learning}: {Multiple} {Time}-{Resolution} {Surrogates} for {Molecular} {Dynamics}},
	shorttitle = {Implicit {Transfer} {Operator} {Learning}},
	url = {http://arxiv.org/abs/2305.18046},
	urldate = {2023-05-30},
	publisher = {arXiv},
	author = {Schreiner, Mathias and Winther, Ole and Olsson, Simon},
	month = may,
	year = {2023},
	note = {arXiv:2305.18046 [physics, stat]},
}

@misc{klein_timewarp_2023,
	title = {Timewarp: {Transferable} {Acceleration} of {Molecular} {Dynamics} by {Learning} {Time}-{Coarsened} {Dynamics}},
	shorttitle = {Timewarp},
	url = {http://arxiv.org/abs/2302.01170},
	doi = {10.48550/arXiv.2302.01170},
	urldate = {2023-03-27},
	publisher = {arXiv},
	author = {Klein, Leon and Foong, Andrew Y. K. and Fjelde, Tor Erlend and Mlodozeniec, Bruno and Brockschmidt, Marc and Nowozin, Sebastian and Noé, Frank and Tomioka, Ryota},
	month = feb,
	year = {2023},
	note = {arXiv:2302.01170 [cond-mat, physics:physics, stat]},
}

@misc{lipman_flow_2023,
	title = {Flow {Matching} for {Generative} {Modeling}},
	url = {http://arxiv.org/abs/2210.02747},
	doi = {10.48550/arXiv.2210.02747},
	urldate = {2023-03-26},
	publisher = {arXiv},
	author = {Lipman, Yaron and Chen, Ricky T. Q. and Ben-Hamu, Heli and Nickel, Maximilian and Le, Matt},
	month = feb,
	year = {2023},
	note = {arXiv:2210.02747 [cs, stat]},
}

@misc{ho_denoising_2020,
	title = {Denoising {Diffusion} {Probabilistic} {Models}},
	url = {http://arxiv.org/abs/2006.11239},
	doi = {10.48550/arXiv.2006.11239},
	urldate = {2022-10-11},
	publisher = {arXiv},
	author = {Ho, Jonathan and Jain, Ajay and Abbeel, Pieter},
	month = dec,
	year = {2020},
	note = {arXiv:2006.11239 [cs, stat]},
}

@article{jumper_highly_2021,
	title = {Highly accurate protein structure prediction with {AlphaFold}},
	issn = {0028-0836, 1476-4687},
	url = {https://www.nature.com/articles/s41586-021-03819-2},
	doi = {10.1038/s41586-021-03819-2},
	language = {en},
	urldate = {2021-08-18},
	journal = {Nature},
	author = {Jumper, John and Evans, Richard and Pritzel, Alexander and Green, Tim and Figurnov, Michael and Ronneberger, Olaf and Tunyasuvunakool, Kathryn and Bates, Russ and Žídek, Augustin and Potapenko, Anna and Bridgland, Alex and Meyer, Clemens and Kohl, Simon A. A. and Ballard, Andrew J. and Cowie, Andrew and Romera-Paredes, Bernardino and Nikolov, Stanislav and Jain, Rishub and Adler, Jonas and Back, Trevor and Petersen, Stig and Reiman, David and Clancy, Ellen and Zielinski, Michal and Steinegger, Martin and Pacholska, Michalina and Berghammer, Tamas and Bodenstein, Sebastian and Silver, David and Vinyals, Oriol and Senior, Andrew W. and Kavukcuoglu, Koray and Kohli, Pushmeet and Hassabis, Demis},
	month = jul,
	year = {2021},
}

@article{hollingsworth_molecular_2018,
	title = {Molecular {Dynamics} {Simulation} for {All}},
	author = {Hollingsworth, Scott A. and Dror, Ron O.},
	journal = {Neuron},
	year = {2018},
	month = sep,
	volume = {99},
	number = {6},
	pages = {1129--1143},
	doi = {10.1016/j.neuron.2018.08.011},
	url = {https://doi.org/10.1016/j.neuron.2018.08.011},
	note = {PMID: 30236283; PMCID: PMC6209097},
}

@book{leimkuhler_molecular_2015,
	title = {Molecular {Dynamics}: With {Deterministic} and {Stochastic} {Numerical} {Methods}},
	author = {Leimkuhler, Benedict and Matthews, Charles},
	year = {2015},
	publisher = {Springer International Publishing},
	series = {Interdisciplinary Applied Mathematics},
	volume = {39},
	doi = {10.1007/978-3-319-16375-8},
	url = {https://doi.org/10.1007/978-3-319-16375-8},
	isbn = {978-3-319-16375-8},
}

@inproceedings{Kloeden1992NumericalSO,
  title={Numerical Solution of Stochastic Differential Equations},
  author={Peter E. Kloeden and Eckhard Platen},
  year={1992},
  url={https://api.semanticscholar.org/CorpusID:116903781}
}

@article{leimkuhler2013robust,
  title={Robust and efficient configurational molecular sampling via Langevin dynamics},
  author={Leimkuhler, Benedict and Matthews, Charles},
  journal={The Journal of chemical physics},
  volume={138},
  number={17},
  year={2013},
  publisher={AIP Publishing}
}

@inproceedings{Stroock1979MultidimensionalDP,
  title={Multidimensional Diffusion Processes},
  author={Daniel W. Stroock and Srinivasa Varadhan},
  year={1979},
  url={https://api.semanticscholar.org/CorpusID:118022077}
}

@misc{kiyohara2025neuralstochasticflowssolverfree,
  title={Neural Stochastic Flows: Solver-Free Modelling and Inference for SDE Solutions},
  author={Naoki Kiyohara and Edward Johns and Yingzhen Li},
  year={2025},
  eprint={2510.25769},
  archivePrefix={arXiv},
  primaryClass={cs.LG},
  url={https://arxiv.org/abs/2510.25769},
}

@misc{mccallum2026strongstochasticflowmaps,
  title={Strong Stochastic Flow Maps},
  author={Sam McCallum and Zander W. Blasingame and Timothy Herschell and Niklas Rindtorff and Alexander Tong and James Foster},
  year={2026},
  eprint={2606.01086},
  archivePrefix={arXiv},
  primaryClass={cs.LG},
  url={https://arxiv.org/abs/2606.01086},
}

@misc{pan2026itomapsanystepsdes,
  title={It\^o maps for any-step SDEs},
  author={Zhengkai Pan and Peter Potaptchik and Wenxi Yao and Michael S. Albergo and Jakiw Pidstrigach},
  year={2026},
  eprint={2606.11156},
  archivePrefix={arXiv},
  primaryClass={stat.ML},
  url={https://arxiv.org/abs/2606.11156},
}

%%%%%%%%%%%%%%%%%%%%%%%%%%%%%%%%%%%%%%%%%%%%%%%%%%%%%%%%%%%%

\newpage
\appendix
\etocdepthtag.toc{appendix}
\etocsettagdepth{mainbody}{none}
\etocsettagdepth{appendix}{subsection}
\etocsettocstyle{\begin{center}\large\bfseries \methodlong: Trajectory-Free Learning for Simulating Molecular Dynamics\\[2pt]
(Supplementary Material)\end{center}\vspace{0.5em}}{}
\tableofcontents
\clearpage
% Keep natural vertical spacing throughout the appendix.
\raggedbottom

\renewcommand{\thefigure}{A\arabic{figure}}  % For figures
\renewcommand{\thetable}{A\arabic{table}}    % For tables
\setcounter{figure}{0}  % Reset figure counter
\setcounter{table}{0}   % Reset table 

\section{Proofs}
\label{app:proofs}

\subsection{Identification of the transition kernel}
\label{app:kernel-identification}

Here we show the proof of a classical result for the identification of the kernel \citep{Stroock1979MultidimensionalDP}.
\begin{proof}[Proof of \cref{prop:kernel-identification}]
    Fix $s<t$ and an initial state $x$.
    Let $\tau=t-s$, set $t_j=s+j\tau/n$, and define the Markov chain
    \begin{align*}
        X^{(n)}_s &= x, &
        X^{(n)}_{t_{j+1}}\mid X^{(n)}_{t_j}
        &\sim p_{t_j,t_{j+1}}(\cdot\mid X^{(n)}_{t_j}),
        \qquad j=0,\ldots,n-1.
    \end{align*}
    By repeated application of Kolmogorov--Chapman, $X^{(n)}_t$ has density $p_{s,t}(\cdot\mid x)$ for every $n$.
    
    The boundary conditions in \cref{eq:boundary-local,eq:boundary-moment_1,eq:boundary-moment_2} are exactly the assumption of the diffusion approximation \citep{Stroock1979MultidimensionalDP}. Its remaining hypothesis, well-posedness of the martingale problem for $(v, g g^\top)$, follows from the assumptions on $v$ and $g$ in \cref{prop:kernel-identification}: they give a pathwise unique, non-explosive solution of \cref{eq:sde}, hence uniqueness in law, which is equivalent to well-posedness \citep{Stroock1979MultidimensionalDP}. Therefore, this gives weak convergence of the linearly interpolated chains $(X_r^{(n)})_{r\in[s,t]}$ to $(X_r)_{r\in[s,t]}$, where $(X_r)_{r\in[s,t]}$ solves the SDE in \cref{eq:sde} with initial state $X_s=x$.
    Note that to apply the diffusion approximation to time-dependent systems, we can simply apply the result to $(r,X_r)$, treating time as part of the state. We first establish convergence within some bounded region where the boundary condition errors vanish uniformly as $n\to\infty$.
    Second, since the SDE has continuous trajectories defined through $[s, t]$, the probability of leaving the bounded region tends to zero as the region extends to the full state space.
    Therefore, the localization argument gives convergence over the full interval.
    
    Weak convergence thus gives, for every bounded continuous $f$,
    \begin{align*}
        \int dy\,f(y)\,p_{s,t}(y\mid x)
        &=\E \left[f\left(X^{(n)}_t\right)\right]
        \longrightarrow \E[f(X_t)\mid X_s=x].
    \end{align*}
    The left-hand side is independent of $n$, so $p_{s,t}(\cdot\mid x)$ is the SDE's transition density, up to equality almost everywhere. Since $s$, $t$, and $x$ were arbitrary, this identifies the entire transition kernel.
    
    Conversely, the SDE's transition kernel satisfies the boundary conditions \cref{eq:boundary-local,eq:boundary-moment_1,eq:boundary-moment_2}, and its Markov property gives Kolmogorov--Chapman consistency. For the proof, we refer the reader to a textbook \citep{karlin1981second}.
\end{proof}

\subsection{Global minimizers of the KC divergence}
\label{app:ck-minimizer}

\begin{proof}[Proof of global minimizer]
    Here, $\cL_{\mathrm{KC}}$ denotes the KL divergence in \cref{eq:ck-kl}, averaged over the sampled source states and time triples.
    KL divergence is nonnegative, so $\cL_{\mathrm{KC}}(\theta)\geq 0$.
    The SDE's transition kernel $p$ satisfies Kolmogorov--Chapman consistency and belongs to the model class.
    When $q_{s,t}(y \mid x; \theta) = \int dx'\; q_{r,t}(y \mid x' ;\theta)q_{s,r}(x'\mid x;\theta)$, the parameterized kernel achieves $\cL_{\mathrm{KC}}=0$, giving $\min_\theta\cL_{\mathrm{KC}}(\theta)=0$.

    Now suppose $\cL_{\mathrm{KC}}(\theta)=0$.
    The nonnegative KL integrand in \cref{eq:ck-kl} must vanish almost everywhere under the joint sampling distribution.
    Continuity and full support imply that it vanishes for every initial state and time triple $s<r<t$.

    Since the target and learned transition kernels match, zero KL divergence gives
    \begin{align}
        q_{s,t}(y\mid x;\theta)
        &=
        \int dx'\,q_{r,t}(y\mid x';\theta)
             q_{s,r}(x'\mid x;\theta)
    \end{align}
    for every $x$ and $s<r<t$, with equality almost everywhere in $y$.
    Thus, the learned kernel satisfies Kolmogorov--Chapman consistency.
    By assumption, it also satisfies the boundary conditions with limits locally uniform in $(s,x)$.
    Applying \cref{prop:kernel-identification} yields
    \begin{align}
        q_{s,t}(\cdot\mid x;\theta)
        &=p_{s,t}(\cdot\mid x),
        \qquad \forall x,\;s<t.
    \end{align}
\end{proof}

\subsection{Proof of \cref{lemma:gate-orders}}
\label{app:gate-orders}

We use the following notation:
\begin{itemize}
    \item $u \coloneqq \sqrt{\tau}\,g z$ with $z\sim\Normal(0,I)$;
    \item $\kappa_n(\tau) \coloneqq \frac{(\omega\tau)^n}{1+(\omega\tau)^n}$;
    \item $\bar a := \max_k \sup \|a_k\|_\infty$ and $\bar b := \max_k \sup \|b_k\|$ denote the uniform bounds on the conditioners (over $x_1^{(k)}$ and $x$);
    \item $C$ denote constants depending only on $K$, $\omega$, $\bar a$, $\bar b$.
\end{itemize}

\begin{lemma}[Gated layers are small perturbations of the identity]
\label{lemma:gate-perturbation}
For all $\tau\in(0,1]$ and all $u$,
\begin{align}
    \|\Psi(u) - u\| \le C\left(\tau^{n_a}\|u\| + \tau^{n_b}\right)\,.\label{eq:flow-perturbation}
\end{align}
\end{lemma}
\begin{proof}
Consider one layer \cref{eq:time_affine_layer} with input $x=(x_1,x_2)$
\begin{align}
    \|\Psi_k(x)-x\| = \big\|x_2\odot\big(e^{\kappa_{n_a}(\tau) a_k}-1\big) + \kappa_{n_b}(\tau)\, b_k\big\|
    \leq \|x_2\|\,\kappa_{n_a}(\tau)\,\bar a e^{\bar a} + \kappa_{n_b}(\tau)\bar{b}\,,
\end{align}
where we use $|e^c - 1| \leq |c|e^{|c|}$. Then we have
\begin{align}
    \|\Psi_k(x)-x\|
    \leq \|x_2\|\,\kappa_{n_a}(\tau)\,\bar a e^{\bar a} + \kappa_{n_b}(\tau)\bar{b}
    \leq \|x\|\,(\omega\tau)^{n_a}\,\bar a e^{\bar a} + (\omega\tau)^{n_b}\bar{b}
\end{align}
For $x^{(0)}=u$ and $x^{(k)} = \Psi_k(x^{(k-1)})$, we have
\begin{align}
    \norm{x^{(k)}} =~& \norm{\Psi_k(x^{(k-1)}) - x^{(k-1)} + x^{(k-1)}} \leq \norm{\Psi_k(x^{(k-1)}) - x^{(k-1)}} + \norm{x^{(k-1)}}\\
    \leq~& \norm{x^{(k-1)}}\left((\omega\tau)^{n_a}\,\bar a e^{\bar a} + 1\right) + (\omega\tau)^{n_b}\bar{b}\,.
\end{align}
So the norm is multiplied by $\left((\omega\tau)^{n_a}\,\bar a e^{\bar a} + 1\right)$ and added to $(\omega\tau)^{n_b}\bar{b}$. Thus, we have
\begin{align}
    \norm{x^{(k)}} \leq~& \left((\omega\tau)^{n_a}\,\bar a e^{\bar a} + 1\right)^k\left[\norm{x^{(0)}} + k(\omega\tau)^{n_b}\bar{b}\right] \\
    \leq~& \left((\omega\tau)^{n_a}\,\bar a e^{\bar a} + 1\right)^K\left[\norm{x^{(0)}} + K(\omega\tau)^{n_b}\bar{b}\right]
\end{align}
Therefore
\begin{align}
    \|\Psi(u)-u\| \le~& \sum_{k=1}^{K}\big\|\Psi_k(x^{(k-1)})-x^{(k-1)}\big\|
    \le \sum_{k=1}^{K}\big(\|x^{(k-1)}\|\,(\omega\tau)^{n_a}\,\bar a e^{\bar a} + (\omega\tau)^{n_b}\bar{b}\big)\\
    \le~& K\big(\left((\omega\tau)^{n_a}\,\bar a e^{\bar a} + 1\right)^{K}\left[\norm{u} + K(\omega\tau)^{n_b}\bar{b}\right]\,(\omega\tau)^{n_a}\,\bar a e^{\bar a} + K(\omega\tau)^{n_b}\bar{b}\big)\\
    \le~& K\,\bar{a} e^{\bar a}\omega^{n_a}\left(\omega^{n_a}\,\bar a e^{\bar a} + 1\right)^{K}\left[\tau^{n_a}\norm{u} + K(\omega\tau)^{n_b}\bar{b}\right] + K^2(\omega\tau)^{n_b}\bar{b}\\
    \leq~& C\left[\tau^{n_a}\norm{u} + \tau^{n_b}\right]\,,
\end{align}
for some constant $C$.
\end{proof}

\begin{proof}[Proof of \cref{lemma:gate-orders}]
From our parameterization \cref{eq:our_model} we have
\begin{align}
    y - x = \tau v(x) + u + \Psi(u) - u\,.
\end{align}
From \cref{lemma:gate-perturbation}, we have
\begin{align}
    \norm{\Psi(u) - u} \leq C(\tau^{n_a}\norm{u} + \tau^{n_b})\,,
\end{align}
and since $u$ is Gaussian, we can write  $\mean\|u\|^m = c_m\,\tau^{m/2}$ with $c_m$ being a constant depending only on $g,m$ and number of dimensions, so we have
\begin{align}
    \mean\norm{\Psi(u) - u}^m \le C\big(\tau^{m n_a}\,\mean\|u\|^m + \tau^{m n_b}\big) = C\big(\tau^{m (n_a + 1/2)} + \tau^{m n_b}\big) = O(\tau^{m\beta})\,,
\end{align}
where we denote $\beta \coloneqq \min\{n_a+\tfrac12,\; n_b\}$, and note that by the condition $n_a>\tfrac12$ and $n_b>1$, we have $\beta > 1$.

The first moment is then
\begin{align}
    \mean[y-x] = \tau v(x) + \mean u + \mean \left[\Psi(u) - u\right] = \tau v(x) + O(\tau^{\beta})\,.
    \label{eq:first-moment-full}
\end{align}
The second moment we write down using $r \coloneqq \Psi(u) - u$. That is
\begin{align}
    \E[(y-x)(y-x)^\top]
    =~& \tau^2 vv^\top + \tau gg^\top + \E[rr^\top] +\tau\big(v\,\E[u+r]^\top + \E[u+r]\,v^\top\big) + \\
    ~&+ \E[ur^\top]+\E[ru^\top]\,.
\end{align}
Furthermore, using $\mean u = 0$ and Cauchy-Schwarz, we have
\begin{align}
    \norm{\mean[ur^\top]} \leq~& \sqrt{\mean\norm{u}^2 \mean \norm{r}^2} = O(\tau^{1/2 + \beta})\\
    \norm{\mean[rr^\top]} \leq~& \mean \norm{r}^2 = O(\tau^{2\beta})\\
    \norm{v\,\E[u+r]^\top} = \norm{v\,\E r^\top} \leq~& \sqrt{\mean\norm{v}^2 \mean \norm{r}^2} = O(\tau^{\beta})\,.
\end{align}
Since $\beta > 1$, we, finally, have
\begin{align}
    \E[(y-x)(y-x)^\top] = \tau gg^\top + o(\tau)\,.
\end{align}

To prove the no jump condition in \cref{eq:boundary-local}, we can use Markov's inequality i.e., for every $\eps>0$,
\begin{align}
    \mathbb{P}\big(\|y-x\|>\eps\big) \le \frac{\E\|y-x\|^4}{\eps^{4}} \leq \frac{1}{\eps^{4}} 27\big(\tau^4\|v(x)\|^4+\mean\|u\|^4+\mean\|r\|^4\big) = O(\tau^2)\,,
\end{align}
where we use $(a + b + c)^4 \leq 27(a^4 + b^4 + c^4)$. Thus, we show $\mathbb{P}\big(\|y-x\|>\eps\big) / \tau \to 0$. Finally, we show that the truncation does not matter for the moments $k\in\{1,2\}$, by Cauchy--Schwarz, 
\begin{align}
    \E\big[\|y-x\|^{k}\,\mathbf{1}\{\|y-x\|>\eps\}\big]
    \le \sqrt{\E\|y-x\|^{2k}}\,\sqrt{\mathbb{P}(\|y-x\|>\eps)}
    = O(\tau^{k/2})\,O(\tau) = o(\tau)\,. \nonumber
\end{align}
\end{proof}

\section{Derivations}
\label{app:derivations}

\subsection{Kolmogorov--Chapman KL loss derivation}
\label{app:ck-kl-loss}
Fix a source state $x$ and times $s<r<t$, and hold $\eta$ fixed when optimizing $\theta$.
The two-step target density is
\begin{align}
    q_{s,r,t}(y\mid x;\eta)
    &=\int dx'\,q_{r,t}(y\mid x';\eta)q_{s,r}(x'\mid x;\eta).
\end{align}
We expand the KL divergence as
\begin{align}
    &\KL\left(q_{s,r,t}(\cdot\mid x;\eta)
      \,\middle\|\,q_{s,t}(\cdot\mid x;\theta)\right)\nonumber\\
    &\quad=\int dy\,q_{s,r,t}(y\mid x;\eta)
      \log\frac{q_{s,r,t}(y\mid x;\eta)}{q_{s,t}(y\mid x;\theta)}\nonumber\\
    &\quad=-\int dy\,q_{s,r,t}(y\mid x;\eta)
      \log q_{s,t}(y\mid x;\theta)+C\nonumber\\
    &\quad=-\int dx'\,dy\,q_{s,r}(x'\mid x;\eta)q_{r,t}(y\mid x';\eta)
      \log q_{s,t}(y\mid x;\theta)+C\nonumber\\
    &\quad=\E_{\substack{
        x'\sim q_{s,r}(\cdot\mid x;\eta)\\
        y\sim q_{r,t}(\cdot\mid x';\eta)
      }}\left[-\log q_{s,t}(y\mid x;\theta)\right]+C,
\end{align}
where
\begin{align}
    C&=\int dy\,q_{s,r,t}(y\mid x;\eta)\log q_{s,r,t}(y\mid x;\eta)
\end{align}
is the negative entropy of the fixed target and is independent of $\theta$.
The two objectives therefore have the same minimizers,
\begin{align}
    &\argmin_\theta\KL\left(q_{s,r,t}(\cdot\mid x;\eta)
      \,\middle\|\,q_{s,t}(\cdot\mid x;\theta)\right)\nonumber\\
    &\quad=\argmin_\theta\E_{\substack{
        x'\sim q_{s,r}(\cdot\mid x;\eta)\\
        y\sim q_{r,t}(\cdot\mid x';\eta)
      }}\left[-\log q_{s,t}(y\mid x;\theta)\right].
\end{align}
Thus, we sample two-step transitions and evaluate their likelihood under the direct transition model without evaluating the composed density.
Taking the expectation over the sampled source states and times gives \cref{eq:ck-kl-mle}. With $\eta=\theta^\star$, $\theta^\star$ minimizes $KL\left(q_{s,r,t}(\cdot\mid x;\theta^\star)\,\middle\|\,q_{s,t}(\cdot\mid x;\theta)\right)$ over $\theta$. If the minimum is zero for every $x$ and $s<r<t$, then $q_{s,t}(\cdot \mid x;\theta^\star)$ statisfies the Kolmogorov--Chapman equation \cref{eq:chapman-kolmogorov}.

\subsection{Learning the boundary conditions}
\label{app:learning-boundary-conditions}

In \cref{sec:parameterization}, we enforce the boundary conditions through the parameterization. Alternatively, we can retain the locality and diffusion conditions in the parameterization and learn the infinitesimal drift through a training objective.
We consider two approaches: direct regression to the known vector field and likelihood training on short-time samples generated from the SDE's drift and diffusion.
We also describe a conditional regression formulation for generative modeling, where the marginal vector field is unavailable but conditional training targets can be constructed.

\paragraph{Drift regression}
Instead of adding the known displacement $\tau v(x)$ as in \cref{eq:our_model}, we apply the flow as follows,
\begin{align}
    y_\theta &= \Psi(x+\sqrt{\tau}\,gz;x,\theta),
    \qquad z\sim\Normal(0,I).
\end{align}
We use the coupling layers in \cref{eq:time_affine_layer} with $n_a=n_b=1$, allowing the flow to introduce a displacement proportional to $\tau$ while preserving the locality and diffusion conditions.

Assume log-scale and shift are uniformly bounded over their inputs for sufficiently small $\tau$, and jointly continuous in their inputs and $\tau$ at $\tau=0$.
The model's infinitesimal drift is then
\begin{align}
    v_\theta(x)
    &:=\left.\deriv{}{\tau}\E_z[y_\theta-x]\right|_{\tau=0}
      =\left.\deriv{\Psi(x;x,\theta)}{\tau}\right|_{\tau=0},
\end{align}
where the derivative of $\Psi$ holds its input and source state fixed.

Since every coupling layer is the identity at $\tau=0$, differentiating their composition gives
\begin{align}
    v_\theta(x)
    &=\sum_{k=1}^{K}
      \left.\deriv{\Psi_k(x;x,\theta)}{\tau}\right|_{\tau=0}.
\end{align}
For $n_a=n_b=1$, both gates have derivative $\omega$ at $\tau=0$.
A layer that transforms the second block therefore contributes
\begin{align}
    \left.\deriv{\Psi_k(x;x,\theta)}{\tau}\right|_{\tau=0}
    &=\omega
    \begin{pmatrix}
        0\\
        x_2\odot a_k(x_1;x,\theta)+b_k(x_1;x,\theta)
    \end{pmatrix},
\end{align}
where $x=[x_1,x_2]$ and the conditioners are evaluated at $\tau=0$.
For layers transforming the first block, the block roles are exchanged.
Thus, we can evaluate the infinitesimal drift directly from the scale and shift conditioners.
Expanding the gated transformations gives $\E[y_\theta-x]=\tau v_\theta(x)+o(\tau)$.
By \cref{lemma:gate-perturbation} with $n_a=n_b=1$, the flow correction is at most $C\tau(\|x+\sqrt{\tau}gz\|+1)$, so $\beta=1$ in \cref{app:gate-orders}. The diffusion and locality estimates there need only $\beta>\tfrac12$.

We learn this drift by minimizing
\begin{align}
    \cL_{\mathrm{drift}}(\theta)
    &=\E_{x\sim\pi}\left[\|v_\theta(x)-v(x)\|^2\right],
\end{align}
where $\pi$ is the distribution of source states used for training.

For generative modeling, we can sample states and vector-field targets from prescribed conditional probability paths, following conditional flow matching \citep{lipman_flow_2023}. When the conditional targets average to the marginal vector field given the state and time, this regression objective differs from regression against the marginal field only by a constant independent of $\theta$.

\paragraph{Short-time likelihood}
Alternatively, we can supervise the model using short-time Euler--Maruyama transitions.
For a fixed small elapsed time $\tau>0$, we sample
\begin{align}
    y &= x+\tau\,v(x)+\sqrt{\tau}\,gz,
    \qquad z\sim\Normal(0,I).
\end{align}
We maximize the likelihood of these samples under the model's transition density at elapsed time $\tau$,
\begin{align}
    \cL_{\mathrm{EM}}(\theta)
    &=\E_{x\sim\pi, \, z\sim\Normal(0,I)}
      \left[-\log q_{0,\tau}(y\mid x;\theta)\right].
\end{align}
The target has mean displacement $\tau\,v(x)$ and covariance $\tau\,gg^\top$, giving the required infinitesimal moments as $\tau\to0$.
At a fixed $\tau$, this objective fits the Euler--Maruyama approximation and does not by itself guarantee the boundary conditions, which concern the limit $\tau\to0$.

\subsection{Gaussian reference kernel}
\label{app:gaussian-reference}
We derive the Gaussian reference kernels for the first- and second-order autonomous parameterizations in \cref{eq:our_model,eq:our_model_second}.
The source state $x$ denotes the position in the first-order case and the joint position--momentum state in the second-order case.
By time homogeneity, we take the source time to be zero and write $p_\tau(y\mid x)$ for the exact transition density over elapsed time $\tau$.

Throughout these derivations, $\Psi$ is set to the identity, so the model reduces to the Gaussian reference density $q^{\mathrm{ref}}_\tau(y\mid x)$ with mean $\mu=x+m$ and covariance $\Sigma=LL^\top$.
All expectations and covariances are conditional on the source state, held fixed in time derivatives.
We assume a smooth vector field with bounded derivatives and at most linear growth, and constant isotropic noise with scalar coefficient $g$, to justify the expansions and integrations by parts below.

\subsubsection{First-order Gaussian reference}
\label{app:first-order-gaussian-reference}

Let $p_\tau(y\mid x)$ denote the conditional density of $X_\tau$ given $X_0=x$, evaluated at $y$.
Conditioning on $X_0=x$, we have the SDE
\begin{align}
    dX_\tau&=v(X_\tau)\,d\tau+g\,dW_\tau,
\end{align}
and its corresponding Fokker--Planck equation
\begin{align}
    \deriv{p_\tau(y\mid x)}{\tau}
    &=-\langle\nabla_y,p_\tau(y\mid x)v(y)\rangle
      +\frac{g^2}{2}\Delta_y p_\tau(y\mid x),
    \label{eq:first-order-fokker-planck}
\end{align}
with $p_0(y\mid x)=\delta(y-x)$.

To obtain a Gaussian reference with a closed-form sampler, we approximate the vector field about the source position.
Its Taylor expansion is
\begin{align}
    v(y)&=v(x)+\deriv{v(x)}{x}(y-x)+O(\|y-x\|^2).
    \label{eq:reference-coefficient-expansion}
\end{align}
Retaining only the zeroth-order term gives the reference SDE
\begin{align}
    dy_\tau&=v(x)\,d\tau+g\,dW_\tau,
    \qquad y_0=x.
\end{align}
Its transition density satisfies
\begin{align}
    \deriv{q^{\mathrm{ref}}_\tau(y\mid x)}{\tau}
    &=-\langle\nabla_y,q^{\mathrm{ref}}_\tau(y\mid x)v(x)\rangle
      +\frac{g^2}{2}\Delta_y q^{\mathrm{ref}}_\tau(y\mid x),
\end{align}
with $q^{\mathrm{ref}}_0(y\mid x)=\delta(y-x)$.
The reference SDE has constant drift and diffusion, so its transition kernel is Gaussian. 
We derive its mean and covariance below.

\paragraph{Mean and covariance}
Starting with the first moment, we have
\begin{align}
    \mu(x,\tau) &= \int dy\,y\, q^{\mathrm{ref}}_\tau(y\mid x).
\end{align}
Differentiating with respect to $\tau$ and integrating by parts gives
\begin{align}
    \deriv{\mu}{\tau}
    &=\int dy\,y\,\deriv{q^{\mathrm{ref}}_\tau(y\mid x)}{\tau}\nonumber\\
    &=- \int dy\,y\, \langle\nabla_y,q^{\mathrm{ref}}_\tau(y\mid x)v(x)\rangle
      +\frac{g^2}{2}\int dy\,y\,\Delta_y q^{\mathrm{ref}}_\tau(y\mid x)\nonumber\\
    &= \int dy\,q^{\mathrm{ref}}_\tau(y\mid x)v(x) = v(x).
\end{align}
Using the initial condition $\mu(x,0) = x$, we obtain
\begin{align}
    \mu(x,\tau) &= x + \tau v(x).
\end{align}
Similarly, differentiating the second moment and integrating by parts gives
\begin{align}
    \deriv{}{\tau}\E[yy^\top]
    &=\int dy\,yy^\top\,
    \deriv{q^{\mathrm{ref}}_\tau(y\mid x)}{\tau}\nonumber\\
    &=v(x)\mu^\top+\mu v(x)^\top+g^2I_d.
\end{align}
Since $\Sigma=\E[yy^\top]-\mu\mu^\top$, its derivative is
\begin{align}
    \deriv{\Sigma}{\tau}
    &=\deriv{}{\tau}\E[yy^\top]
    -\deriv{\mu}{\tau}\mu^\top
    -\mu\left(\deriv{\mu}{\tau}\right)^\top\nonumber\\
    &=g^2I_d.
\end{align}
Using the initial condition $\Sigma(0)=0$, we obtain
\begin{align}
    \Sigma(\tau)&=\tau g^2I_d.
\end{align}

The reference density and sampler are
\begin{align}
    q^{\mathrm{ref}}_\tau(y\mid x)
    &=\cN\bigl(y\mid\mu(x,\tau),\Sigma(\tau)\bigr),\\
    y&=x+\tau v(x)+u,
    \qquad u=\sqrt{\tau}\,gz,
    \quad z\sim\Normal(0,I_d).
    \label{eq:first-order-reference-sampling}
\end{align}
For nonzero $g$, the reference covariance is positive definite whenever $\tau>0$.

\subsubsection{Second-order Gaussian reference}
\label{app:second-order-gaussian-reference}

We start from the second-order Langevin dynamics
\begin{align}
    dX_\tau&=P_\tau\,d\tau,\nonumber\\
    dP_\tau&=\bigl(v(X_\tau)-\gamma P_\tau\bigr)\,d\tau+g\,dW_\tau,
\end{align}
where $\gamma > 0$ is the friction coefficient. For simplicity, we work in mass-weighted coordinates $M^{1/2}X$ and $M^{-1/2}P$, where the dynamics have unit mass, vector field $M^{-1/2}v$ and isotropic noise $\sqrt{2\gamma k_BT}$; the general case follows by rescaling the position and momentum rows of $\mu$ and $L$ by $M^{-1/2}$ and $M^{1/2}$. The dynamics are conditioned on the source state $x=(X,P)\in\mathbb{R}^{2d}$, where $X,P\in\mathbb{R}^d$, with $X_0=X$ and $P_0=P$. 

As in the first-order case, we hold the vector field fixed at the source position, $v(X)$, while keeping the momentum damping term unchanged.
Let $y(\tau)=(y_x(\tau),y_p(\tau))$ denote the reference process, which satisfies
\begin{align}
    dy_x(\tau)&=y_p(\tau)\,d\tau,\nonumber\\
    dy_p(\tau)&=\bigl(v(X)-\gamma y_p(\tau)\bigr)\,d\tau+g\,dW_\tau,
    \label{eq:second-order-reference-sde}
\end{align}
with $y(0)=x$. Here, $W_\tau$ is standard Brownian motion in $\mathbb{R}^d$, and $g$ is a scalar noise coefficient.
Its density $q^{\mathrm{ref}}_\tau(y\mid x)$, evaluated at $y=(y_x,y_p)$, satisfies
\begin{align}
    \deriv{q^{\mathrm{ref}}_\tau(y\mid x)}{\tau}
    &=-\langle\nabla_{y_x},y_pq^{\mathrm{ref}}_\tau(y\mid x)\rangle
      -\langle\nabla_{y_p},\bigl(v(X)-\gamma y_p\bigr)q^{\mathrm{ref}}_\tau(y\mid x)\rangle\nonumber\\
    &\qquad+\frac{g^2}{2}\Delta_{y_p}q^{\mathrm{ref}}_\tau(y\mid x),
    \label{eq:second-order-reference-fokker-planck}
\end{align}
with $q^{\mathrm{ref}}_0(y\mid x)=\delta(y_x-X)\delta(y_p-P)$.
The reference dynamics are affine in the joint state and driven by Gaussian noise, so their transition kernel is Gaussian. Now, we derive its mean and covariance.

\paragraph{Mean}
Starting with the first moment, we have
\begin{align}
    \mu(x,\tau)
    &=\begin{pmatrix}\mu_x(x,\tau)\\\mu_p(x,\tau)\end{pmatrix}
      =\int dy\,y\,q^{\mathrm{ref}}_\tau(y\mid x).
\end{align}
Differentiating with respect to $\tau$, substituting \cref{eq:second-order-reference-fokker-planck}, and integrating by parts gives
\begin{align}
    \deriv{\mu}{\tau}
    &=\int dy\,y\,\deriv{q^{\mathrm{ref}}_\tau(y\mid x)}{\tau}\nonumber\\
    &=-\int dy\,y\,\langle\nabla_{y_x},y_pq^{\mathrm{ref}}_\tau(y\mid x)\rangle\nonumber\\
    &\qquad-\int dy\,y\,\langle\nabla_{y_p},\bigl(v(X)-\gamma y_p\bigr)q^{\mathrm{ref}}_\tau(y\mid x)\rangle\nonumber\\
    &\qquad+\frac{g^2}{2}\int dy\,y\,\Delta_{y_p}q^{\mathrm{ref}}_\tau(y\mid x)\nonumber\\
    &=\int dy\,\begin{pmatrix}y_p\\v(X)-\gamma y_p\end{pmatrix}q^{\mathrm{ref}}_\tau(y\mid x)
      =\begin{pmatrix}\mu_p\\v(X)-\gamma\mu_p\end{pmatrix}.
\end{align}
Thus, the position and momentum means satisfy
\begin{align}
    \frac{d\mu_x}{d\tau}&=\mu_p,\nonumber\\
    \frac{d\mu_p}{d\tau}+\gamma\mu_p&=v(X),
\end{align}
with $\mu_x(0)=X$ and $\mu_p(0)=P$. Multiplying the momentum equation by $e^{\gamma\tau}$ and integrating from $0$ to $\tau$ gives
\begin{align}
    \frac{d}{d\tau}\left(e^{\gamma\tau}\mu_p(\tau)\right)
    &=e^{\gamma\tau}v(X),\nonumber\\
    e^{\gamma\tau}\mu_p(\tau)-P
    &=v(X)\int_0^\tau dr\,e^{\gamma r},\nonumber\\
    \mu_p(\tau)
    &=e^{-\gamma\tau}P+\frac{1-e^{-\gamma\tau}}{\gamma}v(X).
\end{align}

For position, integrating its mean equation gives
\begin{align}
    \mu_x(\tau)-X
    &=\int_0^\tau dr\,\mu_p(r)\nonumber\\
    &=P\int_0^\tau dr\,e^{-\gamma r}
      +\frac{v(X)}{\gamma}\int_0^\tau dr\,(1-e^{-\gamma r})\nonumber\\
    &=\frac{1-e^{-\gamma\tau}}{\gamma}P
      +\frac{\gamma\tau-1+e^{-\gamma\tau}}{\gamma^2}v(X).
\end{align}
Using the functions
\begin{align}
    \phi_1(\tau)&=\frac{1-e^{-\gamma\tau}}{\gamma},
    \qquad
    \phi_2(\tau)=\frac{\tau-\phi_1(\tau)}{\gamma},
\end{align}
we can write the mean as $\mu=x+m$, with
\begin{align}
    m&=\begin{pmatrix}
        \phi_1(\tau)P+\phi_2(\tau)v(X)\\
        \phi_1(\tau)\bigl(v(X)-\gamma P\bigr)
    \end{pmatrix}.
    \label{eq:second-order-reference-mean}
\end{align}

\paragraph{Covariance}
To construct the Gaussian sampler, we need the position variance, the momentum variance,
and their cross-covariance. We derive these from the reference process after subtracting its mean:
\begin{align}
    \widetilde X&=y_x-\mu_x,
    \qquad
    \widetilde P=y_p-\mu_p.
\end{align}
Both start at zero because the source state is fixed. Subtracting the mean equations from the reference dynamics gives
\begin{align}
    d\widetilde X&=\widetilde P\,d\tau,\nonumber\\
    d\widetilde P&=-\gamma\widetilde P\,d\tau+g\,dW_\tau,
\end{align}
where we note that the vector field cancels.

For each coordinate $j=1,\ldots,d$, define
\begin{align}
    S_{xx}&=\E[\widetilde X_j^2],
    \qquad S_{xp}=\E[\widetilde X_j\,\widetilde P_j],
    \qquad S_{pp}=\E[\widetilde P_j^2].
\end{align}
These quantities do not depend on $j$ and are initially zero. We write $W_{\tau,j}$ for the $j$th component of the Brownian motion.

Starting with $S_{pp}$, the product rule gives
\begin{align}
    d(\widetilde P_j^2)&=2\widetilde P_j\,d\widetilde P_j+(d\widetilde P_j)^2.
\end{align}
Using the It\^o identities $(dW_{\tau,j})^2=d\tau$ and $d\tau\,dW_{\tau,j}=(d\tau)^2=0$, we obtain
\begin{align}
    d(\widetilde P_j^2)
    &=-2\gamma\widetilde P_j^2\,d\tau+2g\widetilde P_j\,dW_{\tau,j}+g^2\,d\tau.
\end{align}
The stochastic integral has zero expectation, so
\begin{align}
    \frac{dS_{pp}}{d\tau}+2\gamma S_{pp}&=g^2.
\end{align}
Multiplying by $e^{2\gamma\tau}$, integrating, and using $S_{pp}(0)=0$ gives
\begin{align}
    \frac{d}{d\tau}\left(e^{2\gamma\tau}S_{pp}(\tau)\right)
    &=g^2e^{2\gamma\tau},\nonumber\\
    S_{pp}(\tau)&=\frac{g^2}{2\gamma}\left(1-e^{-2\gamma\tau}\right).
    \label{eq:second-order-reference-momentum-variance}
\end{align}

We repeat for $S_{xp}$, 
\begin{align}
    d(\widetilde X_j\,\widetilde P_j)
    &=\widetilde X_j\,d\widetilde P_j+\widetilde P_j\,d\widetilde X_j+d\widetilde X_j\,d\widetilde P_j.
\end{align}
The It\^o identities above give $d\widetilde X_j\,d\widetilde P_j=0$.
Substituting the centered dynamics gives
\begin{align}
    d(\widetilde X_j\,\widetilde P_j)
    &=\bigl(-\gamma\widetilde X_j\,\widetilde P_j+\widetilde P_j^2\bigr)d\tau+g\widetilde X_j\,dW_{\tau,j}.
\end{align}
Taking expectations,
\begin{align}
    \frac{dS_{xp}}{d\tau}+\gamma S_{xp}&=S_{pp},
\end{align}
and solving with the expression for $S_{pp}$ and the initial condition $S_{xp}(0)=0$ gives
\begin{align}
    \frac{d}{d\tau}\left(e^{\gamma\tau}S_{xp}(\tau)\right)
    &=\frac{g^2}{2\gamma}\left(e^{\gamma\tau}-e^{-\gamma\tau}\right),\nonumber\\
    e^{\gamma\tau}S_{xp}(\tau)
    &=\frac{g^2}{2\gamma^2}\left(e^{\gamma\tau}+e^{-\gamma\tau}-2\right),\nonumber\\
    S_{xp}(\tau)&=\frac{g^2}{2\gamma^2}\left(1-e^{-\gamma\tau}\right)^2.
    \label{eq:second-order-reference-cross-covariance}
\end{align}

Finally, for $S_{xx}$,
\begin{align}
    d(\widetilde X_j^2)
    &=2\widetilde X_j\,d\widetilde X_j+(d\widetilde X_j)^2
      =2\widetilde X_j\,\widetilde P_j\,d\tau.
\end{align}
Taking expectations and integrating gives
\begin{align}
    \frac{dS_{xx}}{d\tau}&=2S_{xp},\nonumber\\
    S_{xx}(\tau)
    &=\frac{g^2}{\gamma^2}\int_0^\tau dr\,\left(1-e^{-\gamma r}\right)^2\nonumber\\
    &=\frac{g^2}{\gamma^2}\left[
        \tau-\frac{2(1-e^{-\gamma\tau})}{\gamma}
        +\frac{1-e^{-2\gamma\tau}}{2\gamma}
    \right].
    \label{eq:second-order-reference-position-variance}
\end{align}

The same covariance structure holds for each coordinate, with zero covariance between different coordinates. Thus, the joint covariance is
\begin{align}
    \Sigma(\tau)&=\begin{pmatrix}
        S_{xx}I_d&S_{xp}I_d\\
        S_{xp}I_d&S_{pp}I_d
    \end{pmatrix}.
    \label{eq:second-order-reference-covariance}
\end{align}

\paragraph{Reference kernel and sampler}
The covariance factor satisfying $LL^\top=\Sigma$ is
\begin{align}
    L(\tau)&=\begin{pmatrix}
        \sqrt{S_{xx}}\,I_d&0\\[2pt]
        \dfrac{S_{xp}}{\sqrt{S_{xx}}}\,I_d&
        \sqrt{S_{pp}-\dfrac{S_{xp}^2}{S_{xx}}}\,I_d
    \end{pmatrix}.
    \label{eq:second-order-reference-cholesky}
\end{align}
For $g>0$ and $\tau>0$, this is the Cholesky factor with positive diagonal. At $\tau=0$, we define $L(0)=0$ by continuity.

The reference kernel and sampler are therefore
\begin{align}
    q^{\mathrm{ref}}_\tau(y\mid x)
    &=\cN\bigl(y\mid\mu(x,\tau),\Sigma(\tau)\bigr),
\end{align}
\begin{align}
    y_x&=\mu_x+\sqrt{S_{xx}}\,z_x,\nonumber\\
    y_p&=\mu_p+\frac{S_{xp}}{\sqrt{S_{xx}}}\,z_x
      +\sqrt{S_{pp}-\frac{S_{xp}^2}{S_{xx}}}\,z_p,
    \label{eq:second-order-reference-sampling}
\end{align}
where $z_x,z_p \sim \Normal(0,I_d)$ are independent. The shared $z_x$ produces the position--momentum correlation derived above.

\subsubsection{Midpoint reference and leading corrections}
\label{app:midpoint-reference}

For the duration schedule $h(\tau)$ of \cref{sec:design}, $h(\tau)=\tau$ for $\tau\leq\tau_0$, so the schedule does not affect the expansions below.
We retain the smoothness and moment assumptions stated above. Evaluating the vector field at $X_{\mathrm{mid}}=X+\tfrac12 h(\tau)P=X+\tfrac{\tau}{2}P$ for $\tau\leq\tau_0$ gives the reference mean
\begin{align}
    \mu_x&=X+\phi_1(\tau)P+\phi_2(\tau)v(X_{\mathrm{mid}}),\nonumber\\
    \mu_p&=e^{-\gamma\tau}P+\phi_1(\tau)v(X_{\mathrm{mid}}).
    \label{eq:midpoint-reference-mean}
\end{align}
Its covariance remains that of \cref{eq:second-order-reference-covariance}.
We denote its blocks by $\Sigma_{xx}=S_{xx}I_d$, $\Sigma_{xp}=S_{xp}I_d$, and $\Sigma_{pp}=S_{pp}I_d$.
Throughout this subsection, $\mu$ denotes the midpoint reference mean.

\paragraph{Mean corrections}
Let $\mu_x^\star=\E[X_\tau]$ and $\mu_p^\star=\E[P_\tau]$ denote the exact conditional means.
Taking expectations of the SDE gives
\begin{align}
    \deriv{\mu_x^\star}{\tau}&=\mu_p^\star,\nonumber\\
    \deriv{\mu_p^\star}{\tau}&=\E[v(X_\tau)]-\gamma\mu_p^\star,
    \qquad \mu_x^\star(0)=X,\quad \mu_p^\star(0)=P.
\end{align}
Since position has no direct diffusion term, differentiating the expected vector field at the source gives
\begin{align}
    \left.\deriv{}{\tau}\E[v(X_\tau)]\right|_{\tau=0}
    &=\deriv{v(X)}{X}P,\\
    \left.\deriv{^2}{\tau^2}\E[v(X_\tau)]\right|_{\tau=0}
    &=\deriv{^2v(X)}{X^2}[P,P]
      +\deriv{v(X)}{X}\bigl(v(X)-\gamma P\bigr).
\end{align}
Here, $\deriv{^2v(X)}{X^2}[P,P]$ is the second spatial derivative applied to $(P,P)$.
The midpoint vector field has the expansion
\begin{align}
    v(X_{\mathrm{mid}})
    &=v(X)+\frac{\tau}{2}\deriv{v(X)}{X}P
      +\frac{\tau^2}{8}\deriv{^2v(X)}{X^2}[P,P]+O(\tau^3).
\end{align}
Taylor expanding the exact mean equations and subtracting \cref{eq:midpoint-reference-mean} yields
\begin{align}
    \mu_x^\star-\mu_x
    &=-\frac{\tau^3}{12}\deriv{v(X)}{X}P+O(\tau^4),\\
    \mu_p^\star-\mu_p
    &=\tau^3\biggl[
        \frac1{24}\deriv{^2v(X)}{X^2}[P,P]
        +\frac16\deriv{v(X)}{X}v(X)\nonumber\\
    &\qquad-\frac{\gamma}{12}\deriv{v(X)}{X}P
      \biggr]+O(\tau^4).
    \label{eq:midpoint-mean-corrections}
\end{align}
Thus, both mean corrections start at order $\tau^3$.
Evaluating the vector field at the source instead leaves a momentum mean correction $\tfrac{\tau^2}{2}\deriv{v(X)}{X}P+O(\tau^3)$.
The midpoint removes this term while preserving the reference's infinitesimal drift.
Since the covariance is unchanged and the reference remains Gaussian, its diffusion and locality conditions are also preserved.

\paragraph{Covariance corrections}
The It\^o product rule gives the exact conditional covariance equations
\begin{align}
    \deriv{}{\tau}\Cov(X_\tau)
    &=\Cov(X_\tau,P_\tau)+\Cov(P_\tau,X_\tau),\\
    \deriv{}{\tau}\Cov(X_\tau,P_\tau)
    &=\Cov(P_\tau)-\gamma\Cov(X_\tau,P_\tau)+\Cov(X_\tau,v(X_\tau)),\\
    \deriv{}{\tau}\Cov(P_\tau)
    &=-2\gamma\Cov(P_\tau)+g^2I_d
      +\Cov(v(X_\tau),P_\tau)+\Cov(P_\tau,v(X_\tau)).
\end{align}
All covariance blocks vanish at $\tau=0$.
The reference captures the leading terms
\begin{align}
    \Sigma_{pp}&=g^2\tau I_d+O(\tau^2),\nonumber\\
    \Sigma_{xp}&=\frac{g^2\tau^2}{2}I_d+O(\tau^3),\nonumber\\
    \Sigma_{xx}&=\frac{g^2\tau^3}{3}I_d+O(\tau^4).
\end{align}
Expanding the vector field about $X$ gives
\begin{align}
    \Cov(v(X_\tau),P_\tau)
    &=\frac{g^2\tau^2}{2}\deriv{v(X)}{X}+O(\tau^3),\\
    \Cov(X_\tau,v(X_\tau))
    &=\frac{g^2\tau^3}{3}\left(\deriv{v(X)}{X}\right)^\top+O(\tau^4).
\end{align}
Subtracting the reference covariance equations and integrating the leading terms therefore yields
\begin{align}
    \Cov(P_\tau)-\Sigma_{pp}
    &=\frac{g^2\tau^3}{6}
      \left[\deriv{v(X)}{X}+\left(\deriv{v(X)}{X}\right)^\top\right]+O(\tau^4),\\
    \Cov(X_\tau,P_\tau)-\Sigma_{xp}
    &=\frac{g^2\tau^4}{24}
      \left[\deriv{v(X)}{X}+3\left(\deriv{v(X)}{X}\right)^\top\right]+O(\tau^5),\\
    \Cov(X_\tau)-\Sigma_{xx}
    &=\frac{g^2\tau^5}{30}
      \left[\deriv{v(X)}{X}+\left(\deriv{v(X)}{X}\right)^\top\right]+O(\tau^6).
    \label{eq:midpoint-covariance-corrections}
\end{align}
These corrections are the same for the source and midpoint references because their covariances are identical.

\paragraph{Free-flight term}
\label{app:free-flight-term}
For atom $i$, we scale the position shift as $b_{k,i}=\ell_i(\tau)\widetilde b_{k,i}$, where $\widetilde b_{k,i}$ is the bounded, dimensionless network output. In physical units,
\begin{align}
    \ell_i(\tau)
    &=\frac{v_{\mathrm{th},i}\tau}
    {\sqrt{1+\bigl(v_{\mathrm{th},i}\tau/\ell_{\max}\bigr)^2}},\label{eq:free-flight-factor}
\end{align}
with thermal velocity $v_{\mathrm{th},i}=\sqrt{k_BT/M_i}$ and $\ell_{\max}=0.01\,\mathrm{nm}$.
This factor follows the ballistic free-flight displacement at small elapsed times and saturates to $\ell_{\max}$.
Since $\ell_i(\tau)=v_{\mathrm{th},i}\tau+O(\tau^3)$, it contributes one additional power of $\tau$ to the position shift, giving $O(\tau^4)$ when combined with the cubic gate.
In dimensionless coordinates, we divide $\ell_i$ by the position normalization scale.

\phantomsection
\paragraph{Gate orders}
\label{app:reference-gate-orders}
We select gate orders empirically within the family that preserves the boundary conditions, using $(n_a,n_b)=(2,3)$ for momentum updates and $(3,3)$ for position updates.
The leading mean and covariance corrections in \cref{eq:midpoint-mean-corrections,eq:midpoint-covariance-corrections} provide a starting point for these choices. Both mean corrections are $O(\tau^3)$. The momentum and position covariance corrections are $O(\tau^3)$ and $O(\tau^5)$, respectively, compared with reference variances of $O(\tau)$ and $O(\tau^3)$.
Thus, both covariance corrections are of relative order $O(\tau^2)$. Comparing these powers suggests log-scale gates of order $O(\tau^2)$ and shifts of order $O(\tau^3)$.

For momentum, the quadratic log-scale gate and cubic shift follow these orders. For position, the cubic log-scale gate suppresses scale updates by one additional power of $\tau$. The cubic shift gate also includes the $O(\tau)$ free-flight factor in \cref{eq:free-flight-factor}, giving an effective shift of order $O(\tau^4)$, one order higher than the leading mean correction. Our empirical choices therefore suppress position updates more strongly at small $\tau$.
This comparison motivates the gate powers without requiring the learned flow to match every leading mean and covariance correction. We establish boundary preservation below.

\paragraph{Boundary preservation}
Both second-order references satisfy the boundary conditions and have $\E\|L(\tau)z\|^2=O(\tau)$. Under the boundedness assumptions of \cref{lemma:gate-orders}, define the flow correction as
\begin{align}
    \xi_\tau
    &=\Psi(L(\tau)z;x,\theta)-L(\tau)z.
\end{align}
With $n_a\geq2$ for both blocks and effective shift powers of at least two, the perturbation bound in \cref{eq:flow-perturbation} gives
\begin{align}
    \left(\E\|\xi_\tau\|^2\right)^{1/2}
    &\leq C\left[
        \tau^2\left(\E\|L(\tau)z\|^2\right)^{1/2}+\tau^2
    \right]
    =O(\tau^2)=o(\tau).
\end{align}
The correction therefore leaves the infinitesimal drift and diffusion unchanged.
More generally, $\min\{n_a+\tfrac12,n_b\}>1$ suffices, as in \cref{lemma:gate-orders}, which also covers the lower-order gates of \cref{tab:al2_ablation_flow}.
For every $\epsilon>0$, locality follows from
\begin{align}
    \Pr(\|y-x\|>\epsilon)
    &\leq
    \Pr(\|m+L(\tau)z\|>\epsilon/2)
    +\frac{4\E\|\xi_\tau\|^2}{\epsilon^2}
    =o(\tau).
\end{align}

\section{Implementation Details}
\label{app:implementation-details}

\subsection{Model Details}
\label{app:model-details}

\paragraph{Equivariance}
\label{app:equivariance}
We center positions at the center of mass and keep positions and momenta in the center-of-mass-free subspace, where densities are evaluated. The learned flow acts on this subspace. We propagate the center-of-mass position and total momentum separately using the exact force-free Langevin transition for the total mass. For peptides, the source configuration defines a rotation $R=[e_1,e_2,e_3]$ from the N, C$_\alpha$ and C atoms of the first complete residue,
\begin{align}
    e_1=\frac{X_{\mathrm{N}}-X_{\mathrm{C}_\alpha}}{\|X_{\mathrm{N}}-X_{\mathrm{C}_\alpha}\|},\qquad
    e_2\propto(X_{\mathrm{C}}-X_{\mathrm{C}_\alpha})-\langle X_{\mathrm{C}}-X_{\mathrm{C}_\alpha},e_1\rangle e_1,\qquad
    e_3=e_1\times e_2.
\end{align}
The flow acts on inputs rotated by $R^\top$, and its outputs are rotated back by $R$. Since $R$ depends only on the source and is orthogonal, the log-determinant is unchanged and the kernel is rotation equivariant. This does not extend to LJ, so instead we apply a random rotation augmentation to each training example.

\paragraph{Gaussian reference duration schedule}
\label{app:duration-schedule}
For $\tau>\tau_0$, the schedule $h(\tau)$ of \cref{sec:design} rescales the transport and vector-field terms of the midpoint mean \cref{eq:midpoint-reference-mean} and the position rows of the Cholesky factor \cref{eq:second-order-reference-cholesky}, while remaining terms keep the full elapsed time:
\begin{align}
    \mu_x&=X+\tfrac{h(\tau)}{\tau}\phi_1(\tau)P+\bigl(\tfrac{h(\tau)}{\tau}\bigr)^2\phi_2(\tau)v(X_{\mathrm{mid}}),\nonumber\\
    \mu_p&=e^{-\gamma\tau}P+\tfrac{h(\tau)}{\tau}\phi_1(\tau)v(X_{\mathrm{mid}}),\nonumber\\
    L_h(\tau)&=\mathrm{diag}\bigl(\tfrac{h(\tau)}{\tau}I_d,\,I_d\bigr)L(\tau),
\end{align}
with $X_{\mathrm{mid}}=X+\tfrac12h(\tau)P$. For $\tau\leq\tau_0$, $h(\tau)/\tau=1$, resulting in the reference of \cref{app:midpoint-reference}.

\paragraph{Reference forces}
The midpoint force is projected to remove its net force and its torque about the source positions, so that the reference mean satisfies the same conservation constraints as the flow.

\paragraph{Coupling layers}
The flow has $K=8$ coupling blocks that alternate between position and momentum updates, starting with positions, without permutations.
Each block projects its affine update onto the allowed subspace:
\begin{align}
    x_2^{(k+1)}
    =x_2^{(k)}+\Pi_k\Biggl[
    \left(\exp\,\left(\frac{(\omega\tau)^{n_a}}{1+(\omega\tau)^{n_a}}\cdot a_k\right)-1\right)\odot x_2^{(k)}
    +\frac{(\omega\tau)^{n_b}}{1+(\omega\tau)^{n_b}}\cdot b_k\Biggr].
\end{align}
In position blocks, $\Pi_k$ removes changes to the center of mass.
In momentum blocks, it removes changes to total momentum and angular momentum about the source positions.
We evaluate the log-determinant on this constrained subspace.

We bound the learned scales and shifts in the standardized units of \cref{app:standardization}. Position blocks use
\begin{align}
    a_k=\tanh\alpha_k,\qquad
    \widetilde b_{k,i}=4\tanh\beta_{k,i},\qquad
    b_{k,i}=\frac{\ell_i(\tau)}{X_{\mathrm{std}}}\widetilde b_{k,i},
\end{align}
where $\ell_i(\tau)$ is the free-flight factor in \cref{eq:free-flight-factor}. Momentum blocks use
\begin{align}
    a_k=3\tanh\alpha_k,\qquad b_k=32\tanh\beta_k.
\end{align}
These bounds satisfy the assumption in \cref{lemma:gate-orders}.

\paragraph{Angular-momentum map}
We observed that the learned model could introduce excess spin that accumulated over successive steps and destabilized long rollouts.
For a rotationally invariant potential, internal forces exert no net torque, so angular momentum evolves through friction and thermal noise.
We use the Gaussian reference to approximate this evolution and constrain the coupling layers to preserve its angular momentum about the source positions.
A final momentum transformation transfers this angular-momentum value to the proposed geometry.

Let $P'$ denote the momenta after the coupling layers, before this transformation. Define
\begin{align}
    \Lambda=\sum_iX_{t,i}\times P'_i,\qquad
    \Delta L=\Lambda-\sum_iX_{t+\tau,i}\times P'_i.
\end{align}
Let $I_{t+\tau}$ denote the inertia tensor at the proposed positions. Keeping these positions fixed, we add a rigid rotation:
\begin{align}
    T(P')_i=P'_i+M_i\bigl[(I_{t+\tau}^{-1}\Delta L)\times X_{t+\tau,i}\bigr],
\end{align}
This gives $\sum_iX_{t+\tau,i}\times T(P')_i=\Lambda$, retaining the reference angular-momentum distribution, including its friction and noise.

For fixed source and proposed positions, $T$ is linear in the momenta. To express its determinant, define the matrix
\begin{align}
    I_{\mathrm{mix}}(a,b)=\sum_iM_i\bigl[(a_i\cdot b_i)\,\mathrm{Id}-b_ia_i^\top\bigr],
\end{align}
so that $I_{t+\tau}=I_{\mathrm{mix}}(X_{t+\tau},X_{t+\tau})$.
When $I_{t+\tau}$ and $I_{\mathrm{mix}}(X_t,X_{t+\tau})$ are nonsingular, the map is invertible and
\begin{align}
    \log|\det T|
    =\log|\det I_{\mathrm{mix}}(X_t,X_{t+\tau})|-\log|\det I_{t+\tau}|.
\end{align}
Extending $T$ to the joint state by leaving positions unchanged, the model becomes
\begin{align}
    y=T\bigl(x+m(x,\tau)+\Psi(L(\tau)z;x,\theta)\bigr).
\end{align}
We include the log-determinant of $T$ when evaluating the model density.

\paragraph{Conditioner network}
Per-atom tokens of width $384$ combine the source and mean embeddings with embeddings of atom type, residue type, residue index, chain length and atom name, and pass through one cross-attention layer over these chemical embeddings and two self-attention layers \citep{vaswani_attention_2023}.
A per-atom conditioning vector is computed from the $\tau$ embedding, the covariance features, the speed and impulse expansions, and the angular-momentum features by a linear map and two residual SwiGLU layers, where $\tau$ enters through a sinusoidal embedding \citep{peebles2023scalable}.
These conditioning features are detailed in the paragraph below.
Each coupling block adds to the tokens a linear embedding of the variables it conditions on, divided by the reference widths without the duration schedule, $\sqrt{S_{xx}/M_i}$ and $\sqrt{M_iS_{pp}}$, and applies two self-attention and SwiGLU layers with six heads, hidden width $768$ and AdaLN-Zero modulation \citep{peebles2023scalable} by the conditioning vector.
A zero-initialized linear head outputs $(\alpha_k,\beta_k)$ for every atom, so the flow is the identity at initialization.

\begin{samepage}
\paragraph{Conditioning features}
\label{app:conditioning-features}
Besides the source state and $\tau$ embedding, each particle $i$ receives features of the Gaussian reference.
For first-order dynamics, the reference has mean displacement $m(x,\tau)=h(\tau)\,v(x)$ and width $\sigma(\tau)=g\sqrt{\tau}$,
\begin{align}
    f_{\mathrm{ref},i}=\bigl(m_i(x,\tau),\,\log\sigma(\tau)\bigr).
\end{align}
\end{samepage}
For second-order dynamics, with $m(x,\tau)$ from \cref{eq:second-order-reference-mean} and the covariance blocks of \cref{eq:second-order-reference-covariance} for atom $i$,
\begin{align}
    f_{\mathrm{ref},i}=\bigl(m_i(x,\tau),\,\log\sigma_{x,i},\,\log\sigma_{p,i},\,\rho_{xp,i}\bigr),
\end{align}
where, with $S_{xx}$, $S_{xp}$ and $S_{pp}$ from \cref{app:second-order-gaussian-reference} and atom mass $M_i$,
\begin{align}
    \sigma_{x,i}=\tfrac{h(\tau)}{\tau}\sqrt{S_{xx}/M_i},\qquad
    \sigma_{p,i}=\sqrt{M_iS_{pp}},\qquad
    \rho_{xp,i}=\frac{S_{xp}}{\sqrt{S_{xx}S_{pp}}}.
\end{align}
Both log widths are min--max normalized to $[-1,1]$ in standardized units, using their smallest value over atoms at $\tau=0.01$\,fs and their largest at $T_{\max}$, with $\tau$ floored at $10^{-6}$\,ps.
The mean enters with the source state through four per-atom vectors, $\widehat X_i$, $\widehat P_i$, $m_{x,i}/(v_{\mathrm{th},i}\tau)$ and $m_{p,i}T_{\mathrm{std}}/(P_{\mathrm{std},i}\tau)$, with each vector $u$ mapped to $(u\,\mathrm{asinh}\|u\|/\|u\|,\log(1+\|u\|))$, which keeps its direction and log-scales its magnitude, embedded by an MLP, and the four embeddings are added.
Following \citet{ripken2026hfm}, each atom also receives a Gaussian basis expansion of its speed $\|\widehat P_i\|$ ($8$ functions on $[0,6]$), and likewise of its reference impulse $\tfrac{h(\tau)}{\tau}\phi_1(\tau)\|v_i(X_{\mathrm{mid}})\|/P_{\mathrm{std},i}$ ($16$ functions on $[0,3.5]$), together with the cosine between the two.
All atoms also share two angular-momentum features: $\Lambda$ itself and its deviation from the damped source value $e^{-\gamma\tau}\sum_iX_{t,i}\times P_{t,i}$, each divided by its thermal standard deviation and passed into $\mathrm{asinh}$, which is linear near zero and logarithmic for large values.

\subsection{Training Details}

\paragraph{Standardization}
\label{app:standardization}
After centering molecular positions at the center of mass, we standardize the state and elapsed time as
\begin{align}
    \widehat X_i=\frac{X_i}{X_{\mathrm{std}}},\qquad
    \widehat P_i=\frac{P_i}{P_{\mathrm{std},i}},\qquad
    \widehat\tau=\frac{\tau}{T_{\mathrm{std}}}.
\end{align}
The position scale is the root mean square over training configurations, atoms and Cartesian coordinates, $X_{\mathrm{std}}^2=(3N)^{-1}\E_{\mathrm{train}}[\sum_i\|X_i\|^2]$, where $N$ is the number of atoms.
We use $P_{\mathrm{std},i}=\sqrt{M_i k_B T}$ at the system temperature and $T_{\mathrm{std}}=10\,\mathrm{fs}$.
These scales remain fixed during training and inference, including momentum augmentation.
Generated states are converted back to physical units using the same scales.

\begin{table}[ht]
\centering
\small
\setlength{\tabcolsep}{5pt}
\caption{\textbf{Standardization scales.} Position scales are fitted to the training split.
Momentum scales are listed by element at 300\,K for AL2/AL4 and 310\,K for AL6.}
\label{tab:standardization}
\begin{tabular}{@{}llrrr@{}}
\toprule
Scale & Unit & AL2 & AL4 & AL6 \\
\midrule
$X_{\mathrm{std}}$ & nm & 0.167 & 0.228 & 0.300 \\
$T_{\mathrm{std}}$ & fs & 10 & 10 & 10 \\
$P_{\mathrm{std},i}$ (H) & amu\,nm\,ps$^{-1}$ & 1.585 & 1.585 & 1.611 \\
$P_{\mathrm{std},i}$ (C) & amu\,nm\,ps$^{-1}$ & 5.473 & 5.473 & 5.564 \\
$P_{\mathrm{std},i}$ (N) & amu\,nm\,ps$^{-1}$ & 5.910 & 5.910 & 6.008 \\
$P_{\mathrm{std},i}$ (O) & amu\,nm\,ps$^{-1}$ & 6.317 & 6.317 & 6.421 \\
\bottomrule
\end{tabular}
\end{table}

\phantomsection
\paragraph{Momentum Augmentation}
\label{app:momentum-augmentation}
As done in \citep{ripken2026hfm}, we expose the model to a wider range of source momenta. In particular, we sample $P_i\sim\cN(0,M_i k_B T)$ for each atom $i$, then enforce zero total momentum. For 75\% of training examples, we draw $T \sim \cN(500K, (150K)^2)$, and we clamp the temperature below at $50\,\mathrm{K}$. The remaining examples use the system temperature. This augmentation changes the source momentum distribution while the target Langevin dynamics remain at the system temperature.

\begin{algorithm}[H]
\caption{Training \method with $\cL_{\mathrm{KC}}$ for Peptides}
\label{alg:training}
\begin{algorithmic}[1]
\REQUIRE Kernel $q_{s,t}(y\mid x;\theta)$, target parameters $\eta$, configurations $\cD$, time prior $p(\tau)$
\REPEAT
\STATE $X\sim\cD$, \quad $T\sim\cN(500\,\mathrm{K},(150\,\mathrm{K})^2)$, \quad $P\sim\cN(0,M\,k_BT)$, \quad $x\leftarrow(X,P)$
\STATE $\tau\sim p(\tau)$, \quad $r\sim T_{\min}/2+(\tau-T_{\min})\,\mathrm{Beta}(2,2)$
\STATE $x'\sim q_{0,r}(x'\mid x;\eta)$, \quad $y\sim q_{r,\tau}(y\mid x';\eta)$
\STATE $\cL_{\mathrm{KC}}\leftarrow-\log q_{0,\tau}(y\mid x;\theta)$
\STATE $\theta\leftarrow\textsc{AdamW}(\theta,\nabla_\theta\cL_{\mathrm{KC}})$, \quad $\eta\leftarrow\operatorname{EMA}(\theta)$
\UNTIL{converged}
\end{algorithmic}
\end{algorithm}

\subsection{Inference Details}
\label{app:inference-stabilization}

We apply the following corrections around to our map's proposal to prevent repeated transitions from accumulating errors during long molecular simulations. \Cref{alg:inference} summarizes one step.

\paragraph{Force capping}
While rare, the model can generate a bad configuration which could result in a large force in the subsequent step. To prevent large forces increments, and ultimately prevent non-finite values, we measure the force increment from the reference Gaussian kernel in thermal-momentum units as $\tfrac{h(\tau)}{\tau}\phi_1(\tau)\|v_i(X_{\mathrm{mid}})\|/\sqrt{M_i k_B T}$, where $h(\tau)$ is the duration schedule of \cref{sec:design} and $\phi_1(\tau)=(1-e^{-\gamma\tau})/\gamma$, and we rescale the forces according to a specified cap. The rescaled forces enter the reference mean and conditioning features.

\paragraph{Momentum correction}
Following \citet{ripken2026hfm}, we retain the proposed positions and minimally adjust the proposed momenta to match an internal kinetic-energy target $K_{\mathrm{target}}$. We construct this target to account for the friction and thermal noise of the Langevin dynamics.

To adjust angular momentum and internal kinetic energy separately, we decompose the momenta into rigid rotational and internal components. For positions $X_i$, define the angular momentum and inertia tensor as
\begin{align}
    L=\sum_i X_i\times P_i,
    \qquad
    I=\sum_i M_i\bigl(\|X_i\|^2\mathrm{Id}-X_iX_i^\top\bigr).
\end{align}
The decomposition is
\begin{align}
    P_i=M_i\bigl(I^{-1}L\times X_i\bigr)+P_i^{\mathrm{int}}.
\end{align}
The internal momenta carry no angular momentum, $\sum_i X_i\times P_i^{\mathrm{int}}=0$. Consequently, for any angular velocity $\Omega$,
\begin{align}
    \sum_i\frac{P_i^{\mathrm{int}}\cdot[M_i(\Omega\times X_i)]}{M_i}
    =\Omega\cdot\sum_i X_i\times P_i^{\mathrm{int}}=0.
\end{align}
Therefore, the two components are orthogonal in the mass-weighted inner product, and their kinetic energies add:
\begin{align}
    K^{\mathrm{tot}}=K_{\mathrm{rot}}(L,I)+K^{\mathrm{int}},
    \qquad
    K_{\mathrm{rot}}(L,I)=\frac{L^\top I^{-1}L}{2k_BT},
    \qquad
    K^{\mathrm{int}}=\sum_i\frac{\|P_i^{\mathrm{int}}\|^2}{2M_ik_BT}.
\end{align}
This separation allows us to set the internal kinetic energy by rescaling the internal component while keeping the rigid component. We evaluate these quantities at the source state and the proposed state, denoted by subscripts $t$ and $t+\tau$. Below, $U$ denotes potential energy and $K$ denotes internal kinetic energy, both in units of $k_BT$.

The angular momentum needs no correction, since the map $T$ of \cref{app:model-details} is part of the model and already sets the angular momentum that the flow keeps about the source positions.

For the internal kinetic-energy target, we need to account for friction and thermal noise. As a result, we use the internal-energy law induced by an Ornstein--Uhlenbeck momentum update at fixed positions. This is the closed-form law of the stochastic velocity-rescaling thermostat \citep{bussi2007canonical}:
\begin{align}
    K_{\mathrm{bath}}
    =\frac12\bigl(a\sqrt{2K_t}+\sqrt{1-a^2}\,Z\bigr)^2
    +\frac12(1-a^2)\,Y,
    \qquad a=e^{-\gamma\tau},
\end{align}
where $Z\sim\cN(0,1)$ and $Y\sim\chi^2_{n-1}$ are independent, and $n=3N-6$ is the number of internal degrees of freedom. Then, we account for the change in potential energy and the change in rotational energy at fixed source angular momentum:
\begin{align}
    K_{\mathrm{target}}
    &=U_t-U_{t+\tau}+K_{\mathrm{bath}}+\Delta K_{\mathrm{rot}},\\
    \Delta K_{\mathrm{rot}}
    &=K_{\mathrm{rot}}(L_t,I_t)-K_{\mathrm{rot}}(L_t,I_{t+\tau}).
\end{align}
Here, $\Delta K_{\mathrm{rot}}$ represents rotational energy released into internal motion as the inertia tensor changes. When $\gamma=0$, we have $K_{\mathrm{bath}}=K_t$ where if angular momentum is also conserved, this target enforces conservation of total energy. Unlike \citet{ripken2026hfm}, who conserve angular momentum and total energy and apply a thermostat separately, we incorporate the Langevin thermostat into the target construction.

With the target specified, we choose the smallest change to the proposed momenta in the mass-weighted norm:
\begin{align}
    \min_{P'}\ \sum_i\frac{\|P'_i-P_{t+\tau,i}\|^2}{2M_i}
    \quad\text{s.t.}\quad
    \sum_i X_{t+\tau,i}\times P'_i=L_{t+\tau},
    \qquad
    K^{\mathrm{int}}(P')=K_{\mathrm{target}}.
\end{align}
Orthogonality of the rigid and internal components gives the closed-form correction
\begin{align}
    P'_i
    =M_i\bigl(I_{t+\tau}^{-1}L_{t+\tau}\times X_{t+\tau,i}\bigr)
    +\sqrt{\frac{K_{\mathrm{target}}}{K_{t+\tau}}}\,
    P^{\mathrm{int}}_{t+\tau,i}.
\end{align}
Here $L_{t+\tau}=\sum_iX_{t+\tau,i}\times P_{t+\tau,i}$, so the first term keeps the angular momentum, while the second sets the internal kinetic energy. The positions and total momentum remain unchanged.

For numerical stability, both kinetic energies in the rescaling ratio are floored at $0.1$.

\paragraph{Adaptive proposal retries}
We redraw a proposal if it is non-finite,
$|(U_{t+\tau}+K_{t+\tau})-(U_t+K_t)|>H_{\mathrm{threshold}}$,
$K_{\mathrm{target}}<0.1$, or $\det T<0.05$.
Each attempt uses fresh proposal noise and its own heat-bath draw to compute
$K_{\mathrm{target}}$.
After $N_{\mathrm{retry}}$ unsuccessful attempts at the same step size,
we halve $\tau$, down to a minimum of $1$\,fs.
If all attempts fail, we use the last proposal at the minimum step size.
Only in this case, if $K_{\mathrm{target}}<0.1$, we replace it with
$\max(K_{\mathrm{bath}}+\Delta K_{\mathrm{rot}},0.1)$.
A chain stops if the committed state is non-finite.
The momentum correction uses the duration and heat-bath draw of the committed
proposal, and the chain clock advances by that duration.
The next step starts again at the maximum step size.

\begin{algorithm}[H]
\caption{One inference step}
\label{alg:inference}
\newcommand{\lhs}[1]{\makebox[2.3cm][r]{$#1$}}
\begin{algorithmic}[1]
\REQUIRE Kernel parameters $\theta$, current state $x_t=(X_t,P_t)$, step size $\tau$, $z\sim\Normal(0,I)$
\STATE \lhs{v(X_{\mathrm{mid}})} $\leftarrow\textsc{ForceCapFilter}\bigl(v(X_{\mathrm{mid}})\bigr)$
\STATE \lhs{X_{t+\tau},P_{t+\tau}} $\leftarrow T\Bigl(x_t+m(x_t,\tau)+\Psi\bigl(L(\tau)z;x_t,\theta\bigr)\Bigr)$
\STATE $(X',P',\tau_{\mathrm{used}},K_{\mathrm{bath}})
    \leftarrow\textsc{RedrawFilter}(x_t,X_{t+\tau},P_{t+\tau},\tau)$
\STATE $P'\leftarrow\textsc{MomentumCorrectionFilter}
    (x_t,X',P',\tau_{\mathrm{used}},K_{\mathrm{bath}})$
\STATE \textbf{return} $(X',P',\tau_{\mathrm{used}})$
\end{algorithmic}
\end{algorithm}

\section{Experimental Setup}
\label{app:experimental-setup}

\subsection{Lennard-Jones systems}
\label{app:setup-lj}

\paragraph{Potential}
We use the confined Lennard-Jones systems studied by \citet{kohler2020equivariant,akhound2025pita}.
For $N$ particles, let $d_{ij}=\|x_i-x_j\|$ and $\bar x=N^{-1}\sum_i x_i$. The potential is
\begin{align}
    E(x)=\epsilon\sum_{i<j}\left[\left(\frac{r_m}{d_{ij}}\right)^{12}-2\left(\frac{r_m}{d_{ij}}\right)^{6}\right]
    +\frac{c}{2}\sum_i\|x_i-\bar x\|^2,
\end{align}
with $\epsilon=2$, $r_m=1$ and $c=1$. We learn overdamped Langevin transitions with drift $D\nabla\log\pi(x)$ and diffusion $\sqrt{2D}$, where $\pi(x)\propto\exp(-E(x)/k_BT)$, $k_BT=1$ and $D=0.02$.
Positions and noise are restricted to the zero-center-of-mass subspace, with $3(N-1)$ degrees of freedom.

\paragraph{Data}
At source temperature $T_S=1$, we use the public splits of \citet{akhound-sadegh_iterated_2024}, with 100k/10k/10k training/validation/test configurations for LJ-13 and 10k/10k/10k for LJ-55.
At $T_S=4$, we generate 200k configurations per system using the NUTS \citep{hoffman2014nuts} implementation as done in \citet{akhound2025pita}, with energy $E/4$. In both settings, the target dynamics and evaluation data remain at $T=1$.
Training draws source configurations independently and uniformly with replacement, with a random rotation for each draw.

\paragraph{Model}
The Gaussian reference has mean $x+D\,h(\tau)\nabla\log\pi(x)$ and covariance $2D\tau I$ on the center-of-mass-free subspace, using the duration cap in \cref{sec:design} with $\tau_0=\delta=0.01$. The flow has 9 coupling blocks, each updating one Cartesian coordinate across all particles conditioned on the other two coordinates. Each coupling network uses one transformer layer with AdaLN-Zero conditioning and a SwiGLU feed-forward layer. The context includes the source geometry, time embedding of $\tau$, the reference mean and variance, and force on the source geometry.

\subsection{Alanine peptides}
\label{app:setup-peptides}

\begin{table}[h]
\centering
\small
\setlength{\tabcolsep}{4pt}
\caption{\textbf{Peptide data and physical settings.} Training uses random configurations from these datasets.
The learned dynamics use the listed force fields, temperatures and friction coefficients.
The source trajectories use a 1\,fs integration step with flexible bonds.}
\label{tab:peptide-datasets}
\begin{tabular}{@{}lccc@{}}
\toprule
& AL2 & AL4 & AL6 \\
\midrule
Sequence & Ace-A-Nme & Ace-AAA-Nme & AAAAAA \\
Atoms (dimension $d$) & 22 (66) & 42 (126) & 63 (189) \\
Force field & Amber ff99SB-ILDN & Amber ff99SB-ILDN & Amber ff14SB \\
Implicit solvent & OBC2 & OBC2 & OBC1 \\
Nonbonded cutoff & none & none & 2\,nm \\
Temperature & 300\,K & 300\,K & 310\,K \\
Friction $\gamma$ & 1\,ps$^{-1}$ & 0.3\,ps$^{-1}$ & 0.3\,ps$^{-1}$ \\
Trajectory (frame stride) & 1\,$\mu$s (1\,ps) & 7\,$\mu$s (10\,ps) & 3\,$\mu$s (10\,ps) \\
Random split: train / val / test & 800k / 100k / 100k & 560k / 70k / 70k & 240k / 30k / 30k \\
\bottomrule
\end{tabular}
\end{table}

\paragraph{Data}
For training, AL2 uses the 1\,$\mu$s trajectory of \citet{kohler2021smooth}. AL4 and AL6 use the datasets of \citet{tan_scalable_2025}.  \Cref{tab:peptide-datasets} lists the force fields \citep{lindorff2010ff99sbildn,maier2015ff14sb} and OBC solvent models \citep{onufriev2004obc}.
We randomly split the available configurations 80/10/10 with seed 42, following the split procedure of \citet{ripken2026hfm} for AL2. For AL4 and AL6, we combine the original SBG splits before reshuffling and doing the random splits. During training, we draw positions independently and uniformly from the training split, and do not use trajectory order or paired transitions. Source momenta are sampled independently using the augmentation in \cref{app:momentum-augmentation}. Energies and forces are implemented in JAX \citep{jax2018github} and checked against OpenMM \citep{eastman2017openmm}.

\subsection{Training}
\label{app:setup-training}

\paragraph{Objective}
\label{app:objective}
Our main models use the Kolmogorov--Chapman objective in \cref{eq:ck-kl-mle} with two transitions from the EMA target kernel (\cref{alg:training}).
We split each sampled elapsed time $\tau$ into two durations using a rescaled $\mathrm{Beta}(2,2)$ distribution, with each step lasting at least $T_{\min}/2$. The minimum elapsed time $T_{\min}$ is listed in \cref{tab:training-hparams}. For numerical stability, we remove the parameter-independent terms from the NLL, divide by the number of unconstrained degrees of freedom and cap each sample's NLL at 10. Additionally, we train with the detailed-balance objective in \cref{eq:db-obj} in the last $15\%$ of training steps over $25\%$ of the batch, with $\lambda$ increasing linearly from $0$ to $0.2$ over $5K$ steps for all molecular systems. Source momenta for the detailed-balance objective are sampled at the target temperature. We omit this objective in the ablations.

\paragraph{Optimization}
We use AdamW \citep{loshchilov2019adamw} with $\beta=(0.9,0.95)$ for AL2 and AL4 and $\beta_2=0.9025$ for AL6, $\epsilon=10^{-8}$, weight decay $10^{-8}$ and a global gradient-norm limit of 1. The learning rate warms up linearly to $3\times10^{-4}$ and then decays to $3\times10^{-6}$ with a cosine schedule.  Elapsed times follow the mixture prior in \cref{sec:design}, with $T_{\max}$ increased log-linearly during early training. \Cref{tab:training-hparams} lists the settings for each system. The LJ PF-ODE comparison uses the same architecture, with $T_{\max}=0.05$ annealed from 0.008. Its training budget matches the FLOPs of KC-KL at $T_S=1$, resulting in 275k updates for LJ-13 and 545k for LJ-55. We train on a heterogenous cluster consisting of L40s and H100s.

\begin{table}[h]
\centering
\small
\setlength{\tabcolsep}{4pt}
\caption{\textbf{Training hyperparameters.} LJ times are in reduced units. For LJ, $T_{\max}$ is given for $T_S=1$ / $T_S=4$.}
\label{tab:training-hparams}
\begin{tabular}{@{}lccccc@{}}
\toprule
& LJ-13 & LJ-55 & AL2 & AL4 & AL6 \\
\midrule
Blocks $\times$ attention layers & $9\times1$ & $9\times1$ & $8\times2$ & $8\times2$ & $8\times2$ \\
Channels / heads & 256 / 4 & 256 / 4 & 384 / 6 & 384 / 6 & 384 / 6 \\
Parameters & 9.4M & 9.4M & 46.6M & 46.6M & 46.6M \\
Gate orders $(n_a,n_b)$ & $(1,2)$ & $(1,2)$ & \multicolumn{3}{c}{$x$: $(3,3)$, $p$: $(2,3)$} \\
Fixed $\omega$ (all gates) & \multicolumn{2}{c}{$40$} & \multicolumn{3}{c}{$333\,\mathrm{ps}^{-1}$} \\
Reference cap $\tau_0=\delta$ & 0.01 & 0.01 & \multicolumn{3}{c}{0.75\,fs} \\
Minimum elapsed time $T_{\min}$ & $2\times10^{-5}$ & $2\times10^{-5}$ & \multicolumn{3}{c}{0.02\,fs} \\
$T_{\max}$ & 0.5 / 0.1 & 0.5 / 0.1 & 100\,fs & 85\,fs & 85\,fs \\
$T_{\max}$ ramp (start, duration) & 0.05, 10\% & 0.05, 10\% & \multicolumn{3}{c}{1\,fs, 15\%} \\
Log-uniform weight $w$ & 0.4 & 0.4 & 0.25 & 0.25 & 0.25 \\
Batch size & 256 & 256 & 512 & 512 & 1024 \\
Training steps & 300k & 600k & 750k & 1.5M & 1M \\
Learning-rate warm-up & \multicolumn{5}{c}{$0.5\%$} \\
Two-step EMA decay & \multicolumn{5}{c}{$0.995$} \\
Evaluation EMA decay & \multicolumn{5}{c}{$0.995$} \\
\bottomrule
\end{tabular}
\end{table}

\paragraph{HFM training}
\label{app:hfm-training}
For AL4 and AL6, we train HFM using the public implementation of \citet{ripken2026hfm}, with the same data splits and force fields as our method. The model has six transformer layers, 256 channels and eight attention heads.
We use a global batch size of 512 and seed 42. AdamW uses $\beta=(0.9,0.95)$, no weight decay and gradient clipping at norm 5. The learning rate increases from $10^{-6}$ to $10^{-4}$ over the first 1\% of updates, then follows cosine decay to $10^{-8}$. The maximum elapsed time is 10\,fs, with 75\% of training samples assigned zero elapsed time and the remainder drawn using HFM's improved time sampler. The momentum-augmentation temperature has mean 300\,K for AL4 and 310\,K for AL6, with standard deviation 150\,K. We evaluate the final EMA parameters.

\paragraph{Training compute}
We set the HFM update counts to match the estimated FLOPs of our method on each system (\cref{tab:hfm-training-compute}). We estimate FLOPs using JAX 0.9.2 CPU cost analysis of the loss and its parameter gradients, multiplied by the number of training updates. This includes model evaluations, teacher evaluations, Jacobian--vector products and force computations within the loss.

\begin{table}[ht]
\centering
\small
\setlength{\tabcolsep}{5pt}
\renewcommand{\arraystretch}{1.08}
\caption{\textbf{Training compute on AL4 and AL6.} Parameter counts, completed training updates and estimated FLOPs for the loss and parameter gradients. FLOP totals follow the accounting described above.}
\label{tab:hfm-training-compute}
\begin{tabular}{@{}llrrrr@{}}
\toprule
System & Method & Parameters (M) & Batch size & Updates & FLOPs ($10^{18}$) \\
\midrule
AL4 & HFM     & 9.09  & 512  & $2\,061\,097$ & 15.513 \\
    & \method & 46.58 & 512  & $1\,500\,000$ & 15.513 \\
\midrule
AL6 & HFM     & 9.09  & 512  & $1\,957\,876$ & 31.299 \\
    & \method & 46.58 & 1024 & $1\,000\,000$ & 31.299 \\
\bottomrule
\end{tabular}
\end{table}

\subsection{Evaluation}
\label{app:setup-eval}

\paragraph{Lennard-Jones}
We run 5000 parallel Metropolis--Hastings chains \citep{hastings1970} for 5000 steps, using learned proposals at $\tau=0.05$ for LJ-13 and $\tau=0.02$ for LJ-55. Each step computes the Metropolis-Hastings acceptance ratio to accept or reject the proposal. All chains start from the same energy-minimized configuration, and we evaluate their final states. MALA uses the same initialization and number of chains. We sweep physical step sizes $h\in\{10^{-4},3\times10^{-4},10^{-3},3\times10^{-3},10^{-2},0.02,0.05\}$, with $5000$ steps of sampling on one L40S per setting. We select $h$ using validation Geo-$\mathcal{W}_2$, giving $3\times10^{-4}$ for LJ-13 and $10^{-4}$ for LJ-55.

Energy $\mathcal{W}_2$ compares the 5000 endpoint energies with the 10k test energies.
Geo-$\mathcal{W}_2$ compares the endpoints with 5000 test configurations as done in \citep{klein_equivariant_2023}.  Published baseline values are taken from \citet{potaptchik2025tilt}.

\paragraph{Molecular reference trajectories}
Training uses samples drawn independently from the datasets provided in \citep{kohler2021smooth} for AL2 and \citep{tan_scalable_2025} for AL4 and AL6, whereas evaluation compares generated trajectories with long MD references. The original AL2 data were generated with the legacy \texttt{LangevinIntegrator} in an older OpenMM release \citep{eastman2017openmm}. To reduce discretization bias in evaluation and ensure accurate coverage of all modes, we generate new AL2, AL4, and AL6 references with OpenMM 8.5.1 using \texttt{LangevinMiddleIntegrator} to serve as ground truth \citep{leimkuhler2013robust}, keeping the settings in \cref{tab:peptide-datasets} and a 1\,fs integration step.
The references for all systems were run with 16 independent chains where each chain was simulated for 2\,$\mu$s, with frames saved every 1\,ps and the first 1\,ns discarded (burn-in). Each run took $300$ GPU-hours, $312$ GPU-hours, and $520$ GPU-hours for AL2, AL4, and AL6, respectively.

\paragraph{Molecular sampling}
For the results reported in \cref{tab:mol-results}, we run each setting for one GPU-hour on single L40S with 4 CPU cores, excluding compilation and initialization. All methods start from the same configuration for each molecule, with independently sampled momenta at the target temperature.
\begin{itemize}[leftmargin=*,topsep=3pt,itemsep=4pt,parsep=0pt,partopsep=0pt,before=\raggedright]
    \item \textbf{MD:} OpenMM LangevinMiddle integrator with a 1\,fs step, running 1, 4 or 16 chains.
    \item \textbf{HFM:} the sampler of \citet{ripken2026hfm}, including its energy and momentum corrections.
    We use $\tau\in\{5,7\}$\,fs with $\{1,16,64\}$ chains for AL2 and AL4, and $\tau\in\{3,5\}$\,fs with $\{1,4,16\}$ chains for AL6.
    AL2 uses the released checkpoint. AL4 and AL6 use the final EMA checkpoints described in \cref{app:hfm-training}.
    \item \textbf{\method:} $\tau\in\{5, 10, 20, 30, 40, 50, 60, 70\}$\,fs with $\{8, 16,32\}$ chains for all three molecular systems.
    We apply the corrections in \cref{app:inference-stabilization}, with a dimensionless force cap of 7, $N_{\mathrm{retry}}=5$ attempts per step size and $H_{\mathrm{threshold}}=40$ for energies in units of $k_BT$. Failed proposals trigger step-size halving down to 1\,fs.
\end{itemize}
For each method, \cref{tab:mol-results} reports the configuration with the best average across the three metrics:
MD with 16 chains for all systems; HFM with $\tau=7$\,fs and 16 / 64 chains for AL2 / AL4, and $\tau=5$\,fs with 1 chain for AL6;
\method with $\tau=30$\,fs and 16 chains for AL2 and AL6, and $\tau=40$\,fs and 8 chains for AL4.

\paragraph{Equilibrium Metrics}
For the molecular equilibrium metrics, we discard the first 10\% of each chain's simulated time.
We pool the remaining frames and draw $10^4$ without replacement, weighted by the simulated interval each frame represents.
We compare these with $10^4$ uniformly drawn reference frames using exact optimal transport, and report the mean and standard deviation over three subsampling draws.

Following \citet{tan2026amortized}, we compare the generated and reference configurations using empirical Wasserstein-2 distances.
For samples $\{x_i\}_{i=1}^n$ and $\{y_j\}_{j=1}^m$, define the empirical distributions $\mu=\tfrac{1}{n}\sum_{i=1}^n\delta_{x_i}$ and $\nu=\tfrac{1}{m}\sum_{j=1}^m\delta_{y_j}$. We compute
\begin{align}
    \mathcal{W}_{2,c}(\mu,\nu)
    =\left[\min_{\pi\in\Pi(\mu,\nu)}\sum_{i=1}^n\sum_{j=1}^m\pi_{ij}\,c(x_i,y_j)^2\right]^{1/2},
\end{align}
where $\Pi(\mu,\nu)$ contains couplings with marginals $\mu$ and $\nu$. The squared cost $c(x,y)^2$ determines which properties of the configurations are compared.

For $\mathcal{E}$-$\mathcal{W}_2$, let $\mathcal{E}(x)$ denote the potential energy in units of $k_BT$. We use
\begin{align}
    c_{\mathcal{E}}(x,y)^2=\bigl|\mathcal{E}(x)-\mathcal{E}(y)\bigr|^2.
\end{align}
For $\mathcal{T}$-$\mathcal{W}_2$, let $\operatorname{Dihedrals}(x)$ collect all backbone $(\phi,\psi)$ angles in radians.
Its length is $D=2,6,10$ for AL2, AL4 and AL6, respectively. We account for angle periodicity using
\begin{align}
    c_{\mathcal{T}}(x,y)^2
    =\sum_{\ell=1}^{D}\Bigl[\bigl(\operatorname{Dihedrals}(x)_\ell-\operatorname{Dihedrals}(y)_\ell+\pi\bigr)\bmod 2\pi-\pi\Bigr]^2.
\end{align}
Each term is the squared shortest angular difference, so angles separated by $2\pi$ are equivalent.

For TICA-$\mathcal{W}_2$, we compare configurations along slowly varying coordinates identified by time-lagged independent component analysis \citep{perez2013tica}.
Let $x_t$ denote the mean-centered molecular feature time series, comprising pairwise distances among carbon and nitrogen atoms and the sine and cosine of backbone angles, with the reference feature mean subtracted. For a trajectory of $M$ frames and lag $\ell$, the instantaneous and time-lagged covariance estimates with uniform frame weights are
\begin{align}
    \widehat C_{00}&=\frac{1}{M-\ell}\sum_{t=1}^{M-\ell}x_tx_t^\top,
    \qquad
    \widehat C_{0\ell}=\frac{1}{M-\ell}\sum_{t=1}^{M-\ell}x_tx_{t+\ell}^\top.
\end{align}
TICA finds directions with high autocorrelation at lag $\ell$ by solving the generalized eigenvalue problem
\begin{align}
    \widehat C_{0\ell}w_j=\lambda_j\widehat C_{00}w_j
\end{align}
where $\lambda_j$ measures the autocorrelation of the projected component at lag $\ell$, and $w_j$ defines the corresponding direction, normalized so that $w_j^\top\widehat C_{00}w_j=1$.
We retain the first two components and use the squared transport cost
\begin{align}
    c_{\mathrm{TICA}}(x,y)^2=\sum_{j=1}^{2}\lambda_j^2\bigl(w_j^\top x-w_j^\top y\bigr)^2.
\end{align}
In practice, we fit the Koopman-reweighted covariance estimates using deeptime \citep{hoffmann2021deeptime}. Reference trajectories have frames space at 1\,ps. We use a lag of 100 frames (100\, ps) for AL2 and 1000 frames (1\, ns) for AL4 and AL6. The TICA projections are fitted on the long MD reference trajectories consisting of 16 chains of 2 $\mu$s of simulation per chain. We apply that same projection to every method's samples and the reference configurations.

\paragraph{Mode discovery}
For the discovery curves in \cref{fig:gpu-hour}, we use the TICA projections
fitted on each system's $32\,\mu$s MD reference with a $1$\,ns lag.
We divide the TICA plane into four regions using splits at
$\mathrm{TIC0}=-3$ and $\mathrm{TIC1}=-2$ for AL4, and
$\mathrm{TIC0}=-2$ and $\mathrm{TIC1}=-2$ for AL6.
A mode counts as discovered once a single chain remains in its region for
$20$\,ps of simulated time. We record the GPU time of the frame that completes
this stay. We include burn-in frames from the one-hour runs, but exclude
the startup segments of \method.

\paragraph{Kinetic metrics}
\label{app:kinetic-metrics}
We measure the mean first-passage time (MFPT) between modes and the autocorrelation of mode occupancy, both from trajectories saved every 1\,ps.

\paragraph{Mode labels}
We obtain the modes using the reference MD free energy $F(\phi,\psi)=-k_BT\ln p(\phi,\psi)$, obtained by a watershed on a $120\times120$ grid. Modes separated by saddles below $0.5\,k_BT$ or holding under $0.5\%$ of the population are merged. A mode's core is the region of its basin with $F\leq F_{\min}+2\,k_BT$. A frame inside a mode's core takes that mode's label. A frame outside all cores keeps the label of the last core visited, so brief excursions do not count as transitions. Frames before the first core visit are unlabelled.

\textbf{Mean first-passage time.}
For distinct modes $S$ and $D$, let $s_k$ be the time of the $k$-th entry into $S$ from another mode (the first core visit is not an entry). The first-passage time from this entry is the time until the trajectory first enters the core of $D$,
\begin{align}
    T_k^{S\to D}=\inf\{u>0:X_{s_k+u}\in D\},
    \qquad
    \tau_{S\to D}=\mathbb{E}\big[T_k^{S\to D}\big].
\end{align}
If the trajectory ends before reaching $D$, we only know that $T_k^{S\to D}$ exceeds the remaining time $C_k$.
To keep these censored passages, we model passage times as exponential and use the maximum-likelihood estimate
\begin{align}
    \widehat\tau_{S\to D}
    =\frac{\sum_k\min\big(T_k^{S\to D},C_k\big)}
    {\sum_k\mathbf{1}\big\{T_k^{S\to D}\leq C_k\big\}},
\end{align}
the total observed waiting time divided by the number of completed passages.
Without censoring, this is the sample mean of the passage times.

\textbf{Autocorrelation.}
For each method, let $\pi_A$ be the fraction of its labelled frames in mode $A$.
The autocorrelation of mode occupancy is
\begin{align}
    C_A(t)=\frac{P_{AA}(t)-\pi_A}{1-\pi_A},
\end{align}
where $P_{AA}(t)$ is the fraction of frames labelled $A$ that are labelled $A$ again a lag $t$ later. $C_A$ decays from 1 to 0 as the dynamics forget which mode they started in.

\textbf{Transition probabilities.}
For modes $i$ and $j$, $P_{ij}(t)$ is the fraction of frames labelled $i$ that are labelled $j$ a lag $t$ later, pooled over chains.
As $t\to\infty$, $P_{ij}(t)$ converges to the population of mode $j$, so it tests both the relaxation time scales and the stationary distribution.

\clearpage
\section{Additional Experiments}
\label{app:additional-experiments}
\label{app:ablations}

\subsection{Equilibrium Results across step size}
\label{app:equilibrium-step-size}
\begin{figure}[H]
\centering
\includegraphics[width=0.9\linewidth]{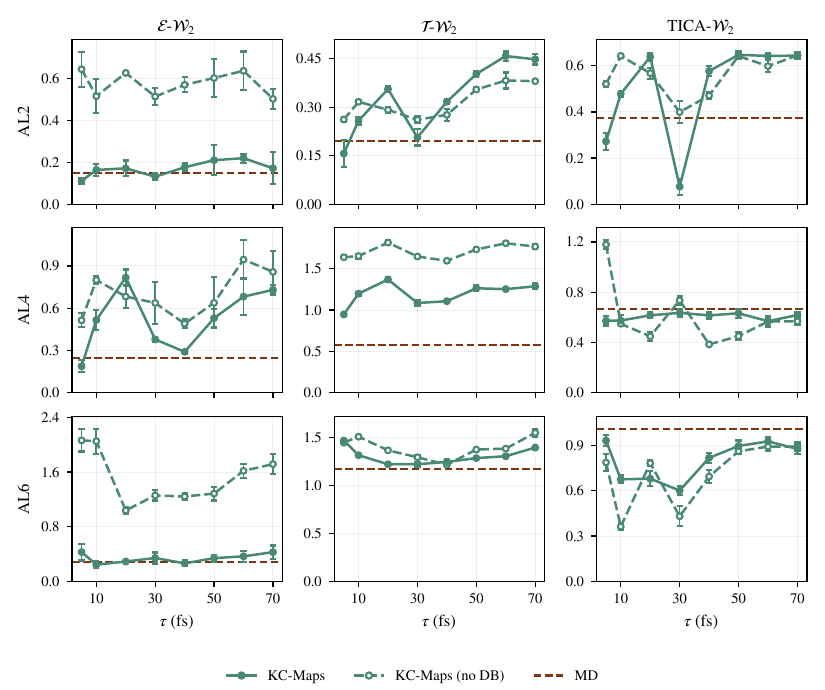}
\caption{\textbf{Equilibrium results across step size.} Equilibrium metrics for \method with and without detailed-balance loss against reference long MD.}
\label{fig:equilibrium-w2-vs-tau}
\end{figure}

In \cref{fig:equilibrium-w2-vs-tau}, we show the performance of \method across $\tau \in [5, 70]\,$fs on equilibrium sampling against the baseline molecular dynamics run on a single GPU-hour. We compare both methods to the reference molecular dynamics simulation (16 chains $\times$ 2 $\mu$s).

\noindent\begin{minipage}{\linewidth}
Moreover, we demonstrate the ability of our method to extrapolate beyond its training distribution and recover modes not seen during training. Namely, for the AL6 system, \cref{fig:al6-extrapolation-tica} shows the reference distribution from long MD, the training data distribution on the train split from \citep{tan_scalable_2025}, which has the missing bottom left mode, and our samples from long simulation from our method (16 chains $\times$ $400$ ns per chain for a total of $40$ GPU-hours).

\begin{figure}[H]
\centering
\includegraphics[width=\linewidth]{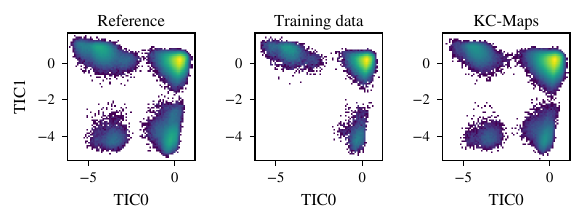}
\caption{\textbf{AL6 sampling beyond the training distribution.}
TICA projections of reference long MD, training data, and \method samples.
The \method samples are pooled from 16 chains of approximately $400\,\mathrm{ns}$ each. The lower-left mode is absent from the training configurations but visited by \method.}
\label{fig:al6-extrapolation-tica}
\end{figure}
\end{minipage}

\subsection{Throughput across step sizes}
\label{app:throughput-step-size}
We show the throughput measured in simulated time across $\tau$ for our method against a single chain of MD. We see in smaller systems, the method lags behind in simulated time for a single chain below $\tau=50$ fs. However, this gap closes as system sizes grow larger e.g. \method exceeds MD throughput for single chain at $\tau=40$ fs for AL6. Moreover, as seen in \cref{fig:gpu-hour}, \method better utilizes GPU resources when simulating parallel chains in a single GPU, highlighting the advantage of deep learning based approaches for accelerated molecular dynamics.

\begin{figure}[H]
\centering
\includegraphics[width=\linewidth]{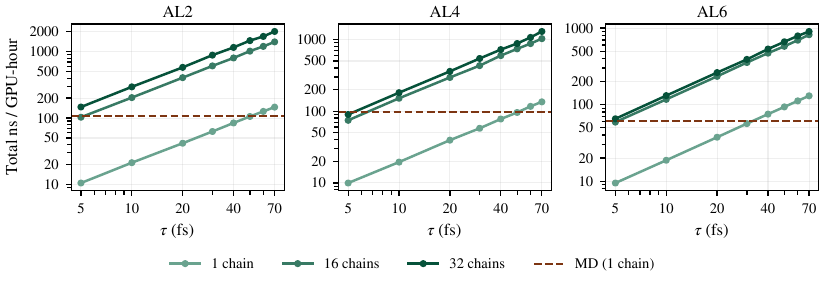}
\caption{Aggregate physical time simulated across chains across $\tau$ for $\texttt{n\_chains} \in {1, 16, 32}$. All runs on a single L40S GPU.}
\label{fig:throughput-vs-tau}
\end{figure}

\subsection{Metropolis--Hastings acceptance across step sizes}
\label{app:mh-acceptance-step-size}
\noindent
\begin{minipage}[t]{0.56\linewidth}
\vspace{0pt}
We show the Metropolis--Hastings acceptance rates across different $\tau$. We observe strong acceptances at small $\tau$ for all systems. We evaluate each system using 10K sources with average across 16 proposals per source across $\tau \in [5, 100]\,$fs for AL2 and $\tau \in [5, 85]\,$fs AL4 and AL6. For AL2, acceptance remains above $20\%$ for the evaluated step sizes $\tau<80$\,fs. For AL4 and AL6, the acceptance rates considerably drop and are practically unusable for simulation for larger $\tau$. The force-field and simulation settings vary across systems (\cref{tab:peptide-datasets}) and may also affect acceptance rates.
\end{minipage}\hfill
\begin{minipage}[t]{0.40\linewidth}
    \vspace{0pt}
    \centering
    \includegraphics[width=\linewidth]{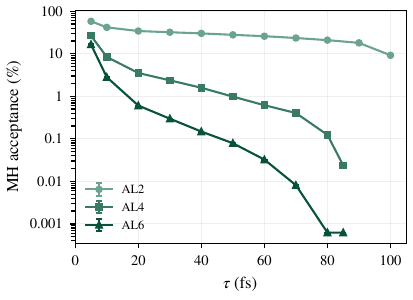}
    \captionsetup{font=footnotesize}
    \captionof{figure}{\textbf{Acceptance rates across step size for \method.}}
\label{fig:mh-acceptance-vs-tau}
\end{minipage}

\subsection{Comparison to MLE on Trajectory Data}
\begin{figure}[H]
\centering
\includegraphics[width=\linewidth]{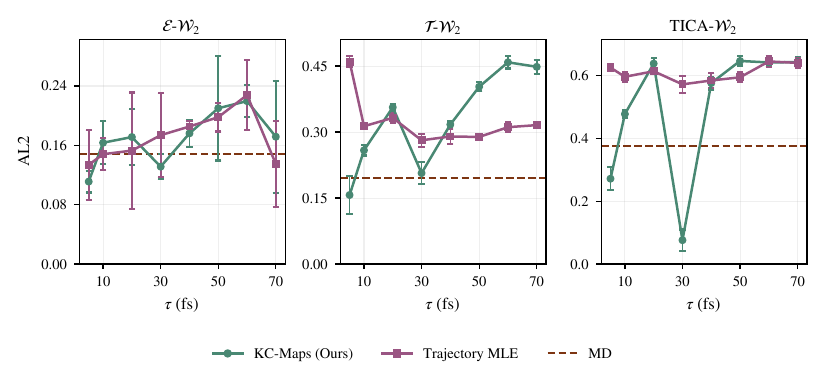}
\caption{\textbf{Equilibrium accuracy on AL2 across step sizes: trajectory MLE vs.\ $\cL_{KC}$.} 2-Wasserstein distances to reference MD for 16 chains run for one hour on a single L40S GPU at each $\tau$.
Error bars are the standard deviation over three subsamples of the trajectory. The dashed line is MD with 16 chains for one hour.}
\label{fig:al2-mle-w2-vs-tau}
\end{figure}

We compare our method under same settings, i.e., model parameterization, training steps, and with $\cL_{DB}$ loss, but instead train using MLE on trajectory data from MD. In \cref{fig:al2-mle-w2-vs-tau}, we see our method is competitive to MLE directly from data, showcasing that we do not need trajectory data to learn accelerated dynamics for $\tau$ up to 70 fs.

\clearpage
\subsection{AL2 transition probabilities}
\label{app:al2-pij}

\begin{figure}[H]
\centering
\includegraphics[width=\linewidth]{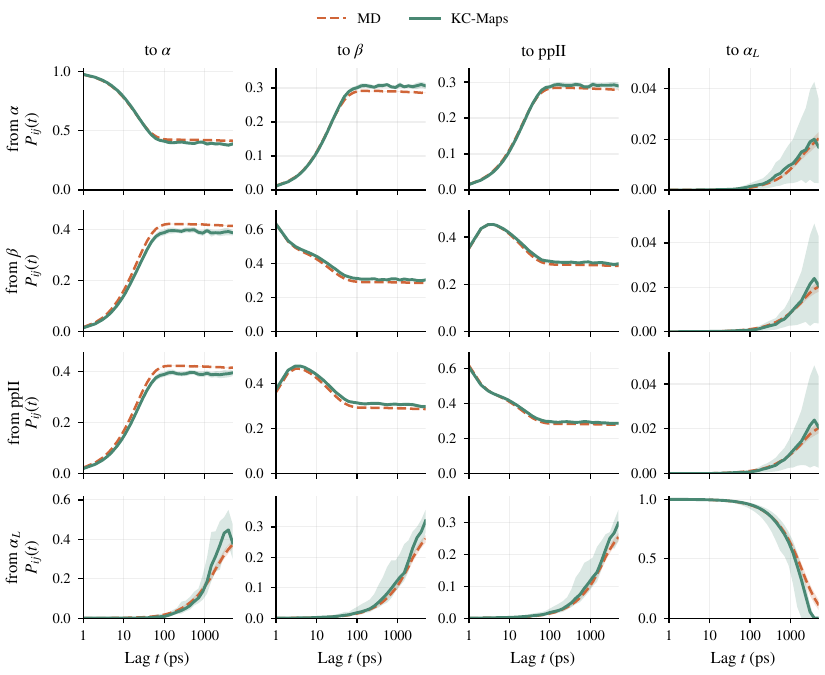}
\caption{\textbf{AL2 transition probabilities.} $P_{ij}(t)$ from mode $i$ (rows) to mode $j$ (columns) against the lag $t$.
Shaded bands are 95\% bootstrap intervals over chains.}
\label{fig:al2-kinetics-pij}
\end{figure}

\cref{fig:al2-kinetics-pij} reports the transition probabilities $P_{ij}(t)$ of \method and reference long MD between the four AL2 modes (\cref{app:kinetic-metrics}). We assign each frame a model label as described in \cref{app:kinetic-metrics}, discarding the first 10\% of each trajectory for \method and the first 1\,ns for each MD trajectory. 

We evaluate the lags on a logarithmically spaced grid from 1\,ps to 5\, ns. At each lag $t$, we pool all pairs of frames separated by $t$. within the same trajectory across 16 trajectories. We estimate $P_{ij}(t)$ as the number of pairs starting in mode $i$ and ending in mode $j$, divided by the number starting in mode $i$.

For each method, we estimate uncertainty by bootstrapping trajectories
\citep{grossfield2018best}. We draw 16 trajectories with replacement,
pool their transition counts, and recompute $P_{ij}(t)$.
We repeat this procedure 2000 times. At each lag, the 2.5th and 97.5th
percentiles of these estimates define the 95\% confidence interval.

\clearpage
\subsection{Target EMA decay}
\begin{figure}[H]
\centering
\includegraphics[width=0.8\linewidth]{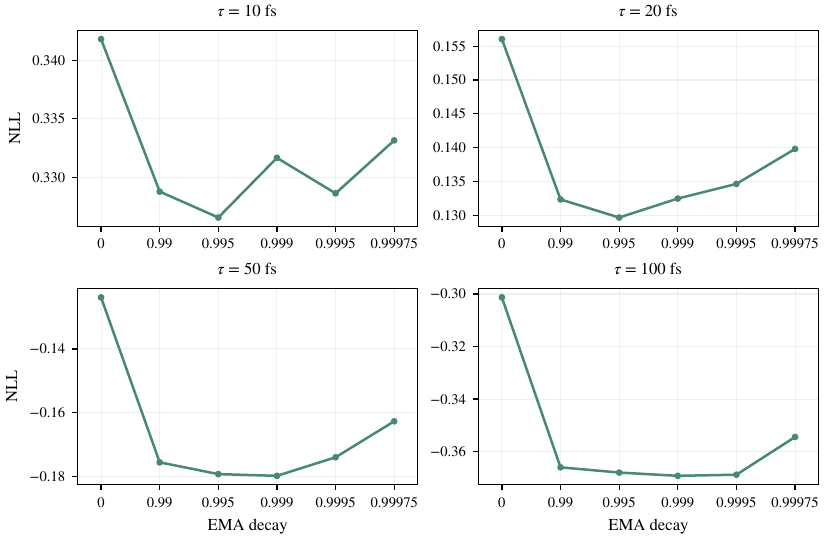}
\caption{\textbf{Target EMA decay on AL2.}
Models differ only in $\mathtt{ema\_decay}$ for the two-step target $\eta=\operatorname{EMA}(\theta)$ in \cref{eq:ck-kl-mle}. All models are trained for 500K steps with batch size 512. We report NLL on MD transition samples at $\tau\in\{10,20,50,100\}$\,fs. A decay of zero corresponds to $\eta=\operatorname{stopgrad}(\theta)$.}
\label{fig:al2-ema-sweep}
\end{figure}

In \cref{fig:al2-ema-sweep}, the stop-gradient target ($\mathtt{ema\_decay}=0$) has the highest NLL at every elapsed time, decays between 0.99 and 0.9995 perform similarly, and the slowest decay, 0.99975, degarding in performance.

\subsection{Flow parameterizations}
\label{app:flow-parameterizations}

We compare three coordinate choices for the affine coupling flow, using the same Gaussian reference $\mu(x,\tau)+L(\tau)z$, where $z\sim\Normal(0,I)$, $\mu(x,\tau)=x+m(x,\tau)$, and $\Sigma(\tau)=L(\tau)L(\tau)^\top$. Writing $\Psi$ for the coupling map in the chosen coordinates, the three parameterizations are
\begin{align}
    \text{Standard:}\qquad &y=\Psi\bigl(x+m(x,\tau)+L(\tau)z;x,\theta\bigr), \nonumber\\
    \text{Displacement:}\qquad &y=x+m(x,\tau)+\Psi\bigl(L(\tau)z;x,\theta\bigr), \nonumber\\
    \text{Whitened:}\qquad &y=x+m(x,\tau)+L(\tau)\Psi(z;x,\theta).
    \label{eq:flow-parameterizations}
\end{align}
The standard flow transforms the full joint state, the displacement flow transforms residuals about the reference mean, and the whitened flow transforms standard Gaussian coordinates. All three recover the Gaussian reference when $\Psi$ is the identity. 

\noindent\begin{minipage}{\linewidth}
We adapt the gates to each coordinate choice. Using $\kappa_n(\tau)=(\omega\tau)^n/[1+(\omega\tau)^n]$, the gates are
\begin{align}
    \begin{array}{l|cc|cc}
        & A_x & A_p & B_x & B_p \\
        \hline
        \text{Standard}     & \kappa_3 & \kappa_2 & \kappa_3 & \kappa_2 \\
        \text{Displacement} & \kappa_2 & \kappa_2 & \kappa_3 & \kappa_2 \\
        \text{Whitened}     & \kappa_2 & \kappa_2 & \sqrt{\kappa_3} & \sqrt{\kappa_3},
    \end{array}
\end{align}
where $A$ gates the log scale and $B$ gates the shift in the corresponding coordinates.
For bounded coupling outputs, these choices give physical corrections of at most $O(\tau^3)$ in position and $O(\tau^2)$ in momentum in all three representations. These gate orders differ from the main model (see \cref{app:reference-gate-orders}).

\begin{figure}[H]
\centering
\includegraphics[width=\linewidth]{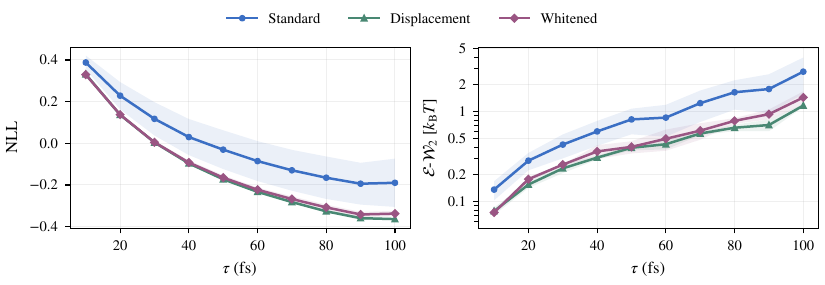}
\caption{\textbf{Flow parameterization on AL2.} Comparison of standard, displacement, and whitened flows, each trained for 500K steps with batch size 512 and evaluated using EMA parameters with decay $0.999$.
Lines and shaded bands show the mean and standard deviation across training seeds: $42$, $43$, and $44$ for all three parameterizations.
Left: NLL per degree of freedom (DoF) on molecular dynamics samples.
Right: $\mathcal{E}$-$\mathcal{W}_2$ in units of $k_\mathrm{B}T$, averaged over source configurations.}
\label{fig:al2-parameterization}
\end{figure}

In \cref{fig:al2-parameterization}, displacement and whitened flows perform similarly at short elapsed times, while the displacement flow has lower mean NLL and $\mathcal{E}$-$\mathcal{W}_2$ at longer elapsed times. Both improve on the standard flow across the evaluated step sizes.
\end{minipage}

%%%%%%%%%%%%%%%%%%%%%%%%%%%%%%%%%%%%%%%%%%%%%%%%%%%%%%%%%%%%

\end{document}